\documentclass[twoside]{article}

\usepackage[accepted]{aistats2025}
\usepackage[round]{natbib}

\usepackage{algorithm}
\usepackage{algorithmic}

\usepackage{microtype}
\usepackage{graphicx}
\usepackage{subfigure}
\usepackage{booktabs}

\usepackage{hyperref}

\usepackage{amsmath}
\usepackage{amssymb}
\usepackage{mathtools}
\usepackage{amsthm}

\usepackage[capitalize,noabbrev]{cleveref}

\theoremstyle{plain}
\newtheorem{theorem}{Theorem}[section]
\newtheorem{proposition}[theorem]{Proposition}

\theoremstyle{definition}

\theoremstyle{remark}

\usepackage[textsize=tiny]{todonotes}

\begin{document}

%

%

\twocolumn[

\aistatstitle{Functional Stochastic Gradient MCMC for Bayesian Neural Networks}

\aistatsauthor{Mengjing Wu \And Junyu Xuan \And  Jie Lu }

\aistatsaddress{ Australian Artificial Intelligence Institute (AAII), University of Technology Sydney } ]

\begin{abstract}
 Classical parameter-space Bayesian inference for Bayesian neural networks (BNNs) suffers from several unresolved prior issues, such as knowledge encoding intractability and pathological behaviours in deep networks, which can lead to improper posterior inference. To address these issues, functional Bayesian inference has recently been proposed leveraging functional priors, such as the emerging functional variational inference. In addition to variational methods, stochastic gradient Markov Chain Monte Carlo (MCMC) is another scalable and effective inference method for BNNs to asymptotically generate samples from the true posterior by simulating continuous dynamics. However, existing MCMC methods perform solely in parameter space and inherit the unresolved prior issues, while extending these dynamics to function space is a non-trivial undertaking. In this paper, we introduce novel functional MCMC schemes, including stochastic gradient versions, based on newly designed diffusion dynamics that can incorporate more informative functional priors. Moreover, we prove that the stationary measure of these functional dynamics is the target posterior over functions. Our functional MCMC schemes demonstrate improved performance in both predictive accuracy and uncertainty quantification on several tasks compared to naive parameter-space MCMC and functional variational inference.
\end{abstract}

\section{Introduction}
\label{sec:intro}
Compared to traditional deep neural networks that rely on a single pattern of model parameters only, Bayesian neural networks (BNNs) demonstrate principled predictive uncertainty estimation and improved generalization by integrating models under the posterior distribution over network parameters \citep{neal1995bayesian, gal2016uncertainty, wilson2020bayesian}. As Bayesian inference methods continue to advance and gain success within the deep learning community, BNNs have been applied in a wide variety of domains and tasks, including dissolution prediction of planetary systems \citep{cranmer2021bayesian}, medical diagnosis as diabetic detection \citep{filos2019systematic, band2022benchmarking}, classification of radio galaxies \citep{mohan2024evaluating}, and other safety-critical environments \citep{rudner2022tractable}. 

Choosing an appropriate prior is critical for BNNs. The common practice is to use isotropy Gaussian priors for all network parameters, often referred to as parameter-space priors. However, these parameter-space priors
have shown several problematic issues. For instance, as the depth of the network increases, the function samples of Gaussian prior over parameters tend to be horizontal \citep{duvenaud2014avoiding, matthews2018gaussian}. Additionally, with ReLU activations, the prior distribution of unit outputs becomes more heavy-tailed for deeper architectures.\citep{vladimirova2019understanding, tran2022all}. Furthermore, the effects of this parameter-space prior to posterior inference remain an open question due to the non-interpretability of network parameters and the complexity of network architectures \cite{fortuin2021bayesian, wild2022generalized}, which make it challenging to translate valid prior knowledge into a corresponding prior distribution over network parameters. Due to these unresolved issues with parameter-space priors, there has been growing interest in performing Bayesian inference directly in function space, where more informative stochastic processes can serve as priors. For example, Gaussian processes (GPs), which can easily encode prior knowledge about the potential functions via kernel functions, have been widely used as priors in functional variational inference for BNNs \citep{sun2019functional, ma2021functional, rudner2022tractable, pielok2022approximate}. These functional variational inference approaches have shown improved performance compared to parameter-space variational inference.


For posterior inference given a prior, Markov Chain Monte Carlo (MCMC) is another valid method for BNNs apart from variational inference. Unlike variational inference, which makes strong assumptions about the posterior distribution, non-parametric MCMC methods are asymptotically exact, making them the gold standard for Bayesian inference \citep{alexos2022structured}. Even better, the computationally expensive MCMC methods became feasible for modern BNNs with large data following the introduction of their scalable stochastic gradient variants (SGMCMC) \citep{welling2011bayesian, patterson2013stochastic, chen2014stochastic, Zhang2020Cyclical, nemeth2021stochastic}. However, existing MCMC methods for BNNs typically perform in parameter space and are based on the diffusion dynamics designed in terms of the posterior over parameters, inheriting similar prior issues as parameter-space variational inference. 
Extending these methods to function space, however, is a non-trivial task. It requires the design of novel dynamics that not only sample effectively from the function space but also guarantee that the resulting stationary distribution corresponds to the target posterior over functions. This involves addressing the inherent complexities of functional spaces while ensuring the correctness and feasibility of the sampling process.

In this work, we propose novel functional MCMC schemes grounded in the fundamental  It{\^o} lemma for stochastic calculus \citep{ito1951formula, brzezniak2008ito,oksendal2013stochastic}. Specifically, the dynamics of BNNs in function space are designed by introducing new potential energy functionals that can incorporate functional priors, such as GPs, into the diffusion processes and simultaneously guarantee the target posterior over functions as the stationary measure. 
Our main contributions are as follows:
\begin{itemize}
    \item We propose novel functional diffusion dynamics, along with their stochastic gradient versions for BNNs that effectively incorporate more informative functional priors into posterior inference. Additionally, we derive corresponding tractable MCMC samplers in parameter space, making them computationally feasible while maintaining the expressiveness of functional priors.
    \item We prove that the stationary measure would be the target posterior over functions under our designed functional Langevin dynamics and functional Hamiltonian dynamics.
    \item Our methods demonstrate improved predictive performance and uncertainty estimation in several benchmark tasks compared to parameter-space SGMCMC and parameter-/function-space variational inference methods.
\end{itemize}

\section{Preliminaries}
\label{sec:prelim}

\subsection{Bayesian Neural Networks}

Bayesian neural networks (BNNs) place probability distribution over network parameters rather than single fixed values used in standard neural networks. Given a dataset $\mathcal{D}=\left\{x_i, y_i\right\}_{i=1}^{N} = \left\{\mathbf{X_\mathcal{D}}, \mathbf{Y_\mathcal{D}}\right\}$ , where $x_i\in \mathcal{X} \subseteq \mathbb{R}^p$ are the training inputs and $y_i\in \mathcal{Y} \subseteq \mathbb{R}^c$ are the corresponding targets,
BNNs are stochastic neural networks characterized by random network parameters $\mathbf{w} \in \mathbb{R}^k$,  treated as a k-dimensional multivariate random variable defined on a probability space $(\Omega, \mathcal{A}, P)$. The prior distribution over parameters is denoted by $p_{0}(\mathbf{w})$, and the likelihood evaluated on the training data is represented by the conditional distribution $p(\mathbf{Y_\mathcal{D}}|\mathbf{X_\mathcal{D}}; \mathbf{w})$. The posterior over parameters can then be inferred using Bayes’ theorem, yielding the canonical form $p(\mathbf{w}|\mathcal{D})\propto p(\mathbf{Y_\mathcal{D}}|\mathbf{X_\mathcal{D}}; \mathbf{w})p_{0}(\mathbf{w})$. Furthermore, the predictive distribution for test data $\left\{x^*, y^*\right\}$ is 
given by the expectation under the posterior as $p(y^*|x^*, \mathcal{D})=\int p(y^*|x^*, \mathbf{w})p(\mathbf{w}|\mathcal{D})\mathrm{d}\mathbf{w}$. All notations used in this paper are listed in \cref{tab:note} in \cref{apd:notation}.

\subsection{Stochastic Gradient MCMC for BNNs}
 
Dynamics-based sampling methods that leverage the gradients of the target log posterior can offer an efficient exploration of the parameter space \citep{ma2015complete, izmailov2021bayesian, li2024entropy}. Furthermore, their stochastic gradient variants, such as the stochastic gradient Langevin dynamics (SGLD) \citep{welling2011bayesian} and the stochastic gradient Hamiltonian Monte Carlo (SGHMC) \citep{chen2014stochastic, baker2019control}, are capable of scaling to large datasets for practical applications. 

\paragraph{SGLD.} 
The objective of MCMC for BNNs is to generate random samples from the posterior distribution over parameters (with the unnormalized form) as $p(\mathbf{w}|\mathcal{D}) \propto \exp(-U(\mathbf{w}))$, where $U(\mathbf{w}) = -\log p(\mathbf{Y_\mathcal{D}}|\mathbf{X_\mathcal{D}}; \mathbf{w}) - \log p_{0}(\mathbf{w})$ is the potential energy function. By combining Langevin dynamics \citep{neal2011mcmc} with the stochastic Robbins-Monro optimization algorithm \citep{robbins1951stochastic}, \citet{welling2011bayesian} proposed SGLD as a scalable MCMC algorithm. The approximated discretization transition rule for posterior samples is 
\begin{equation}
    \mathbf{w}_{t+1} = \mathbf{w}_{t} - \epsilon_{t}\nabla \tilde{U}(\mathbf{w}_{t}) + \sqrt{2\epsilon_{t}} \eta_{t},
    \label{eq:psgld}
\end{equation}
where $\epsilon_{t}$ represents the step-size, and $\eta_{t}$ is a standard Gaussian noise term. The gradient $\nabla \tilde{U}(\mathbf{w})$ is computed using a mini-batch stochastic gradient for the likelihood term given by $\nabla\tilde{U}(\mathbf{w}) = -\frac{N}{n}\sum_{i=1}^{n} \nabla \log p(y_{i}|x_{i}; \mathbf{w}) - \nabla \log p(\mathbf{w})$ with $n\ll N$. As the step-size $\epsilon_{t} \rightarrow 0$, the stochastic gradient noise vanishes more quickly than the injected Gaussian noise $\eta_{t}$, which ensures the discretization error is eliminated. Hence, the Metropolis-Hastings (MH) correction step can be ignored. 

\paragraph{SGHMC.} To enable more efficient exploration of the state space, Hamiltonian dynamics \citep{duane1987hybrid, neal2012bayesian} introduces a momentum variable $\mathbf{z} \in \mathbb{R}^k$. The new target joint distribution is $p(\mathbf{w}|\mathcal{D})p(\mathbf{z})\propto \exp (- H(\mathbf{w}, \mathbf{z}))$ with corresponding energy function $H(\mathbf{w}, \mathbf{z}) = -\log p(\mathbf{Y_\mathcal{D}}|\mathbf{X_\mathcal{D}}, \mathbf{w}) - \log p(\mathbf{w}) - \log p(\mathbf{z})$, where $p(\mathbf{z})$ is commonly assumed to be a zero-mean Gaussian $\mathcal{N}(\mathbf{z}|0, M)$. The sampling rule in the full-batch Hamiltonian Monte Carlo is defined as following with an MH step \citep{neal2012bayesian, cobb2021scaling}:
\begin{equation}
\begin{aligned}
\mathbf{w}_{t+1} &= \mathbf{w}_t+\epsilon_t M^{-1} \mathbf{z}_t, \\
\mathbf{z}_{t+1} &= \mathbf{z}_t-\epsilon_t \nabla U\left(\mathbf{w}_t\right).
\label{eq:phmc}
\end{aligned}
\end{equation}
Like SGLD, SGHMC uses stochastic gradient $\nabla\tilde{U}(\mathbf{w})$ for the sampling of momentum $\mathbf{z}$,  but with an additional friction term $\epsilon_t M^{-1} \mathbf{z}_t$ and a Gaussian injected noise to ensure that the (marginal) stationary distribution of the Hamiltonian dynamics remains the target posterior distribution \citep{chen2014stochastic}. The specific update rule for the momentum $\mathbf{z}$ is given by $\mathbf{z}_{t+1} = \mathbf{z}_t-\epsilon_t \nabla \tilde{U}(\mathbf{w}_t) - \epsilon_t CM^{-1} \mathbf{z}_t + \sqrt{2C\epsilon_{t}}\eta_t$, where $C>0$ is a scalar coefficient.

\section{Functional Dynamics-based MCMC}
\label{sec:fsgmcmc}

In this section, we address the prior issues present in parameter-space MCMC by proposing two novel functional MCMC schemes based on newly designed diffusion dynamics, which allow for the incorporation of more informative functional priors. In \cref{sec:fsgld}, we introduce novel functional Langevin dynamics designed directly for the random function mapping defined by BNNs, along with a corresponding tractable stochastic gradient MCMC sampler for network parameters. We demonstrate that the stationary probability measure of this functional Langevin dynamics corresponds to the target posterior over functions. Additionally, in \cref{sec:fhmc}, we develop functional Hamiltonian dynamics also with a stochastic gradient version for BNNs. We prove that this scheme also guarantees the desired functional posterior as the stationary measure.

\subsection{Functional Langevin Dynamics for BNNs}
\label{sec:fsgld}

Dynamics-based MCMC techniques are grounded in the general framework of It{\^o} diffusion \citep{oksendal2003stochastic}.
The main idea behind parameter-space MCMC methods for BNNs is to simulate a continuous diffusion process of network parameters $\mathbf{w}$ such that its stationary distribution is the target posterior $p(\mathbf{w}|\mathcal{D})$, using appropriately designed drift term and diffusion coefficient (see \cref{apd:fb}). Naturally, to incorporate more informative functional prior into the posterior inference process, we now consider lifting the parameter-space Langevin dynamics onto function space. 

Suppose $f(\cdot; \mathbf{w}): \mathcal{X} \times \mathbb{R}^k \rightarrow \mathcal{Y}$ is the product-measurable random function mapping defined by a BNN. Assums $f(\cdot; \mathbf{w}) \in \mathbb{H}$, where $\mathbb{H}$ is an infinite-dimensional function (Polish) space with Borel $\sigma$-algebra $\mathcal{B}(\mathbb{H})$. The likelihood is defined as $p$: $\mathcal{Y} \times \mathbb{H} \rightarrow [0, \infty)$, mapping $(\mathcal{Y}, f(\mathcal{X};\mathbf{w})) \mapsto p (\mathcal{Y}|f(\mathcal{X};\mathbf{w}))$, where $\mathcal{Y} \subseteq \mathbb{R}^c$ is Borel measurable and $ p(\mathbf{Y}_{\mathcal{D}}|f(\mathbf{X}_{\mathcal{D}};\mathbf{w}))$ denotes the likelihood evaluated on the training data. Let $\mathcal{P}(\mathbb{H})$ represent the space of Borel probability measures on $\mathcal{B}(\mathbb{H})$, and assume the functional prior measure $P_{0} \in \mathcal{P}(\mathbb{H})$ with topological support $\operatorname{supp}(P_{0})$, such as the classical GP prior denoted as $P_0\sim \mathcal{GP}(\mathbf{m}, \mathbf{K})$. The posterior measure $P_{f|\mathcal{D}} \in \mathcal{P}(\mathbb{H})$, induced by the functional prior and likelihood, is defined by the Radon–Nikodym derivative as $P_{f|\mathcal{D}}(\mathrm{d}f) \propto \exp(-\Phi(f)) P_0(\mathrm{d}f)$, where $\Phi(f)$ is the negative log-likelihood as $- \log p(\mathbf{Y}_{\mathcal{D}}|f(\mathbf{X}_{\mathcal{D}};\mathbf{w}))$ \citep{matthews2016sparse, lambley2023strong}.


According to It{\^o} Lemma \citep{ito1951formula},  $f(\cdot; \mathbf{w})$ itself follows an It{\^o} diffusion \citep{brzezniak2008ito} (see \cref{apd:fb}). Thus, we propose to directly simulate the diffusion process of the random function mapping $f(\cdot; \mathbf{w})$ of BNNs, where one can incorporate informative functional prior, $P_{0}$. Suppose $\varnothing \neq E \subseteq \operatorname{supp}(P_{0})$, and let $ I_{0}: E \rightarrow \mathbb{R}$ be an \textit{Onsager–Machlup (OM) functional} for $P_{0}$, which can be heuristically interpreted as the negative logarithm of $P_{0}$ \citep{ayanbayev2021gamma} (see \cref{apd:fb}). Similar to parameter-space Langevin dynamics, the Langevin dynamics in function space can be generally represented by the stochastic differential equation (SDE): $\mathrm{d} f_{t} = - \nabla U(f_{t})\mathrm{d} t + \sqrt{2}\mathrm{d}B_t$, where $B$ is a standard Wiener process ((Brownian motion), and potential energy functional $U: E \rightarrow \mathbb{R}$ is given by $U(f) = \Phi(f) + I_{0}(f)$. This functional $U(f)$ is proven to be an OM functional for $P_{f|\mathcal{D}}$ \citep{lambley2023strong}, ensuring that the target posterior measure $P_{f|\mathcal{D}}$ as the stationary measure $\pi(f)$. However, discretizing and sampling from this diffusion process in function space is intractable due to the infinite-dimensional nature of the problem.

Since $f(\cdot; \mathbf{w})$ is completely determined by the network parameters $\mathbf{w}$ (given a fixed network architecture), we address this challenge by designing a specific functional Langevin dynamics for $f(\cdot; \mathbf{w})$, which maintains the stationary measure $P_{f|\mathcal{D}}$. We then transform this into the corresponding parameter-space dynamics for network parameters. Suppose that function space $\mathbb{H}$ is equipped with a Riemannian manifold structure \citep{wangequivalences}, we first design the following functional Riemannian Langevin dynamics for $f(\cdot; \mathbf{w})$ on $\mathbb{H}$:
\begin{equation}
    \begin{aligned}
       &\mathrm{d}f_{t}(\cdot; \mathbf{w}) = \mu(f_{t}(\cdot; \mathbf{w}))\mathrm{d}t + \sqrt{2G^{-1}(f_{t}(\cdot; \mathbf{w}))}\mathrm{d}B_t \\
        &= \Big[-(\nabla_{\mathbf{w}}f_{t})^{T}(\nabla_{\mathbf{w}}f_{t})\nabla_{f}(-\log p(\mathbf{Y}_{\mathcal{D}}|f_{t}(\mathbf{X}_{\mathcal{D}}; \mathbf{w})) + \\
        &  \quad
        I_{0}(f_{t})) + H_{\mathbf{w}}f_{t} \Big]\mathrm{d}t + \sqrt{2}(\nabla_{\mathbf{w}}f_{t})^{T}\mathrm{d}B_t,
    \end{aligned}
    \label{eq:flg}
\end{equation}
where $\nabla_{\mathbf{w}}f$ denotes the gradient of $f(\cdot; \mathbf{w})$ w.r.t. parameters $\mathbf{w}$, $H_{\mathbf{w}}f$ is the Hessian matrix, and $G^{-1}(f(\cdot; \mathbf{w}))=(\nabla_{\mathbf{w}}f)^{T}(\nabla_{\mathbf{w}}f)$. The stationary probability measure of this dynamics is guaranteed to be $P_{f|\mathcal{D}}$ as stated in the following proposition.

\begin{proposition}
    The stationary probability measure of the functional Langevin dynamics defined in \cref{eq:flg} is the target posterior over functions $P_{f|\mathcal{D}}$. 
    \label{prop:flg}
\end{proposition}

\begin{proof}
The proof is given in \cref{proof}. 
\end{proof}

To establish a tractable functional MCMC scheme, we then transform the proposed functional Riemannian Langevin dynamics for $f(\cdot; \mathbf{w})$ into the corresponding Langevin dynamics for the network parameters $\mathbf{w}$. Note that the differential of the function mapping $f(\cdot; \mathbf{w})$ is given by the It{\^o} Lemma \citep{ito1951formula} as
\begin{equation}
    \begin{aligned}
       &\mathrm{d}f_{t}(\cdot; \mathbf{w}) = (\nabla_{\mathbf{w}}f)^{T}\mathrm{d}\mathbf{w}_{t} + \frac{1}{2}(\mathrm{d}\mathbf{w}_{t})^{T}(H_{\mathbf{w}}f)\mathrm{d}\mathbf{w}_{t} \\
       &= \left((\nabla_{\mathbf{w}}f)^{T}\mu(\mathbf{w}_t) + \frac{1}{2}Tr[\sigma(\mathbf{w}_t)^{T}(H_{\mathbf{w}}f)\sigma(\mathbf{w}_t)]\right)\mathrm{d}t \\
       & + (\nabla_{\mathbf{w}}f)^{T}\sigma(\mathbf{w}_t)\mathrm{d}B_t,
      \label{eq:itof}
    \end{aligned}
\end{equation}
where $\mathrm{d}\mathbf{w}_{t} =\mu(\mathbf{w}_{t}) \mathrm{d}t + \sigma(\mathbf{w}_{t})\mathrm{d}B_{t}$ is the general diffusion form for $\mathbf{w}$. By combining our functional Langevin dynamics defined in \cref{eq:flg} with this differential relationship between $f(\cdot; \mathbf{w})$ and $\mathbf{w}$, it is straightforward to reversely derive the corresponding diffusion process for $\mathbf{w}$ in parameter spaces as follows:
\begin{equation}
\begin{aligned}
     & \mathrm{d}\mathbf{w}_t = \mu(\mathbf{w}_{t}) \mathrm{d}t + \sigma(\mathbf{w}_{t})\mathrm{d}B_t \\
      &= -\nabla_{\mathbf{w}}f_{t} \left[\nabla_{f}(-\log p(\mathbf{Y}_{\mathcal{D}}|f_{t}(\mathbf{X}_{\mathcal{D}}; \mathbf{w})) + I_{0}(f_{t})) \right]\mathrm{d}t \\
      &  + \sqrt{2}\mathrm{d}B_t,
\end{aligned}
\label{eq:fld}
\end{equation}
where the drift term $\mu(\mathbf{w}) = -\nabla_{\mathbf{w}}U(f)$
, and the diffusion coefficient $\sigma(\mathbf{w}_{t}) = \sqrt{2}\cdot\mathbf{I}_{k}$.
Then, by discretizing and sampling from this corresponding diffusion process for the network parameters, we can obtain function samples from the target posterior measure $P_{f|\mathcal{D}}$ in a tractable manner.

In practice, the stochastic gradient variant for the drift term of SDE in \cref{eq:fld} is denoted as $ -\nabla_{\mathbf{w}}\tilde{U}(f)$, which utilizes the minibatch computation for the likelihood term. To estimate the gradient of the prior measure, we assume an approximate GP prior for BNNs. This is motivated by the fact that, in the limit of infinite width, the prior measure of BNNs converges to a GP with the Neural Network Gaussian Process (NNGP) kernel \citep{neal1995bayesian, lee2017deep, matthews2018gaussian, lee2019wide}. Due to the infinite-dimensional nature of the functional distribution, we solve the gradient of functional prior on finite measurement points $\mathbf{X}_{\mathcal{M}} \stackrel{\text { def }}{=}\left[\mathbf{x}_1, \ldots, \mathbf{x}_M\right]^{\mathrm{T}}$ as $\nabla_{f^{\mathbf{X}_{\mathcal{M}}}} \log P_{0}(f^{\mathbf{X}_{\mathcal{M}}})$, where $f^{\mathbf{X}_{\mathcal{M}}}$ are corresponding function values evaluated at $\mathbf{X}_{\mathcal{M}}$. The prior $P_{0}(f^{\mathbf{X}_{\mathcal{M}}})$ reduces to a multivariate Gaussian, allowing for an analytical solution for the gradient. The specific discretization update rule for network parameters samples under functional stochastic gradient Langevin dynamics (fSGLD) is given by
\begin{equation}
    \begin{aligned}
        \mathbf{w}_{t+1} 
         =& \mathbf{w}_{t} - \epsilon_{t}\nabla_{\mathbf{w}}f_{t} \Big [-\frac{N}{n}\sum_{i=1}^{n} \nabla_{f} \log p(y_{i}|f_{t}(x_{i}; \mathbf{w}_{t})) - \\ &\hspace{2.2cm}\nabla_{f^{\mathbf{X}_{\mathcal{M}}}} \log P_{0}(f_{t}^{\mathbf{X}_{\mathcal{M}}}) \Big] + \sqrt{2\epsilon_{t}} \eta_{t}.
    \end{aligned}
    \label{eq:fsgld}
\end{equation}
Like the standard SGLD, with the discretization step-size $\epsilon_{t}$ decays to zero, the stationary measure of fSGLD continues to be the target posterior $P_{f|\mathcal{D}}$ over functions. 

Note that in contrast to the naive parameter-space SGLD in \cref{eq:psgld}, which is driven solely by limited parameter information, the discretization update rule for samples of network parameters in \cref{eq:fsgld} under our fSGLD scheme incorporates information about the function mapping. This additional functional information plays a crucial role in achieving more accurate posterior inference. The pseudocode for fSGLD is shown in \cref{alg:fsgld} in \cref{apdix}.

\subsection{Functional Hamiltonian dynamics for BNNs}
\label{sec:fhmc}
For the naive Hamiltonian dynamics, $X=(\mathbf{w}, \mathbf{z})$ represents the augmentation of model parameters. We now let $X=(f, g)$ and define the Hamiltonian dynamics in function space as $H(f, g) = U(f) - \log p (g)$, where $p (g)$ is the functional auxiliary probability measure induced by the distribution of auxiliary variable $\mathbf{z}$. We then design the functional Hamiltonian dynamics for $f(\cdot; \mathbf{w})$ and $g(\cdot; \mathbf{z})$ as

\begin{equation}
    \begin{aligned}
       & \mathrm{d}f_{t}(\cdot; \mathbf{w}) = -(\nabla_{\mathbf{w}}f_{t})^{T} \nabla_{\mathbf{z}}g_{t} \nabla_{g} \log p(g_{t})\mathrm{d} t,\\
       & \mathrm{d}g_{t}(\cdot; \mathbf{z}) = - (\nabla_{\mathbf{z}}g_{t})^{T} \nabla_{\mathbf{w}}f_{t} \nabla_{f} U(f_{t}) \mathrm{d} t.
    \end{aligned}
    \label{eq:fhmc}
\end{equation}

\begin{table*}[!t]
\caption{Drift term and diffusion coefficient for naive SGMCMC and functional SGMCMC algorithms.}
\label{pfmcmc}
\begin{center}
\begin{small}
\footnotesize
\begin{sc}
\begin{tabular}{lcc}
\toprule
Dynamics &  $\mu(X)$ & $\sigma(X)$  \\
\midrule
        SGLD  & $-\nabla\tilde{U}(\mathbf{w})$ & $\sqrt{2}\cdot\mathbf{I}_d$ \\

        fSGLD  & $-(\nabla_{\mathbf{w}}f)^{T}(\nabla_{\mathbf{w}}f)\nabla_{f}\tilde{U}(f) + H_{\mathbf{w}}f$ & $\sqrt{2}(\nabla_{\mathbf{w}}f)^{T}$ \\
        
        SGHMC  & $-\left(\begin{array}{cc} \mathbf{0} & -\mathbf{I}_d \\ \mathbf{I}_d & C \end{array}\right) \left(\begin{array}{cc} \nabla\tilde{U}(\mathbf{w}) \\ M^{-1}\mathbf{z}\end{array}\right)$ & $\left(\begin{array}{cc} \mathbf{0} & \mathbf{0} \\ \mathbf{0} & \sqrt{2C} \end{array}\right)$\\

        fSGHMC  
        & $-\left(\begin{array}{cc} \mathbf{0} & -(\nabla_{\mathbf{w}}f)^{T} \nabla_{\mathbf{z}} g \\ (\nabla_{\mathbf{z}} g)^{T} \nabla_{\mathbf{w}}f & C(\nabla_{\mathbf{z}} g)^{T} \nabla_{\mathbf{z}} g \end{array}\right) \left(\begin{array}{c} \nabla_{f} \tilde{U}(f) \\
       -\nabla_{g} \log p(g)
       \end{array}\right)$
        & $\left(\begin{array}{cc} \mathbf{0} & \mathbf{0} \\ \mathbf{0} & \sqrt{2C}(\nabla_{\mathbf{z}} g)^{T} \end{array}\right)$\\
\bottomrule
\end{tabular}
\end{sc}
\end{small}
\end{center}
\end{table*}

\begin{proposition}
    The (marginal) stationary probability measure of the functional Hamiltonian dynamics defined in \cref{eq:fhmc} is the target posterior over functions $P_{f|\mathcal{D}}$. 
    \label{prop:fhmc}
\end{proposition}

\begin{proof}
The proof is given in \cref{proof}. 
\end{proof}

Similar to the fSGLD, we also transform the above function-space Hamiltonian dynamics to its corresponding parameter-space dynamics of $\mathbf{w}$ and $\mathbf{z}$ as follows: 
\begin{equation}
    \begin{aligned}
        \mathrm{d} \mathbf{w}_{t} &= \frac{\partial H}{\partial \mathbf{z}} = -\nabla_{\mathbf{z}} g_{t} \cdot \nabla_{g} \log p(g_{t}) \mathrm{d}t, \\
        \mathrm{d} \mathbf{z}_{t} &= -\frac{\partial H}{\partial \mathbf{w}} = -\nabla_{\mathbf{w}} f_{t} \cdot \nabla_{f} U(f_{t}) \mathrm{d}t,
    \end{aligned}
\end{equation}
where the random diffusion term is 0 for the full-batch version. 

For the stochastic gradient version, the discretization update rule for samples of $\mathbf{w}$ and $\mathbf{z}$ under the  functional stochastic gradient Hamiltonian dynamics (fSGHMC) is as follows:
\begin{equation}
\begin{aligned}
\mathbf{w}_{t+1} =&~ \mathbf{w}_t - \epsilon_t\nabla_{\mathbf{z}} g_{t} \cdot \nabla_{g} \log p(g_{t}) \\
\mathbf{z}_{t+1} =&~ \mathbf{z}_t-\epsilon_t\nabla_{\mathbf{w}} f_{t} \cdot \nabla_{f} \tilde{U}(f_{t}) 
\\
&\hspace{0.5cm}+ \epsilon_{t}C\nabla_{\mathbf{z}} g_{t} \cdot \nabla_{g} \log p(g_{t})
+ \sqrt{2C\epsilon_{t}} \eta_t.
\end{aligned}
\label{eq:fsghmc}
\end{equation}
The stationary probability measure of such functional stochastic gradient Hamiltonian can still be guaranteed as the target functional posterior $P_{f|\mathcal{D}}$ (see \cref{proof}). Like the discussion in the last section, the diffusion process for $\mathbf{w}$ and $\mathbf{z}$ under fSGHMC scheme in \cref{eq:fsghmc} incorporates abundant functional information is obviously different from the naive parameter-space SGHMC in \cref{eq:phmc}.
The pseudocode for fSGHMC is shown in \cref{alg:fsghmc} in \cref{apdix}.


        


For our functional MCMC schemes, the predictive distribution is obtained from the following integration process:
\begin{equation}
\begin{aligned}
     p(y^*| x^*, \mathcal{D}) & = \int p(y^*|f(x^*))P_{f|\mathcal{D}}\mathrm{d}f \\
    & \approx \frac{1}{S} \sum_{j=1}^{S} p(y^*|f(x^*; \mathbf{w}^{(j)})),
\end{aligned}
\end{equation}
where function draws of posterior over functions are generated from corresponding samples of network parameters $\mathbf{w}^{(j)} (j=1, 2, ..., S)$ using fSGLD or fSGHMC.

The comparisons of the definitions for the drift term and the diffusion coefficient of popular exiting SGMCMC algorithms and our functional fSGLD and fSGHMC are summarized in \cref{pfmcmc}.

\section{Related Work}
\label{sec:relawok}
\textbf{Functional Variational Inference for BNNs}.
To specify meaningful functional priors like Gaussian processes, \citet{sun2019functional} proposed a functional evidence lower bound (ELBO) to optimize the Kullback–Leibler (KL) divergence between variational posterior and the true posterior in function space explicitly. They solved the KL divergence between infinite-dimensional stochastic processes on finite marginal measurement sets. Due to the analytical intractability of the supremum over marginal KL divergences in functional ELBO, \citet{rudner2022tractable} proposed to approximate the posterior and prior over functions of BNNs as Gaussian distributions through local linearization and derived a more tractable functional variational objective for BNNs. Considering the potential weaknesses of KL divergence for stochastic processes \citet{burt2020understanding}, \citet{tran2022all} proposed to match
a BNN prior to a GP prior by minimizing the 1-Wasserstein distance \citep{kantorovich1960mathematical} to obtain a more interpretable functional prior.

\textbf{Markov Chain Monte Carlo for BNNs}. 
\citet{papamarkou2022challenges} summarized the challenges in applying MCMC approaches to posterior inference for BNNs in four aspects: computational cost, model structure, weight symmetries, and prior specification. MCMC methods had limited success in being broadly adopted for posterior inference in large modern BNNs until the introduction of scalable Monte Carlo algorithms, such as the stochastic gradient MCMC  \citep{welling2011bayesian, chen2014stochastic, ahn2014distributed, garriga2021exact}. \citet{ma2015complete} designed a unifying framework that casts all existing dynamics-based samplers and their minibatch variants into it. Regarding the multimodal properties of posterior distributions, \citet{Zhang2020Cyclical} proposed a cyclical step-size schedule in stochastic gradient MCMC for BNNs to improve the sampling efficiency. On the other hand, \citet{cobb2021scaling} emphasized the limitations of stochastic gradients in HMC and proposed a scaling full-batch HMC for BNNs using a symmetric splitting integration scheme.
However, all these MCMC methods are performed in parameter space.

    

\section{Experiments}
\label{sec:exp}

We evaluate the predictive performance and uncertainty quantification of our functional stochastic gradient MCMC schemes on several benchmark tasks, including a synthetic extrapolation example, multivariate regressions on UCI datasets, image classification tasks, and contextual bandits. We compare fSGLD and fSGHMC with naive SGLD, SGHMC, and competing functional variational inference methods to illustrate the sampling efficiency of our functional stochastic gradient MCMC schemes.

\subsection{Extrapolation on Synthetic Data}
\label{sec:1d}
\begin{figure*}[t!]
    \centering
  {
    \subfigure[fSGLD]{\label{fig:fsgld}%
       \includegraphics[width=0.3\linewidth]{./toy/fsgld}}%
    \subfigure[fSGHMC]{\label{fig:fhmc}%
      \includegraphics[width=0.3\linewidth]{./toy/fhmc2}}%
    \subfigure[FBNN]{\label{fig:fbnn}%
      \includegraphics[width=0.3\linewidth]{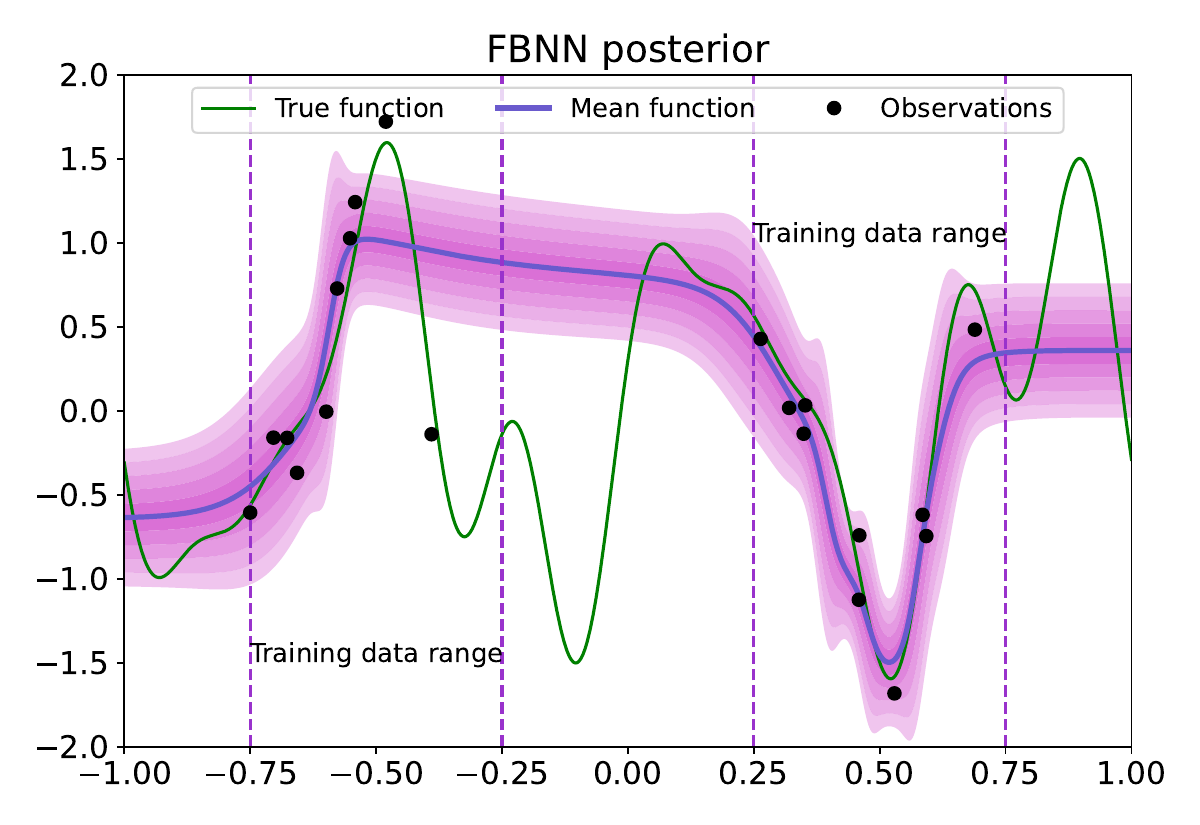}}%
    
    \subfigure[SGLD]{\label{fig:psgld}%
      \includegraphics[width=0.3\linewidth]{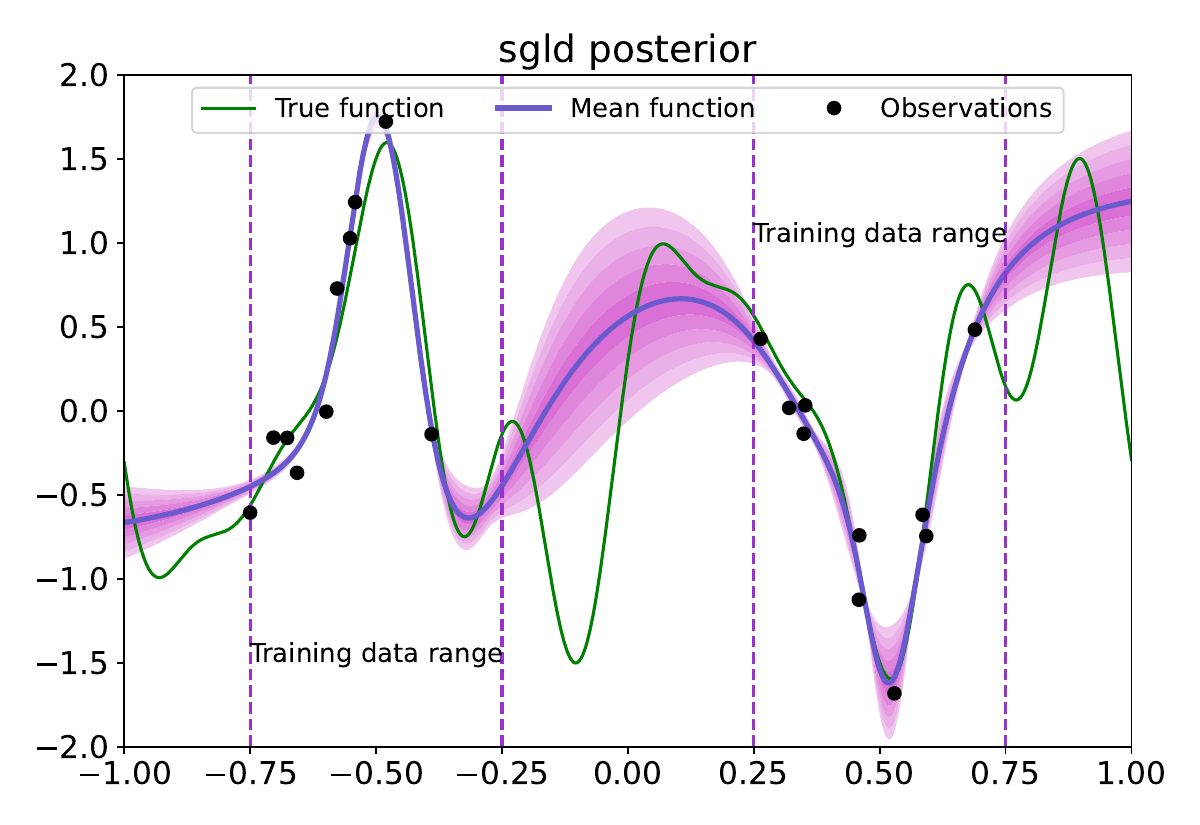}}%
    \subfigure[SGHMC]{\label{fig:phmc}%
      \includegraphics[width=0.3\linewidth]{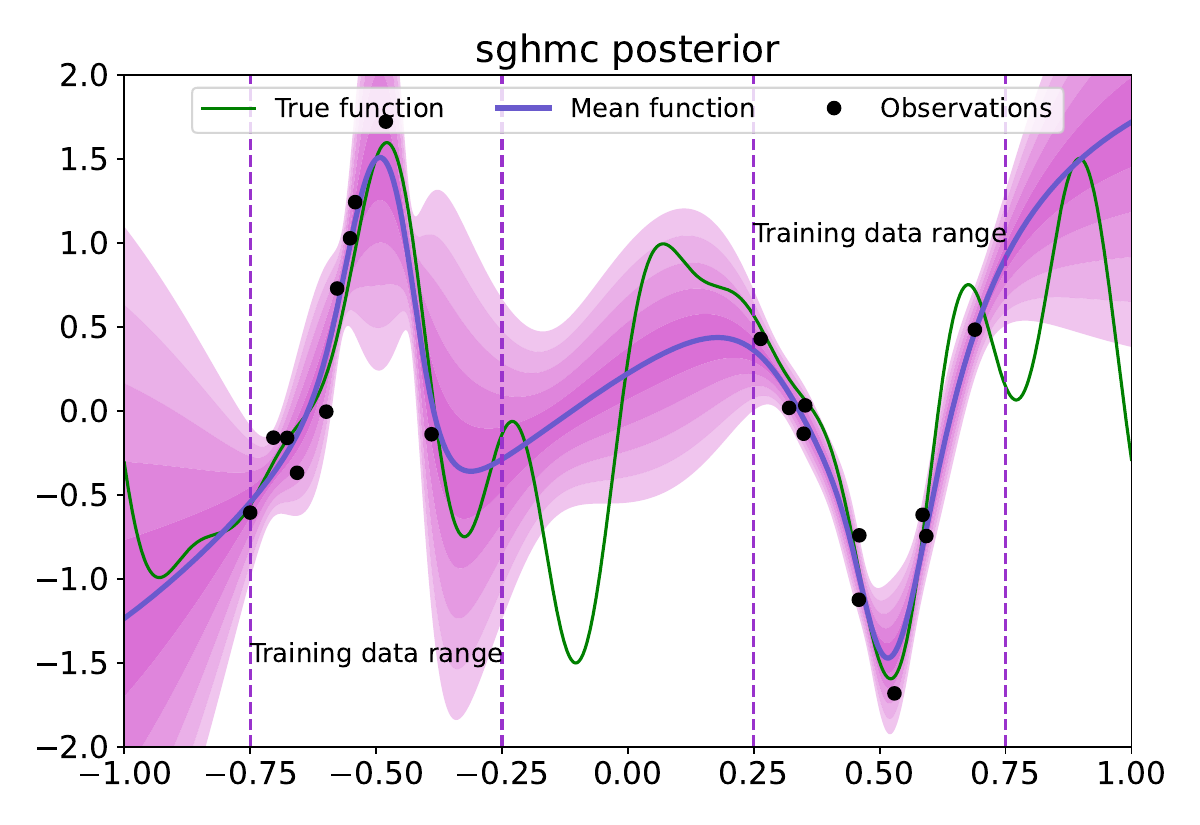}}%
    \subfigure[IFBNN]{\label{fig:ifbnn}%
      \includegraphics[width=0.3\linewidth]{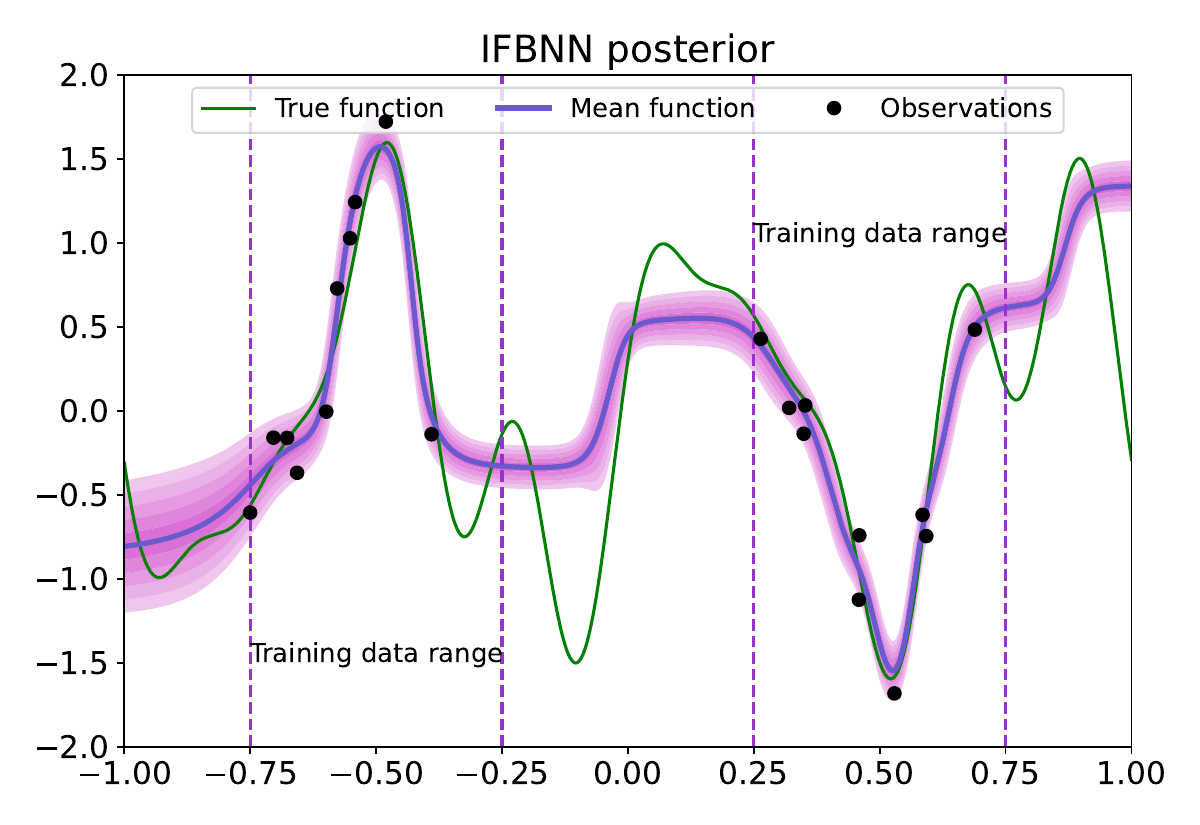}}%
    \caption{1-D extrapolation example. The green line is the ground true function, and the blue lines correspond to the mean of samples from posterior predictions. Black dots denote 20 training points; shadow areas represent the predictive standard deviations. For more details, see \cref{apd:expset}.}
    \label{fig:toy}}
\end{figure*}

To evaluate the fitting ability and uncertainty estimation, we first consider a 1-D oscillation curve extrapolation on a synthetic dataset. For input, we randomly sampled half of 20 observation points from Uniform$(-0.75,-0.25)$, and the other half are from Uniform$(0.25,0.75)$. The output is modeled through the polynomial function:  $y = \sin(3\pi x) + 0.3 \cos(9\pi x) + 0.5 \sin(7\pi x) + \epsilon$ with noise $\epsilon \sim \mathcal{N}(0,0.5^2)$. We compared our fSGLD and fSGHMC with naive SGLD, SGHMC, and two representative functional variational methods: FBNN \citep{sun2019functional} and IFBNN \citep{wu2023indirect}. The model is a fully-connected neural network with two hidden layers. For our functional fSGLD, fSGHMC, and the two functional variational inference methods, we use the same functional GP prior with the RBF kernel pre-trained on the input dataset. There are 40 measurement points for the approximate estimation of the gradient of log functional prior in fSGLD and fSGHMC, which are randomly sampled from training data and an additional 40 inducing points from Uniform $(-1, 1)$. For all sampling methods, we use 2000 burn-in iterations and 8000 iterations for 80 samples (to reduce correlations between samples, we draw separated samples every 100 epochs). Functional variational inference methods are trained for 10000 epochs for fair comparison. Results are shown in \cref{fig:toy}. \cref{fig:fsgld} and \cref{fig:fhmc} show that fSGLD and fSGHMC can recover the key characteristic of the curve in the observation range and provide reasonable uncertainty estimations in the unseen region. fSGHMC performs slightly better than fSGLD. In contrast, for naive SGMCMC methods, \cref{fig:psgld} illustrates that SGLD underestimates the uncertainty in non-observation areas, especially evident in the leftmost non-observed $[-1, -0.75]$ interval. Meanwhile, \cref{fig:phmc} shows that SGHMC suffers from some unreasonable manifestations of the uncertainty estimation, such as the undeserved uncertainty expansion in the interval $[-0.4, -0.25]$ of the input data range. On the other hand, the rightmost column shows results from two functional variational methods. We found that FBNN in \cref{fig:fbnn} is unable to fit the target curve in the training data range $[-0.75, 0.25]$ and performs poorly for uncertainty quantification. \cref{fig:ifbnn} shows that IFBNN also severely underestimates the predictive uncertainty, which is consistent with the findings of \citet{ormerod2010explaining} and \citet{zhang2018advances} that variational approximation typically underestimates the posterior variance. More analysis of mixing rate and computational complexity are shown in \cref{apd:mix} and \cref{apd:compu}, respectively. \cref{apd:abl} presents a wide ablation study about the sample size, measurement points and functional prior.

\subsection{UCI Regression}
\label{sec:uci}
In this experiment, we evaluate our methods on multivariate regression tasks across seven benchmark UCI datasets: \textit{Yacht, Boston, Concrete, Energy, Wine, Kin8nm and Protein}. We use a two-hidden-layer fully connected neural network. For all sampling methods, we run 500 iterations for the burn-in stage and collect 15 samples. Performance is measured via average root mean square error (RMSE) and the test negative log-likelihood (NLL).  For each dataset, we construct 10 random 90-to-10 train-test splits and report the mean and standard deviation of RMSE and NLL over these splits in \cref{tab:rmse} and \cref{tab:nll}, respectively. fSGHMC outperforms all baselines for RMSE on all datasets, and fSGLD also achieves better accuracy values than naive SGMCMC methods and functional variational inference, demonstrating superior performance of our functional SGMCMC in predictive accuracy. Moreover, fSGHMC and fSGLD obtain better NLL results on most of the datasets (5/7), showing that our methods are also very competitive in uncertainty quantification.

\begin{table*}[!t]
\small
\centering
    \caption{The table shows the results of average RMSE for multivariate regression on UCI datasets. We split each dataset randomly into 90\% training data and 10\% test data, and this process is repeated 10 times to ensure validity. Bold indicates statistically significant best results ($p<0.01$ with t-test).}
    \label{tab:rmse}
    \begin{sc}
    \begin{tabular}{lcccccccc}
     \hline & \multicolumn{6}{c}{\text { RMSE }} &  \\
    \cline { 2 - 7 } & \text { SGLD } & \text { fSGLD } & \text { SGHMC } & \text { fSGHMC } & \text { FBNN } & \text { IFBNN } \\
    \hline 
    \text{Yacht} & 
    1.09 $\pm$ 0.10 & 0.41 $\pm$ 0.10 & 1.19 $\pm$ 0.10 & \textbf{0.25} $\pm$ \textbf{0.13} & 1.52 $\pm$  0.08  & 1.24 $\pm$ 0.10 \\
    \text{Boston} & 
    1.23 $\pm$ 0.06 & 0.36 $\pm$ 0.09 & 1.31 $\pm$ 0.06 & \textbf{0.24} $\pm$ \textbf{0.10} & 1.68 $\pm$  0.12 & 1.44 $\pm$ 0.09 
    \\
    \text{Concrete} & 
    1.10 $\pm$ 0.07 & 0.45 $\pm$ 0.04 & 1.15 $\pm$ 0.06 & \textbf{0.28} $\pm$ \textbf{0.05} & 1.27 $\pm$ 0.05 & 1.07 $\pm$ 0.07 
    \\
    \text{Energy} & 
    1.04 $\pm$ 0.06 & 0.24 $\pm$ 0.03 & 1.07 $\pm$ 0.08 & \textbf{0.18} $\pm$ \textbf{0.03} & 1.35 $\pm$ 0.06 & 1.19 $\pm$ 0.05 
    \\
    \text{Wine} & 
     1.08 $\pm$ 0.09 & 0.71 $\pm$ 0.05 & 1.14 $\pm$ 0.09 & \textbf{0.62} $\pm$ \textbf{0.04} & 1.53 $\pm$ 0.05 & 1.21 $\pm$ 0.05 
    \\
    \text{Kin8nm} & 
     1.20 $\pm$ 0.02 & 0.74 $\pm$ 0.02 & 1.26 $\pm$ 0.03 & \textbf{0.37} $\pm$ \textbf{0.01} & 1.45 $\pm$ 0.07 & 1.12 $\pm$ 0.02 
    \\
     \text{Protein} & 
     1.12 $\pm$ 0.01 & 0.84 $\pm$ 0.01 & 1.21 $\pm$ 0.01 & \textbf{0.83} $\pm$ \textbf{0.01} & 1.50 $\pm$ 0.03 & 1.16 $\pm$ 0.01  
     \\
    \hline 
    \end{tabular}
    \end{sc}
\end{table*}

\begin{table*}[!t]
\small
\centering
    \caption{The table shows the results of average NLL for multivariate regression on UCI datasets. We split each dataset randomly into 90\% training data and 10\% test data, and this process is repeated 10 times to ensure validity. Bold indicates statistically significant best results ($p<0.01$ with t-test).}
    \label{tab:nll}
    \begin{sc}
    \begin{tabular}{lcccccccc}
     \hline & \multicolumn{6}{c}{\text { NLL }} &  \\
    \cline { 2 - 7 } & \text { SGLD } & \text { fSGLD } & \text { SGHMC } & \text { fSGHMC } & \text { FBNN } & \text { IFBNN } \\
    \hline 
    \text{Yacht} & 
    -0.37 $\pm$ 0.08 & -2.46 $\pm$  0.28 & -2.06 $\pm$ 0.10 & \textbf{-3.37} $\pm$ \textbf{0.69} & -0.77 $\pm$ 0.86 & -1.25 $\pm$ 1.21
    \\
    \text{Boston} & 
    -0.60 $\pm$ 0.11 & -2.20 $\pm$  0.20 & -2.25 $\pm$ 0.08 & \textbf{-3.26} $\pm$\textbf{0.33} & -1.19 $\pm$ 0.76 & 0.32 $\pm$ 0.30
    \\
    \text{Concrete} & 
    -0.54 $\pm$ 0.05 & -1.63 $\pm$  0.14 & -2.51 $\pm$ 0.08 & \textbf{-2.74} $\pm$\textbf{0.18} & -1.00 $\pm$ 0.52 & -0.39 $\pm$ 0.33
    \\
    \text{Energy} & 
   -0.58 $\pm$ 0.09 & -2.25 $\pm$ 0.25 & -2.39 $\pm$ 0.09 & \textbf{-4.29} $\pm$\textbf{0.23} & -2.14 $\pm$ 0.47 & -1.78 $\pm$ 0.36
    \\
    \text{Wine} & 
    -0.90 $\pm$ 0.12 & \textbf{-2.53} $\pm$ \textbf{0.15} & -2.50 $\pm$ 0.10 & -2.34 $\pm$ 0.25 & 0.52 $\pm$ 0.14 & 0.26 $\pm$ 0.15
    \\
    \text{Kin8nm} & 
     -0.60 $\pm$ 0.04 & -2.24 $\pm$ 0.09 & -2.29 $\pm$ 0.04 & -2.00 $\pm$0.12 & \textbf{-2.45} $\pm$ \textbf{0.62} & -1.01 $\pm$ 0.14
    \\
     \text{Protein} & 
     -0.60 $\pm$ 0.01 & -2.10 $\pm$ 0.08 & \textbf{-2.20} $\pm$ \textbf{0.01} & -2.02 $\pm$0.06 & -1.49 $\pm$ 0.24 & -2.13 $\pm$ 0.34
     \\
    \hline 
    \end{tabular}
    \end{sc}
\end{table*}

\begin{table*}[!t]
\small
\centering
    \caption{Image classification and OOD detection performance.}
    \label{tab:classification}
    \begin{sc}
    \begin{tabular}{lcccccc}
    \hline & \multicolumn{2}{c}{\text { MNIST }} & \multicolumn{2}{c}{\text { FMNIST }}& \multicolumn{2}{c}{\text { CIFAR10 }} \\
    \cline { 2 - 3 } \cline { 4 - 5 } \cline { 6 - 7 } \text { Model } & \text { Test Error } & \text { AUC } & \text { Test Error } & \text { AUC } & \text { Test Error } & \text { AUC } \\
    \hline 
    SGLD & 2.73 $\pm$ 0.00 & 0.859 $\pm$ 0.06 & 13.92 $\pm$ 0.01 & 0.633 $\pm$ 0.13 & 49.56 $\pm$ 0.03 & 0.669 $\pm$ 0.04
    \\
    fSGLD & 1.89 $\pm$ 0.00 & \textbf{0.960} $\pm$ \textbf{0.02} & \textbf{11.66} $\pm$ \textbf{0.01} & 0.849 $\pm$ 0.03 &  33.57 $\pm$ 0.02 & 0.672 $\pm$ 0.03
    \\
    SGHMC & 2.97 $\pm$ 0.00 & 0.853 $\pm$ 0.05 & 13.72 $\pm$ 0.01 & 0.589 $\pm$ 0.07 & 48.13 $\pm$ 0.04 & 0.639 $\pm$ 0.06 
    \\ 
    fSGHMC & \textbf{1.64} $\pm$ \textbf{0.00} & 0.931 $\pm$ 0.02 & 11.91 $\pm$ 0.00 & \textbf{0.880} $\pm$ \textbf{0.03} & \textbf{31.99} $\pm$ \textbf{0.01} & \textbf{0.717} $\pm$ \textbf{0.03}
    \\
    FBNN & 3.99 $\pm$ 0.00 & 0.801 $\pm$ 0.03 & 14.36 $\pm$ 0.00 & 0.814 $\pm$ 0.02 & 53.71 $\pm$ 0.01 & 0.612 $\pm$ 0.03 \\
    
    IFBNN & 3.64 $\pm$ 0.00 & 0.949 $\pm$ 0.03 & 14.15 $\pm$ 0.00 & 0.838 $\pm$ 0.01 & 53.38 $\pm$ 0.01 & 0.616 $\pm$ 0.03 \\
    \hline 
    \end{tabular}
    \end{sc}
\end{table*}

\begin{figure*}[!t]
    \centering
  {
    \subfigure[p = 0.4]{\label{fig:context_0.4}%
       \includegraphics[width=0.33\linewidth]{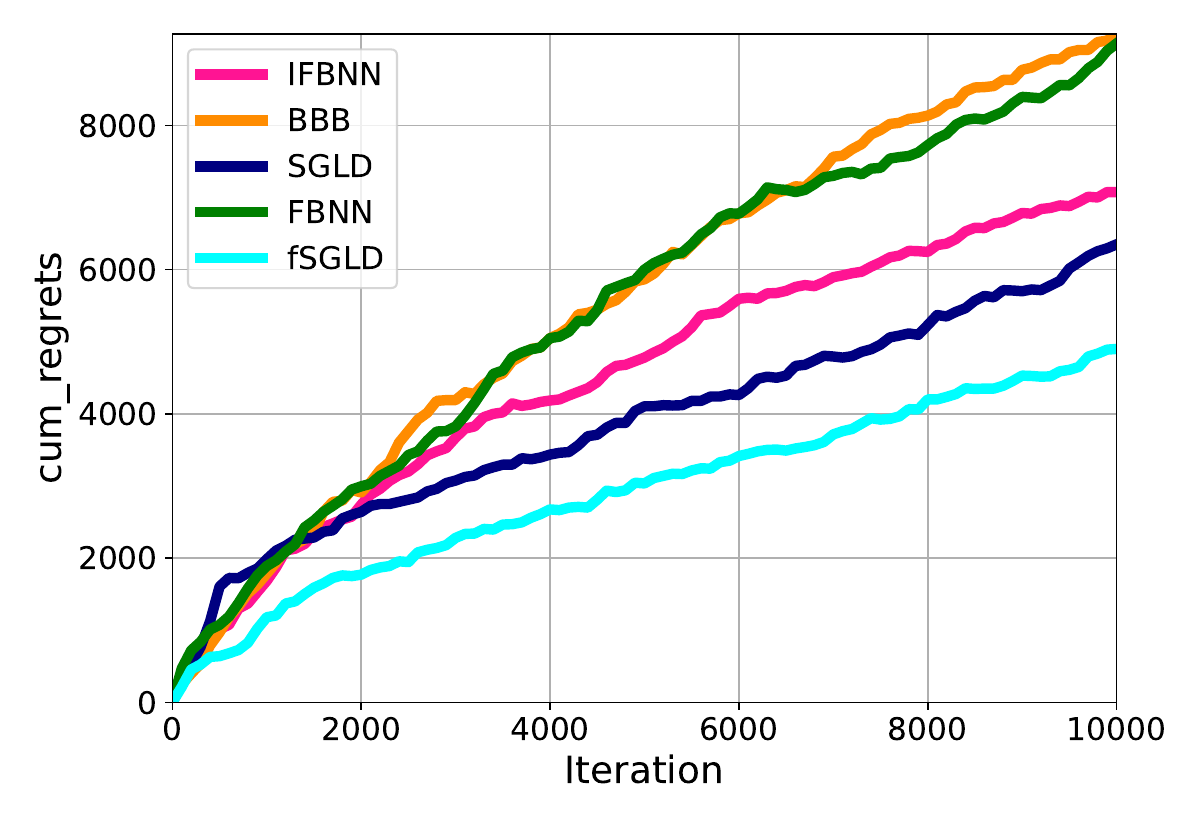}}%
    \subfigure[p = 0.5]{\label{fig:context_0.5}%
      \includegraphics[width=0.33\linewidth]{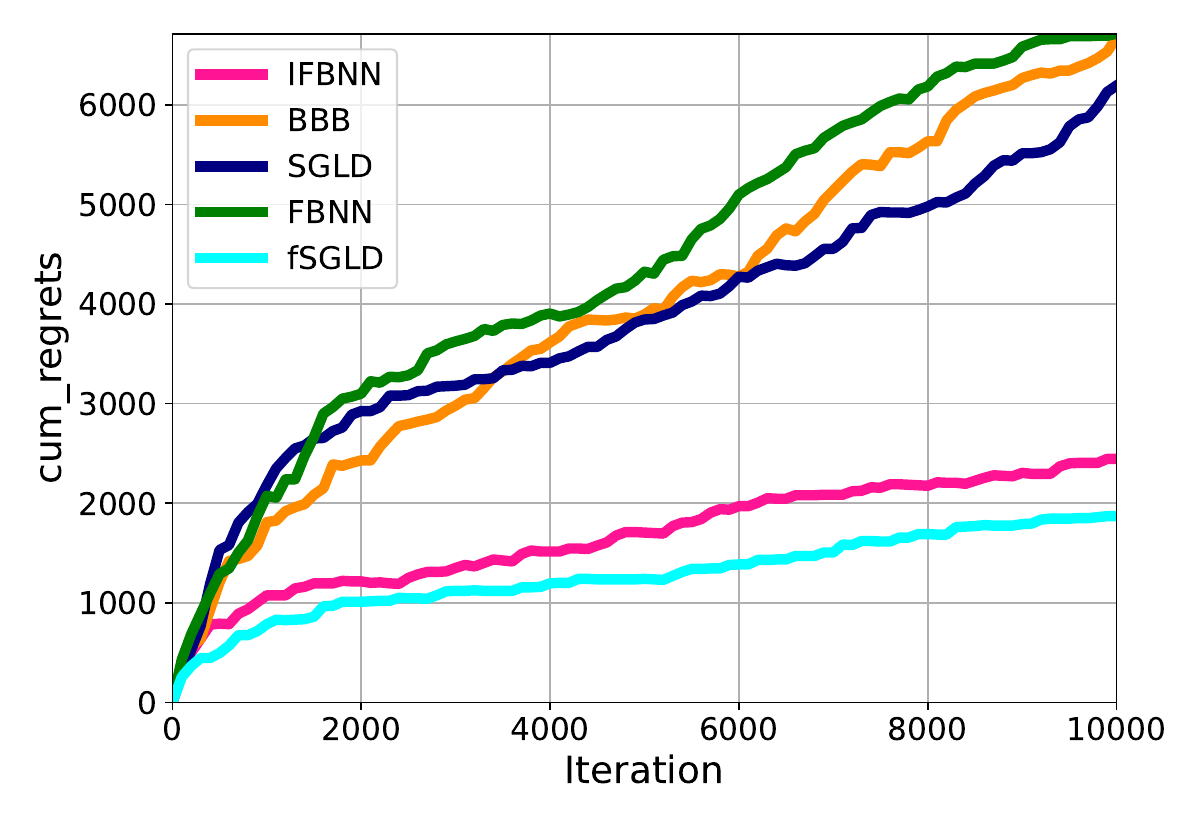}}%
    \subfigure[p = 0.6]{\label{fig:context_0.6}%
      \includegraphics[width=0.33\linewidth]{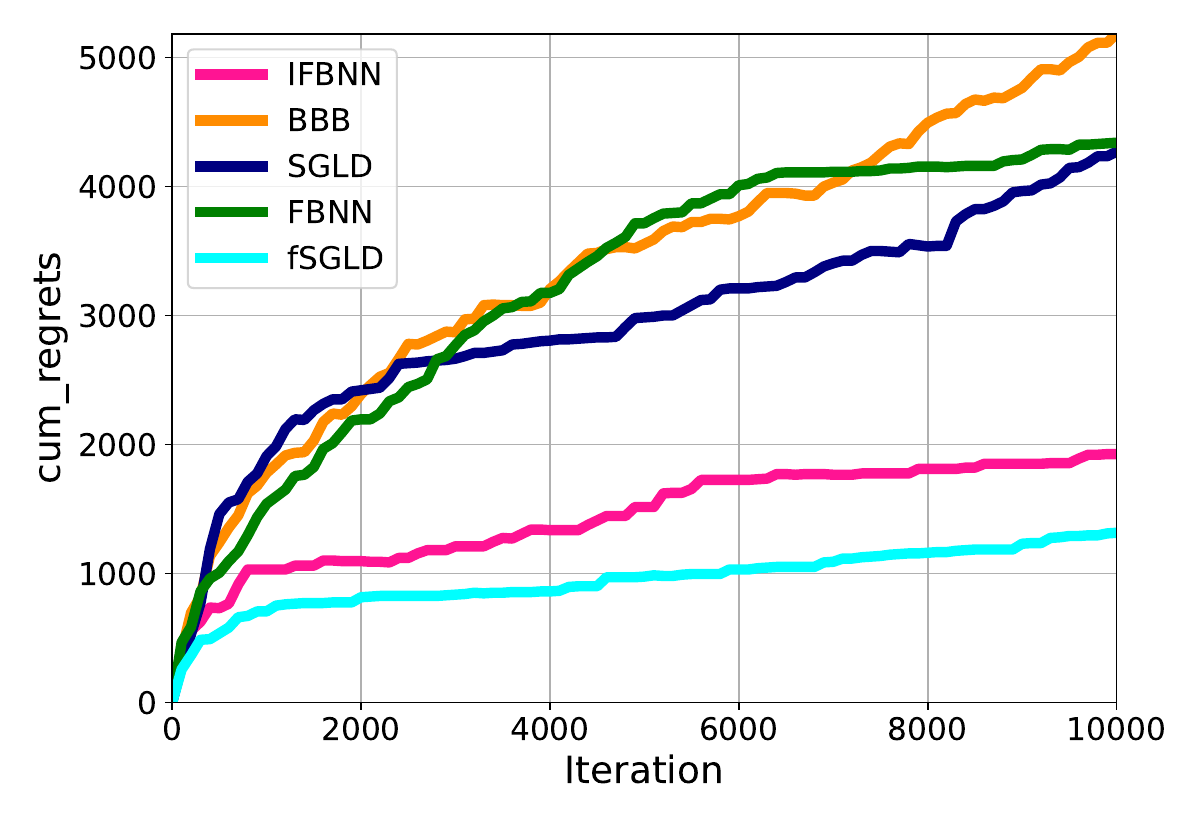}}%
      \caption{Comparisons of cumulative regrets of fSGLD, SGLD, FBNN, IFBNN, BBB for contextual bandit task on the Mushroom dataset. Lower represents better performance.}
    \label{fig:bandits}}
\end{figure*}

\subsection{Imagine Classification and OOD Detection}

In this section, we investigate the scalability of our methods on image classification tasks with high-dimensional inputs. We evaluate the in-distribution predictive performance and out-of-distribution (OOD) detection ability on MNIST \citep{lecun2010mnist}, FashionMNIST\citep{xiao2017fashion} and CIFAR-10\citep{krizhevsky2009learning}. We use ResNet-18 architecture \citep{he2016deep} for all methods on these three datasets, and the batch size is 125. For all sampling methods, we run for 60 burn-in iterations and collect 10 samples on the MNIST dataset, 100 burn-in iterations and collect 10 samples on FashionMNIST, 80 burn-in iterations and draw 10 samples on CIFAR10. We randomly sample 40 points from training data in every iteration as the measurement points for the gradient estimation of log functional prior distribution in fSGLD and fSGHMC. We report the test classification errors for predictive performance and the area under the curve (AUC) of OOD detection pairs FashionMNIST/MNIST, MNIST/FashionMNIST and CIFAR10/SVNH based on predictive entropies in Table \ref{tab:classification}. Our fSGLD and fSGHMC outperform all other methods for both predictive accuracy and OOD detection. 


\subsection{Contextual Bandits}

In many downstream tasks, such as online sequential decision-making scenarios in contextual bandits problems, reliable uncertainty estimation is important to guide the exploration-exploitation process. In such problems, an agent interacts with an unknown environment repeatedly and chooses to take an optimal action in each round of interaction. Thompson sampling \citep{thompson1933likelihood, russo2016information} is a widely used algorithm for exploration strategies in contextual bandits, it (i) first samples a current posterior to obtain a model configuration; (ii) then adaptively chooses an optimal action based on the current context under the sampled model configuration and observes the corresponding reward; (iii) updates the posterior based on the context, action and reward tuples in this interaction round. 

This section evaluates the ability to guide exploration on the UCI Mushroom dataset with 8124 instances. In each instance, the mushroom has 22 features and is labeled as edible or poisonous. An agent can observe mushroom features as the context in each interaction and choose to eat or reject a mushroom. We compare our fSGLD with several baselines, including naive SGLD, functional variational FBNN and IFBNN, parametric-space variational method Bayes By Backprop (BBB) \citep{blundell2015weight}. We follow the basic settings by \citet{blundell2015weight} and consider three different reward patterns: if the agent eats an edible mushroom, it receives a reward of 5 and a reward of 0 if the agent chooses to reject. On the other hand, if the agent eats a poisonous mushroom, it receives a reward of -35 with probabilities of 0.4, 0.5, and 0.6, respectively, for three
different patterns. If the agent chooses to reject a poisonous mushroom, it receives a reward of 0. Suppose an oracle always chooses to eat an edible mushroom and reject the poisonous mushroom. The cumulative regrets with respect to the reward achieved by the oracle measure the exploration-exploitation ability of an agent. We run 10000 iterations for all methods and the minibatch size is 64. \cref{fig:bandits} shows the cumulative regrets for all methods in three reward patterns. Our fSGLD consistently outperforms other inference methods, indicating that our functional SGMCMC can provide reliable
uncertainty estimation in decision-making tasks.

\section{Discussion and Conclusions}
\label{conclude}
In this paper, we lift the It{\^o} diffusion of network parameters in BNNs onto function space and propose novel functional MCMC schemes for BNNs. Specifically, we design functional Langevin dynamics and functional Hamiltonian dynamics for the function mapping defined by BNNs and derive corresponding tractable sampling methods in parameter space, which can incorporate informative functional prior into posterior inference. Moreover, we proved that the stationary measure of our functional dynamics is the posterior over functions. Empirically, we demonstrate that our functional MCMC schemes effectively leverage functional priors to yield superior predictive performance and principled uncertainty quantification on several benchmark tasks.

\bibliography{sample_paper}
\bibliographystyle{apalike}





\newpage
\appendix
\onecolumn
\section{Notation Table}
\label{apd:notation}

\cref{tab:note} is the notation table to demonstrate the notation used in this paper.
\begin{table}[h]
\small
\centering
    \caption{Notation table}
    \label{tab:note}
    \begin{tabular}{cp{10cm}}
    \hline
    \textbf{Notation} & \text{Meanings}\\
    \hline 
    $ \mathcal{D}= \left\{\mathbf{X}_\mathcal{D}, \mathbf{Y}_\mathcal{D}\right\} $ & Training dataset\\
    $\mathcal{X} \subseteq \mathbb{R}^p$ & ($p$-dimensional) input space\\
    $\mathcal{Y} \subseteq \mathbb{R}^c$ & ($c$-dimensional) output space\\
    $(\Omega, \mathcal{A}, P)$ & Probability space on $\mathbb{R}^k$ \\
    $\mathbb{H}$ & Infinite-dimensional function space (Polish space) \\
    $\mathcal{B}(\mathbb{H})$ & Borel $\sigma$-algebra on $\mathbb{H}$ \\ 
    $\mathcal{P}(\mathbb{H})$ & The space of Borel probability measures on $\mathcal{B}(\mathbb{H})$. \\
    $X$ & Stochastic process \\
    $\mathbf{X}_{\mathcal{M}}$ & Finite measurement points from input space\\
    $\mathbf{w} \in \mathbb{R}^k $ & Random network parameters \\
    $\mathbf{z} \in \mathbb{R}^k $ & Momentum variable \\
    $\epsilon$ & Discretization step size \\
    $\eta$ & Standard Gaussian noise \\
    
    $f(\cdot; \mathbf{w})$ & Random function mapping defined by a BNN parameterized by $\mathbf{w}$\\
    $p_0(\mathbf{w})$ & Prior distribution over network parameters \\
    $p(\mathbf{w}|\mathcal{D})$ & Posterior distribution over network parameters \\
    $p(\mathbf{Y}_{\mathcal{D}}|f(\mathbf{X}_{\mathcal{D}};\mathbf{w}))$ & Likelihood function evaluated on the training data \\
    $p(\mathbf{z})$ & Auxiliary probability distribution \\
    
    $P_0$ & Functional Prior measure\\
    $\operatorname{supp}(P_{0})$ & Topological support of $P_{0}$ \\
    $E$ & A non-empty subset of $\operatorname{supp}(P_{0})$ \\
    $P_{f|\mathcal{D}}$ & Posterior measure over functions\\
    $\Phi(f)$ & Negative log-likelihood as $- \log p(\mathbf{Y}_{\mathcal{D}}|f(\mathbf{X}_{\mathcal{D}};\mathbf{w}))$ \\
    $p(g)$ & Functional auxiliary probability measure \\
    
    $U(\mathbf{w})$ & Potential energy function in parameter-space Langevin dynamics \\
    $H(\mathbf{w}, \mathbf{z})$ & Potential energy function in parameter-space Hamiltonian dynamics \\
    $I_{0}(f)$ & Onsager–Machlup (OM) functional for $P_0$ \\
    $U(f)$ & Potential energy functional in functional Langevin dynamics \\
    $H(f, g)$ & Potential energy functional in functional Hamiltonian dynamics \\
    $\mu(\cdot)$ & Drift term in the SDE of It{\^o} diffusion \\
    $\sigma(\cdot)$ & Diffusion coefficient in the SDE of It{\^o} diffusion \\
    $B$ & Standard Wiener process (Brownian motion)\\
    $\pi(\cdot)$ & Stationary distribution (measure) \\

    \hline 
    \end{tabular}
\end{table}

\section{Further Background}
\label{apd:fb}

\paragraph{It{\^o} diffusion}
Dynamics-based MCMC methods are rooted in the general framework of It{\^o} diffusion \citep{oksendal2003stochastic}, which is commonly used to model the evolution of particles in a system.  Given a stochastic process $X:[0, \infty) \times \Omega \rightarrow \mathbb{R}^n$ defined on a probability space $(\Omega, \Sigma, P)$, an It{\^o} diffusion in n-dimensional Euclidean space, driven by the standard Wiener process (Brownian motion), satisfies the following specific type of stochastic differential equation (SDE):
\begin{equation}
    \mathrm{d} X_{t} = \mu(X_{t}) \mathrm{d}t + \sigma(X_{t})\mathrm{d}B_{t},
\end{equation}
where $X_t$ represents the state of the stochastic process at time $t$, $\mu(\cdot): \mathbb{R}^n \rightarrow \mathbb{R}^n$ is a vector field describing the deterministic drift term, and $\sigma(\cdot): \mathbb{R}^n \rightarrow \mathbb{R}^{n \times n}$ is a matrix field denotes the diffusion coefficient for $X_t$. Both $\mu(\cdot)$ and $\sigma(\cdot)$ are assumed to satisfy the standard Lipschitz continuity condition \citep{ghosh2010backward}. The term $B \in \mathbb{R}^n$ refers to an n-dimensional Wiener process (Brownian motion), with $\mathrm{d}B_{t}$ being the increment of a Wiener process, distributed as $\mathrm{d}B_{t} \sim \mathcal{N}(0, \mathrm{d}t \cdot \mathbf{I}_n)$. The probability density function $p(x, t)$ of $X_t$ is governed by the Fokker-Planck (FP) equation \citep{risken1996fokker}:
\begin{equation}
     \frac{\partial p(x, t)}{\partial t} = -\sum_{i} \frac{\partial}{\partial x_{i}}\left [\mu_{i}(X_{t})p(x, t) \right] +\sum_{i, j}\frac{\partial^2}{\partial x_{i} \partial x_{j}}\left[D_{i,j}(x, t)p(x, t) \right],
    \label{eq:fp}
\end{equation}
where $D_{i,j}(x, t) = \frac{1}{2}\sigma(X_{t})\sigma(X_{t})^{T}$. This equation serves as the foundation for deriving dynamics-based MCMC sampling methods. The diffusion process reaches a stationary state when the FP equation equals zero, then simulating this diffusion process is equivalent to sampling from the stationary distribution, denoted by $\pi(x)$. By carefully designing $\mu(\cdot)$ and $\sigma(\cdot)$, one can sample from the target distribution as the stationary distribution. For example, for the commonly used Langevin dynamics, where $\mu(X) = -\nabla U(X)$, $\sigma(X) = \sqrt{2}\cdot\mathbf{I}_{n}$, and $U(\cdot)$ is a potential energy function. The stationary distribution for $p(x, t)$ is given by the Boltzmann distribution $\pi(x) \propto \exp(-U(X))$. Thus, when sampling the posterior $p(\mathbf{w}|\mathcal{D})$ in BNNs, the potential energy function is given by $U(\mathbf{w}) = -\log p(\mathbf{Y_\mathcal{D}}|\mathbf{X_\mathcal{D}}; \mathbf{w}) - \log p(\mathbf{w})$, and $\sigma(\mathbf{w}) = \sqrt{2}\cdot\mathbf{I}_{k}$, ensuring that the target posterior as the stationary distribution.

\paragraph{It{\^o} Lemma}
It{\^o} Lemma \citep{ito1951formula} is a fundamental result in stochastic calculus to find the differential of a function of a stochastic process, which serves as the stochastic calculus counterpart of the chain rule. For an It{\^o} diffusion, let $f(X_{t})$ be an arbitrary twice differentiable scalar function of real variables $X_{t}$, then the differential of $f(X_{t})$ can be derived from the Taylor series expansion of the function as
\begin{equation}
    \begin{aligned}
        \mathrm{d}f(X_{t}) = &\left(\mu(X_{t}) \frac{\partial f}{\partial x} + \frac{\sigma^2(X_{t})}{2} \frac{\partial^2 f}{\partial x^2}\right) \mathrm{d}t + \sigma(X_{t}) \frac{\partial f}{\partial x} \mathrm{d}B_t,
    \end{aligned}
\end{equation}
which implies that $f(X_{t})$ is itself an It{\^o} diffusion \citep{brzezniak2008ito}.

\paragraph{Onsager–Machlup functional} (see e.g., \citet{lambley2023strong}, Definition 2.4.) Suppose $\mathbb{H}$ is a separable metric space and $P_{0} \in  \mathcal{P}(\mathbb{H})$. Assume that $\varnothing \neq E \subseteq \operatorname{supp}(P_{0})$, then a functional $I_{0}: E \rightarrow \mathbb{R}$ is an Onsager–Machlup (OM) functional for $P_{0}$ if for all $f, f' \in E$,
\begin{equation}
    \lim_{r\rightarrow 0} \frac{P_{0}(B_{r}(f)}{P_{0}(B_{r}(f')} = \exp (I_{0}(f') - I_{0}(f)),
\end{equation}
where $B_{r}(\cdot)$ denotes the closed ball of radius $r$ in $\mathbb{H}$ and $\operatorname{supp}(P_{0}):= \{f\in \mathbb{H}\ |\ P_{0}(B_{r}(f))>0 \ \text{for all}\ r>0\}$. \citet{lambley2023strong} (in Proposition 2.6.) proved that posterior measure  $P_{f|\mathcal{D}} \in \mathcal{P}(\mathbb{H})$ defined as $P_{f|\mathcal{D}}(\mathrm{d}f) \propto \exp(-\Phi(f)) P_0(\mathrm{d}f)$ has OM functional $I^{y}:  E \rightarrow \mathbb{R}$ given by $I^{y}(f) = I_{0}(f) + \Phi(f)$, where $\Phi(f)$ is the continuous \textit{potential}, in essence the negative log-likelihood as $- \log p(\mathbf{Y}_{\mathcal{D}}|f(\mathbf{X}_{\mathcal{D}};\mathbf{w}))$. OM functional can be interpreted heuristically as the negative logarithm of the Lebesgue density, but this interpretation cannot be taken literally in infinite-dimensional spaces, as there is no Lebesgue measure in such settings. For instance, the OM functional for a Gaussian measure on an infinite-dimensional Banach space is typically defined only on a specific subspace, known as the Cameron-Martin space, e.g.,  for a centred Gaussian measure, the OM functional is defined as $I: E \rightarrow \mathbb{R}, I(f) = \frac{1}{2}||f||^{2}_{E}$. As a result, the OM functional does not need to be defined over the entire space $\mathbb{H}$.

\section{Theoretical Proof}
\label{proof}

The proof of \cref{prop:flg} and \cref{prop:fhmc} will be concise based on the general framework for dynamics-based sampling methods proposed by \citet{ma2015complete}. They proved in Theorem 1 that if $\mu(X_{t})$ and $\sigma(X_{t})$ are restricted to the following form
\begin{equation}
    \begin{aligned}
         \mu(X) =& -[\mathbf{D}(X) + \mathbf{Q}(X)]\nabla G(X) + \Gamma(X),\\
         \Gamma(X) =& \sum \frac{\partial}{\partial X}(\mathbf{D}_{ij}(X) + \mathbf{Q}_{ij}(X)), \\
         \sigma(X) =& \sqrt{2\mathbf{D}(X)},
    \end{aligned}
    \label{eq:gf}
\end{equation}
where $G(\cdot)$ is the potential energy functional, $\mathbf{D}(\cdot)$ is a positive semidefinite matrix, and $\mathbf{Q}(\cdot)$ is a skew-symmetric curl matrix, then the Fokker–Planck equation can be transformed into a more compact form \citep{yin2006existence, shi2012relation} given by
\begin{equation}
    \begin{aligned}
        \frac{\partial p(x, t)}{\partial t} = \nabla^T \cdot([\mathbf{D}(X)+\mathbf{Q}(X)][p(x, t) \nabla G(X)+\nabla p(x, t)]).
    \label{eq:fk}
    \end{aligned}
\end{equation}
Then, it is straightforward to verify that the stationary distribution of the diffusion process is exactly as $\pi(X)\propto \exp(-G(X))$. 

For the proof of \cref{prop:flg}, it is clear that the functional Langevin dynamics in \cref{eq:flg} can be cast into the above general framework as 
\begin{equation}
    \begin{aligned}
        \mathrm{d}f_{t}(\cdot; \mathbf{w}) & = [-(\nabla_{\mathbf{w}}f_{t})^{T}(\nabla_{\mathbf{w}}f_{t})\nabla_{f}U(f_{t}) + H_{\mathbf{w}}f]\mathrm{d}t + \sqrt{2}(\nabla_{\mathbf{w}}f_{t})^{T}\mathrm{d}B_t\\
        & = \left(-[\mathbf{D}(f) + \mathbf{Q}(f)]\nabla_{f} U(f_{t}) + \Gamma(f)\right)\mathrm{d}t + \sqrt{2\mathbf{D}(f)}\mathrm{d}B_t,
    \end{aligned}
\end{equation}
where $ U(f) = -\log p(\mathbf{Y}_{\mathcal{D}}|f(\mathbf{X}_{\mathcal{D}}; \mathbf{w})) + I_{0}(f)$ is an OM functional for $P_{f|\mathcal{D}}$, $\mathbf{D}(f) = (\nabla_{\mathbf{w}}f)^{T}(\nabla_{\mathbf{w}}f)$, $\mathbf{Q}(f) = \mathbf{0}$, $\Gamma(f) = H_{\mathbf{w}}f$. Therefore, the Fokker-Planck equation of the probability measure $p(f, t)$ for $f_{t}(\cdot; \mathbf{w})$ then can be derived as 
\begin{equation}
    \frac{\partial p(f, t)}{\partial t} = \nabla^{T} \cdot \left([\mathbf{D}(f) + \mathbf{Q}(f)][p(f, t)\nabla_{f} U(f) + \nabla p(f, t)]\right),
\end{equation}
from which the stationary probability measure of $p(f, t)$ can be verified as $\exp (- U(f)) = P_{f|\mathcal{D}}$, that is, the target posterior measure over functions. 

For the proof of \cref{prop:fhmc}, the functional Hamiltonian dynamics defined in \cref{eq:fhmc} can be written as 
\begin{equation}
    \begin{aligned}
        \mathrm{d}\left[\begin{array}{l}
        f_{t}(\cdot; \mathbf{w}) \\
        g_{t}(\cdot; \mathbf{z})
        \end{array}\right] 
        & =-\left[\begin{array}{cc}
        0 & -(\nabla_{\mathbf{w}}f)^{T} \nabla_{\mathbf{z}} g_{t} \\
        (\nabla_{\mathbf{z}} g)^{T} \nabla_{\mathbf{w}}f_{t} & 0
       \end{array}\right]\left[\begin{array}{c}
       \nabla_{f} U(f_{t}) \\
       -\nabla_{g} \log p(g_{t})
       \end{array}\right] \mathrm{d} t\\
       & = -\mathbf{Q}(f, g)\nabla H(f_{t}, g_{t})  \mathrm{d} t,
    \end{aligned}
    \label{eq:fhmc2}
\end{equation}
where $\mathbf{Q}(f, g)$ is s a skew-symmetric curl matrix, $\mathbf{D}(f, g) = \mathbf{0}$. Therefore, the stationary measures of the functional Hamiltonian dynamics defined in \cref{eq:fhmc} is $\pi(f, g) \propto \exp (H(f, g))$, and simply (marginalize) discard the auxiliary functions can obtain the target true posterior $P_{f|\mathcal{D}}$.

For the stochastic gradient version, the functional stochastic gradient Hamiltonian dynamics (fSGHMC) for the discretization update
rule for samples of $\mathbf{w}$ and $\mathbf{z}$ defined in \cref{eq:fsghmc} is defined as follows:
\begin{equation}
    \begin{aligned}
        \mathrm{d}\left[\begin{array}{l}
        f_{t}(\cdot; \mathbf{w}) \\
        g_{t}(\cdot; \mathbf{z})
        \end{array}\right] 
        & =-\left[\begin{array}{cc}
        0 & -(\nabla_{\mathbf{w}}f)^{T} \nabla_{\mathbf{z}} g_{t} \\
        (\nabla_{\mathbf{z}} g)^{T} \nabla_{\mathbf{w}}f_{t} & C(\nabla_{\mathbf{z}} g)^{T} \nabla_{\mathbf{z}} g_{t}
       \end{array}\right]\left[\begin{array}{c}
       \nabla_{f} \tilde{U}(f_{t}) \\
       -\nabla_{g} \log p(g_{t})
       \end{array}\right] \mathrm{d} t + \left[\begin{array}{cc} \mathbf{0} & \mathbf{0} \\ \mathbf{0} & \sqrt{2C}(\nabla_{\mathbf{z}} g_{t})^{T} \end{array}\right] \mathrm{d} t,
    \end{aligned}
    \label{eq:ffsghmc}
\end{equation}

where $\mathbf{D}(f, g)=\left(\begin{array}{cc}
\mathbf{0} & \mathbf{0} \\
\mathbf{0} & C(\nabla_{\mathbf{z}} g)^{T} \nabla_{\mathbf{z}} g
\end{array}\right)$ and $\mathbf{Q}(f, g) = \left(\begin{array}{cc} \mathbf{0} & -(\nabla_{\mathbf{w}}f)^{T} \nabla_{\mathbf{z}} g \\ (\nabla_{\mathbf{z}} g)^{T} \nabla_{\mathbf{w}}f & \mathbf{0} \end{array}\right)$ is the same as that in full-batch functional Hamiltonian dynamics. Then, the (marginal) stationary probability measure of such functional stochastic gradient Hamiltonian is still the target functional posterior $P_{f|\mathcal{D}}$. 

Moreover, the comparisons of matrix $\mathbf{D}(\cdot)$ and $\mathbf{Q}(\cdot)$ for naive SGMCMC methods and our functional SGMCMC are shown in \cref{tab:fwmcmc}.

\begin{table*}[!t]
\caption{General framework of naive SGMCMC and functional SGMCMC algorithms.}
\label{tab:fwmcmc}
\begin{center}
\begin{small}
\begin{sc}
\begin{tabular}{lcccc}
\toprule
Dynamics~~ &  $X$~~ & $G(X)$~~ & $\mathbf{Q}(X)$ & $\mathbf{D}(X)$ \\
\midrule
        SGLD & $\mathbf{w}$ & $\tilde{U}(\mathbf{w})$ & $\mathbf{0}$ & $\mathbf{I}_d$ \\

        fSGLD & $f$ & $\tilde{U}(f)$ & $\mathbf{0}$ & $(\nabla_{\mathbf{w}}f)^{T}(\nabla_{\mathbf{w}}f)$ \\
        
        SGHMC  & $(\mathbf{w}, \mathbf{z})$ & $\Tilde{H}(\mathbf{w}, \mathbf{z})$ & $\left(\begin{array}{cc} \mathbf{0} & -\mathbf{I}_d \\ \mathbf{I}_d & \mathbf{0} \end{array}\right)$ & $\left(\begin{array}{cc} \mathbf{0} & \mathbf{0} \\ \mathbf{0} & C \end{array}\right)$\\

        fSGHMC  
        & $(f, g)$ 
        & $\Tilde{H}(f, g)$ 
        & ~~$\left(\begin{array}{cc} \mathbf{0} & -(\nabla_{\mathbf{w}}f)^{T} \nabla_{\mathbf{z}} g \\ (\nabla_{\mathbf{z}} g)^{T} \nabla_{\mathbf{w}}f & \mathbf{0} \end{array}\right)$~~ 
        & $\left(\begin{array}{cc} \mathbf{0} & \mathbf{0} \\ \mathbf{0} & C(\nabla_{\mathbf{z}} g)^{T} \nabla_{\mathbf{z}} g \end{array}\right)$\\
\bottomrule
\end{tabular}
\end{sc}
\end{small}
\end{center}
\end{table*}

\section{Pseudocode for Functional SGMCMC.}
\label{apdix}
\cref{alg:fsgld} presents the pseudocode for fSGLD. And the pseudocode for fSGHMC is shown in \cref{alg:fsghmc}.

\begin{algorithm}[]
   \caption{Functional SGLD (fSGLD)}
   \label{alg:fsgld}
\begin{algorithmic}
   \STATE {\bfseries Input:} Dataset $\mathcal{D}$, pre-trained GP prior $p(f)$, initialized $\mathbf{w}_{0} \sim \mathcal{N}(0, I)$, step size $\epsilon_{t}$, noise $\eta_t \sim \mathcal{N}(0, 1)$, number of burn-in iterations $N_b$, number of sample $K$
   \STATE $\#$ Burn-in stage :
   \FOR{$t=0$ {\bfseries to} $N_b$}
   \STATE draw measurement set $\mathbf{X}_{\mathcal{M}}$;
   \STATE update $\mathbf{w}_{t+1}\leftarrow\mathbf{w}_{t}- \epsilon_{t}\nabla_{\mathbf{w}}\tilde{U}(f_{t})) + \sqrt{2\epsilon_{t}} \eta_{t} $ using \cref{eq:fsgld};
   \ENDFOR
   
   \STATE $\#$ Sampling stage :
   \STATE $S\leftarrow \varnothing$;
   \STATE $\mathbf{w}_0 \leftarrow \mathbf{w}_{N_b}$;
   \FOR{$t=0$ {\bfseries to} $K$}
   \STATE draw measurement set $\mathbf{X}_{\mathcal{M}}$;
   \STATE $\mathbf{w}_{t+1}\leftarrow\mathbf{w}_{t}- \epsilon_{t}\nabla_{\mathbf{w}}\tilde{U}(f_{t})) + \sqrt{2\epsilon_{t}} \eta_{t} $ using \cref{eq:fsgld};
   \STATE $S\leftarrow S \cup \left\{\mathbf{w}_{t+1}\right\}$;
   \ENDFOR
    
\end{algorithmic}
\end{algorithm}

\begin{algorithm}[]
   \caption{Functional SGHMC (fSGHMC)}
   \label{alg:fsghmc}
\begin{algorithmic}
   \STATE {\bfseries Input:} Dataset $\mathcal{D}$, pre-trained GP prior $p(f)$, initialized $\mathbf{w}_{0} \sim \mathcal{N}(0, I)$, $\mathbf{z}_{0} \sim \mathcal{N}(0, M)$, step size $\epsilon_{t}$, noise $\eta_t \sim \mathcal{N}(0, 1)$,  number of burn-in iterations $N_b$, leapfrog steps $m$, number of sample $K$
   \STATE $\#$ Burn-in stage :
   \FOR{$t=0$ {\bfseries to} $N_b$}
   \STATE resample momentum $\mathbf{z}_{t} \sim \mathcal{N}(0, M)$ 
   \STATE $(\mathbf{w}_0, \mathbf{z}_0)\leftarrow(\mathbf{w}_t, \mathbf{z}_t)$;
   \FOR{$i=1$ {\bfseries to} $m$}
   \STATE draw measurement set $\mathbf{X}_{\mathcal{M}}$;
   \STATE update $(\mathbf{w}_i, \mathbf{z}_i)$ using \cref{eq:fsghmc};
   \ENDFOR
   \STATE $(\mathbf{w}_{t+1}, \mathbf{z}_{t+1})\leftarrow(\mathbf{w}_{m}, \mathbf{z}_{m})$
   \ENDFOR
   
   \STATE $\#$ Sampling stage :
   \STATE $S\leftarrow \varnothing$;
   \STATE $\mathbf{w}_0 \leftarrow \mathbf{w}_{N_b}$;
   \FOR{$t=0$ {\bfseries to} $K$}
   \STATE resample momentum $\mathbf{z}_{t} \sim \mathcal{N}(0, M)$ 
   \STATE $(\mathbf{w}_0, \mathbf{z}_0)\leftarrow(\mathbf{w}_t, \mathbf{z}_t)$;
   \FOR{$i=1$ {\bfseries to} $m$}
   \STATE draw measurement set $\mathbf{X}_{\mathcal{M}}$;
   \STATE update $(\mathbf{w}_i, \mathbf{z}_i)$ using \cref{eq:fsghmc};
   \ENDFOR
   \STATE $(\mathbf{w}_{t+1}, \mathbf{z}_{t+1})\leftarrow(\mathbf{w}_{m}, \mathbf{z}_{m})$
   \STATE $S\leftarrow S \cup \left\{\mathbf{w}_{t+1}\right\}$;
   \ENDFOR

\end{algorithmic}
\end{algorithm}

\section{Experimental Setting}
\label{apd:expset}
\paragraph{Extrapolation on Synthetic Data}
 In this experiment, we use $2\times100$ fully connected tanh neural networks for all models. The functional GP prior with the RBF kernel is pre-trained on the 20 training points for 100 epochs for all functional methods. we use a decreasing step-size schedule for $\epsilon_t$ with 1e-3 as the initial value for all sampling methods. The specific decreasing schedule is as every 5000 iterations, the decay factor is 0.9.

\paragraph{Multivariate regression on UCI datasets}
In this experiment, we use two-hidden-layer fully connected neural networks and each layer with 10 hidden units for all models. The functional GP prior with the RBF kernel is pre-trained for 100 epochs. We run 500 iterations for the burn-in stage and collect 15 samples in the following 1500 iterations for all sampling methods. The initial $\epsilon_t$ is 1e-3, and the decay factor is 0.9. Functional variational inference methods are trained for 2000 epochs for fair comparison. 

\paragraph{Imagine Classification and OOD Detection}
In this experiment, we use ResNet-18 architecture for all methods. The functional prior is a Dirichlet-based GP designed for classification tasks and is pre-trained for 500 epochs. The initial $\epsilon_t$ is 1e-2, the decay period is 20, and the decay factor is 0.99. For all sampling methods, we run for 60 burn-in iterations and collect 10 samples in the following 20 iterations on the MNIST dataset, 100
burn-in iterations and collect 10 samples in the following 100 iterations on Fashion-MNIST, 80 burn-in iterations and draw 10 samples in the following 20 iterations on CIFAR10. Functional variational inference methods are trained for 80, 200, and 100 epochs on MNIST, FashionMNIST and CIFAR-10, respectively.

\paragraph{Contextual Bandits}
In this experiment, we use fully connected neural networks with input-100-100-output architecture for all methods. The GP prior with RBF kernel is pre-trained on 1000 randomly sampled points from training data for 100 epochs. The initial $\epsilon_t$ is 1e-2, and the decay factor is 0.9. ALL models are trained using the last 4096 input-output tuples in the training buffer with a batch size of 64 and training frequency of 64 for each iteration.

\section{Analysis of Mixing Time}
\label{apd:mix}
\begin{figure}[t!]
    \centering
  {
    \subfigure[fSGLD-toy]{\label{fig:fp_toy}%
       \includegraphics[width=0.4\linewidth]{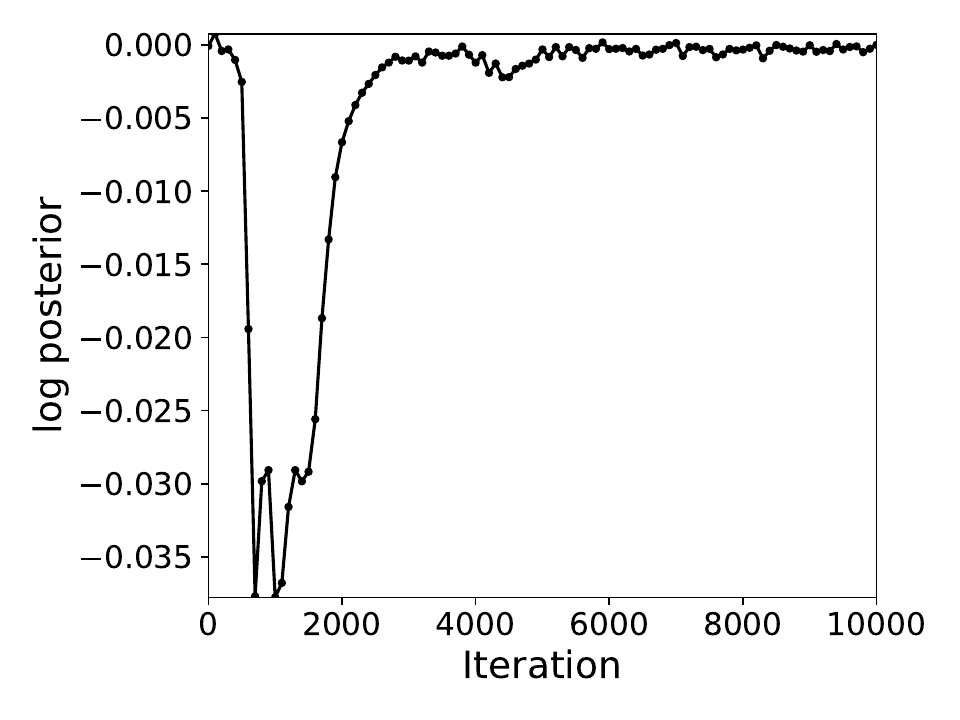}}%
    \subfigure[fSGLD-uci]{\label{fig:fp_uci}%
      \includegraphics[width=0.4\linewidth]{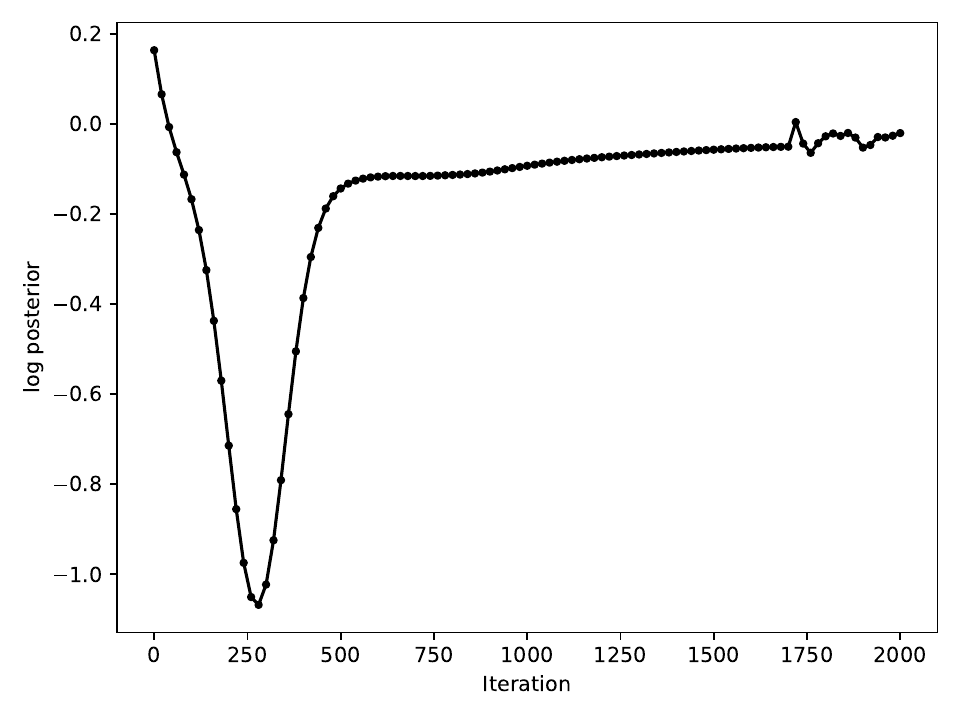}}%
    
    \subfigure[SGLD-toy]{\label{fig:pp_toy}%
      \includegraphics[width=0.4\linewidth]{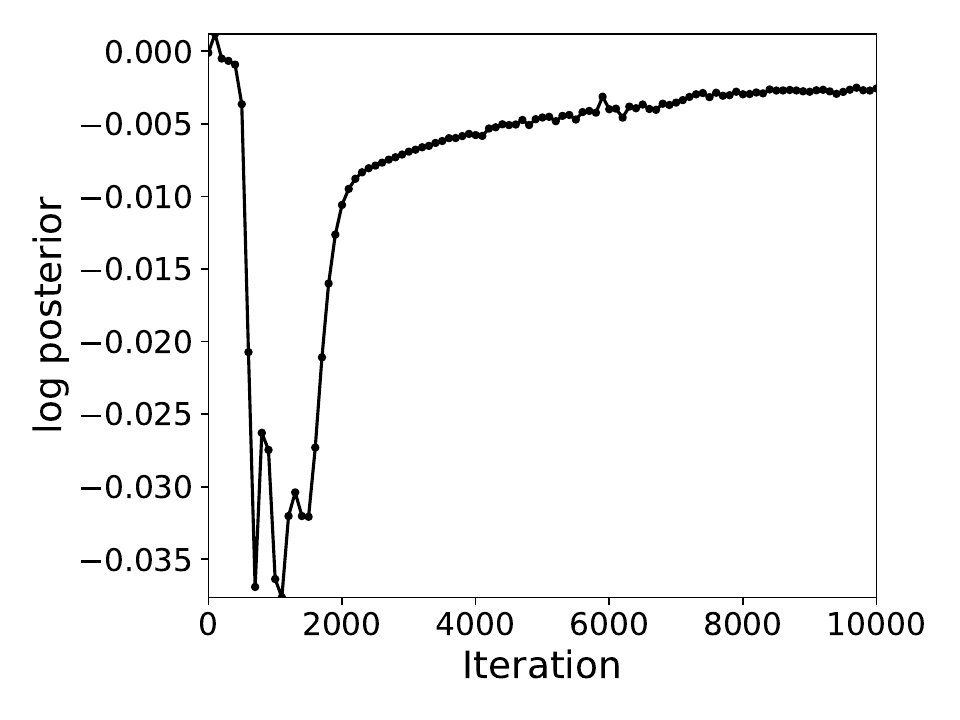}}%
    \subfigure[SGLD-uci]{\label{fig:pp_uci}%
      \includegraphics[width=0.4\linewidth]{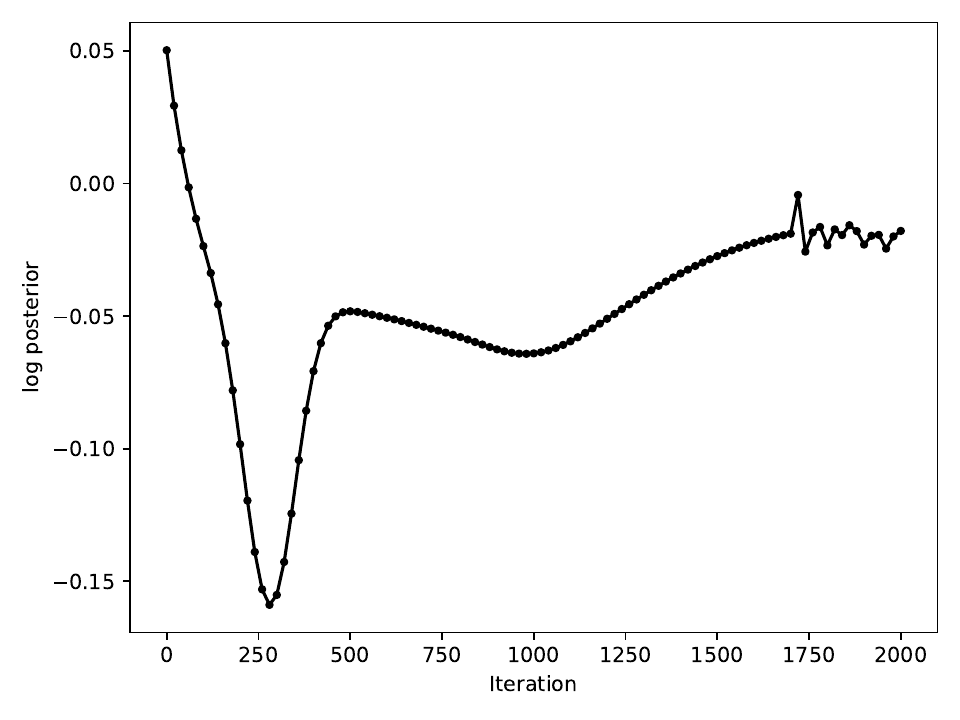}}%
    \caption{Log-posterior probability versus the number of iterations.}
    \label{fig:log_posterior}}
\end{figure}

The trajectories of our fSGLD and the naive parameter-space SGLD for both the 1-D extrapolation and UCI regression (Yacht) experiments are shown in \cref{fig:log_posterior}. For the extrapolation example, we use a two-hidden-layer fully connected neural network with 100 hidden units in each layer (resulting in a 10401-dimensional parameter space). For the UCI regression task, we use a 2×10 fully connected network (141-dimensional parameter space). As illustrated in \cref{fig:log_posterior}, our fSGLD converges rapidly to the stationary measure—within 2000 iterations for the extrapolation example and 500 iterations for the UCI regression task. This convergence rate is comparable to that of the parameter-space SGLD, highlighting the rapid mixing rate of our method.

\section{Computational Complexity}
\label{apd:compu}
The runtime comparisons of a single iteration for all methods on the 1-D extrapolation, UCI regression (yacht) and image classification (MNIST) are provided in \cref{tab:runtime}. For our fSGLD, fSGHMC, functional variational IFBNN and FBNN methods, the functional GP prior was pre-trained for 100, 100, and 500 epochs, respectively, in these three experiments, which added only an extra 1s for both the 1-D and UCI experiments, and 3s for the classification task. Compared to functional variational inference methods, our functional SGMCMC schemes exhibit significantly better computational performance, and are only slightly slower than the naive parameter-space SGMCMC. This demonstrates the computational efficiency of our method while maintaining the advantages of incorporating functional priors for improved inference.

\begin{table}[t]
\caption{Runtime comparison of single iteration for all methods (in seconds).}
\label{tab:runtime}
\vskip 0.15in
\begin{center}
\begin{small}
\begin{sc}
\begin{tabular}{lcccc}
\toprule
Model & 1-D extrapolation & UCI & Image classification  \\
\midrule
SGLD    & 0.0016 & 0.0025 & 4 \\
fSGLD   & 0.0044 & 0.0045 & 5.83 \\
SGHMC    & 0.0023 & 0.02 & 10.95 \\
fSGHMC    & 0.0051 & 0.044 & 13.64 \\
IFBNN     & 0.0094 & 0.0075 & 6.81 \\
FBNN      & 0.129 & 0.114 & 23.25\\
\bottomrule
\end{tabular}
\end{sc}
\end{small}
\end{center}
\vskip -0.1in
\end{table}

\section{Ablation Study}
\label{apd:abl}
\subsection{The effects of the sample size for naive SGMCMC and functional SGMCMC}
\begin{figure*}[t!]
    \centering
  {
    \subfigure[fSGLD-10]{\label{fig:fsgld10}%
       \includegraphics[width=0.3\linewidth]{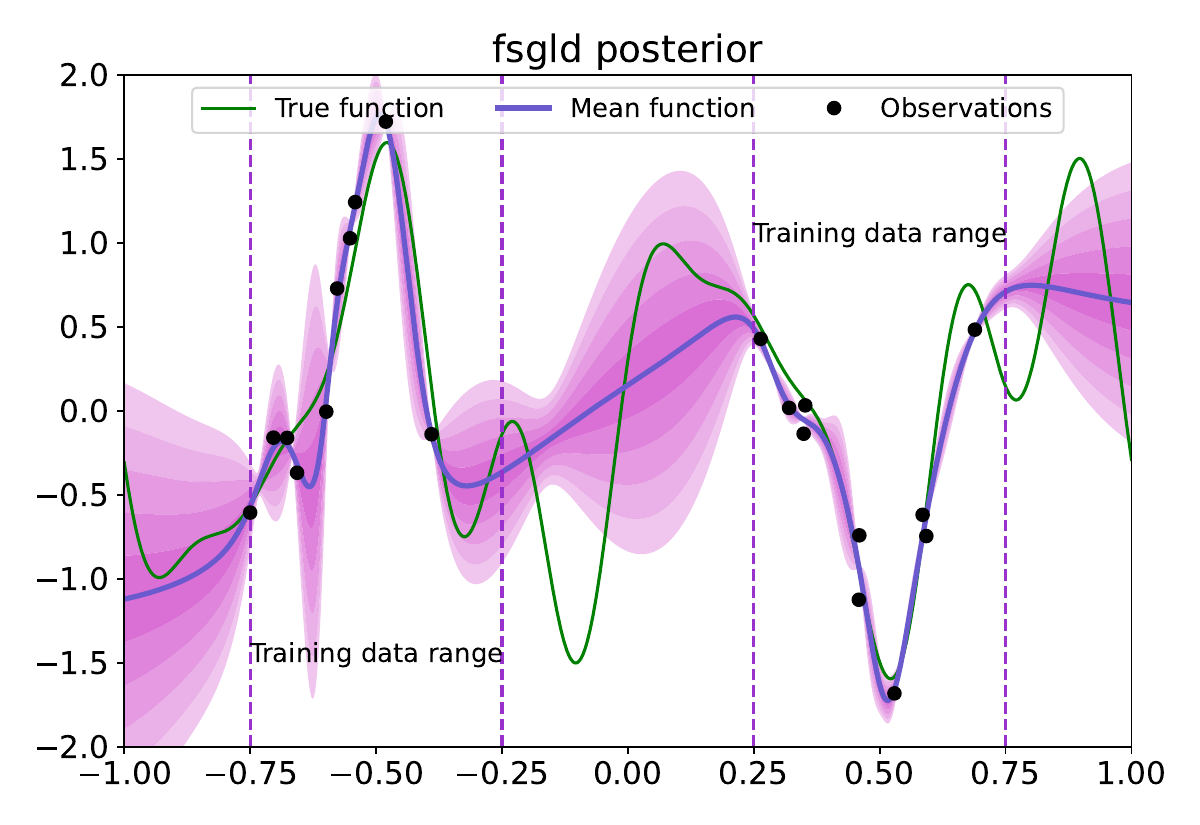}}%
    \subfigure[fSGLD-80]{\label{fig:fsgld80}%
      \includegraphics[width=0.3\linewidth]{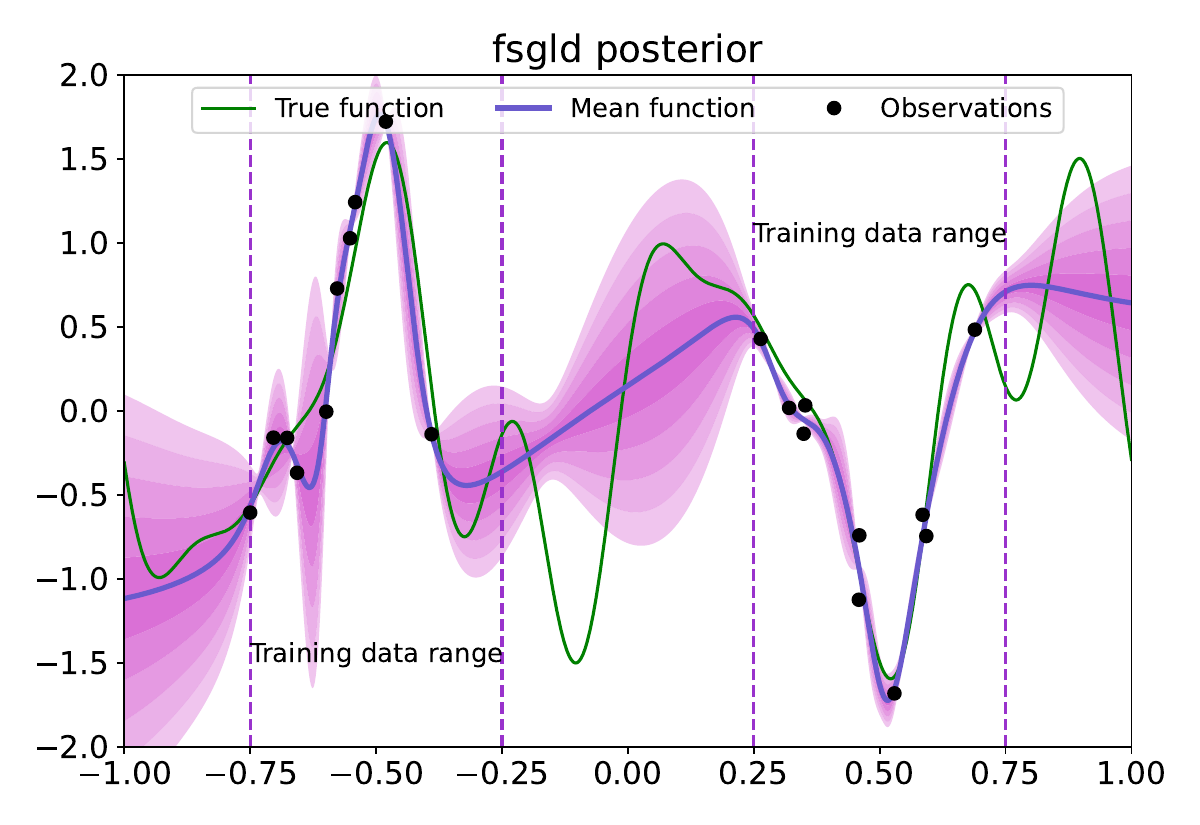}}%
    \subfigure[fSGLD-200]{\label{fig:fsgld200}%
      \includegraphics[width=0.3\linewidth]{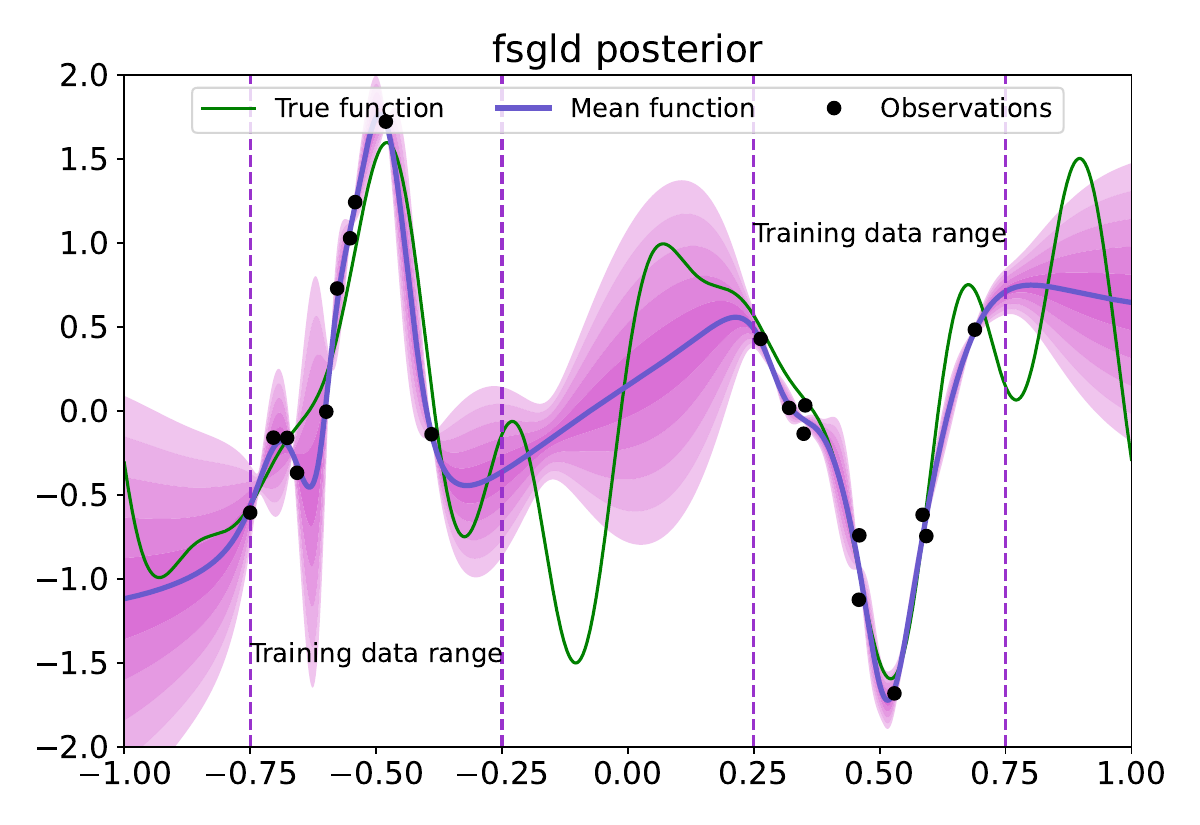}}%
    
    \subfigure[SGLD-10]{\label{fig:psgld10}%
      \includegraphics[width=0.3\linewidth]{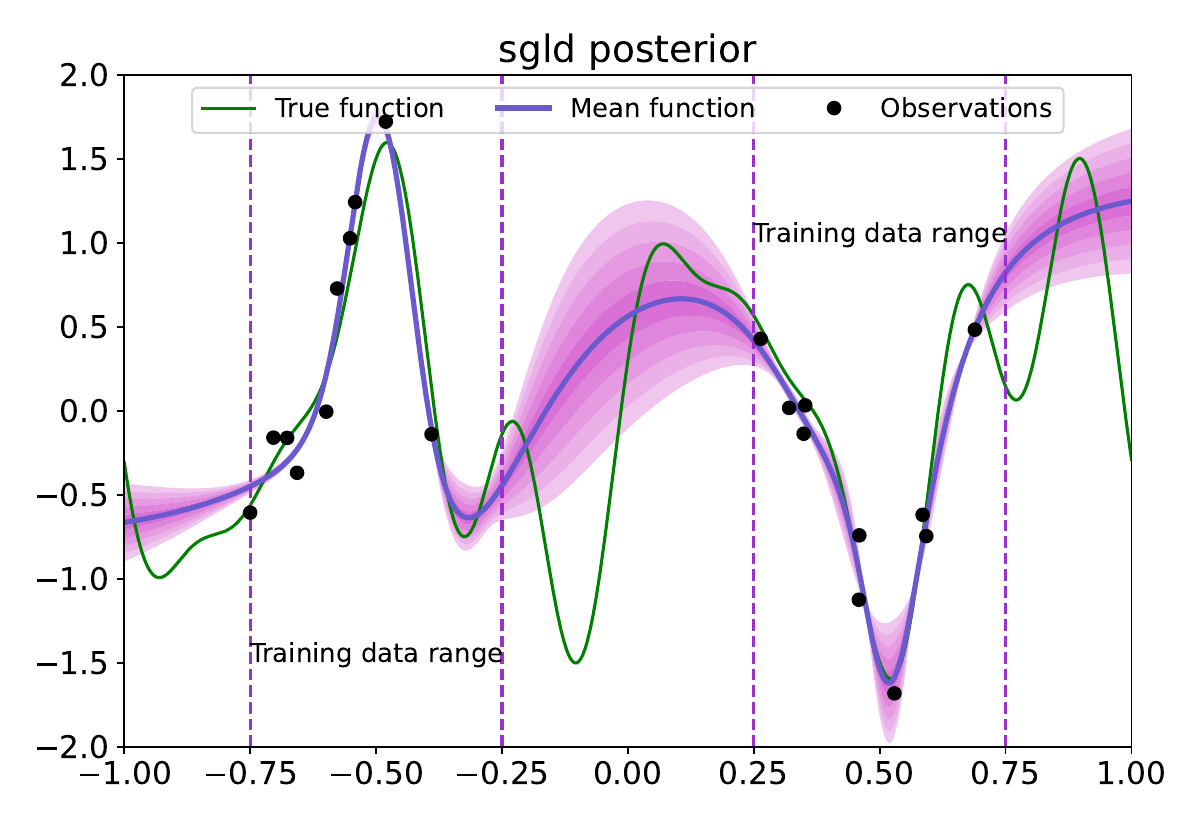}}%
    \subfigure[SGLD-80]{\label{fig:psgld80}%
      \includegraphics[width=0.3\linewidth]{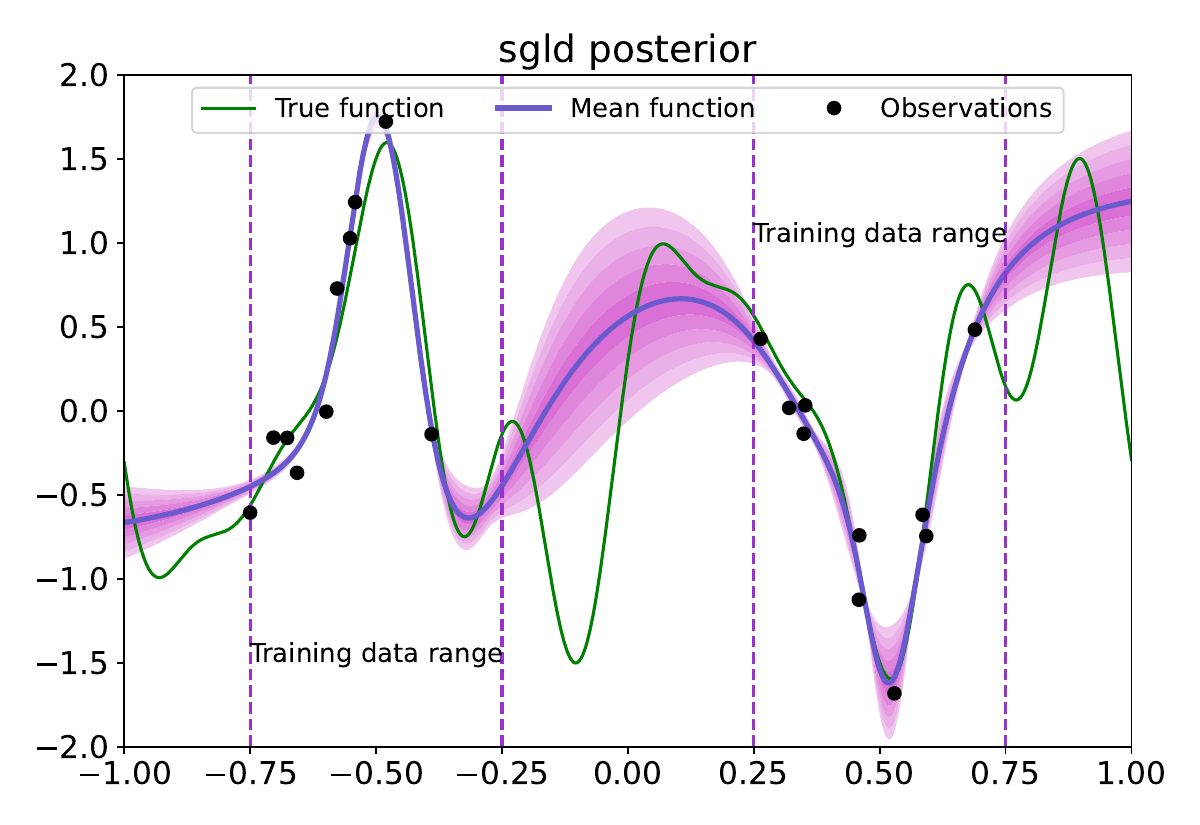}}%
    \subfigure[SGLD-200]{\label{fig:psgld200}%
      \includegraphics[width=0.3\linewidth]{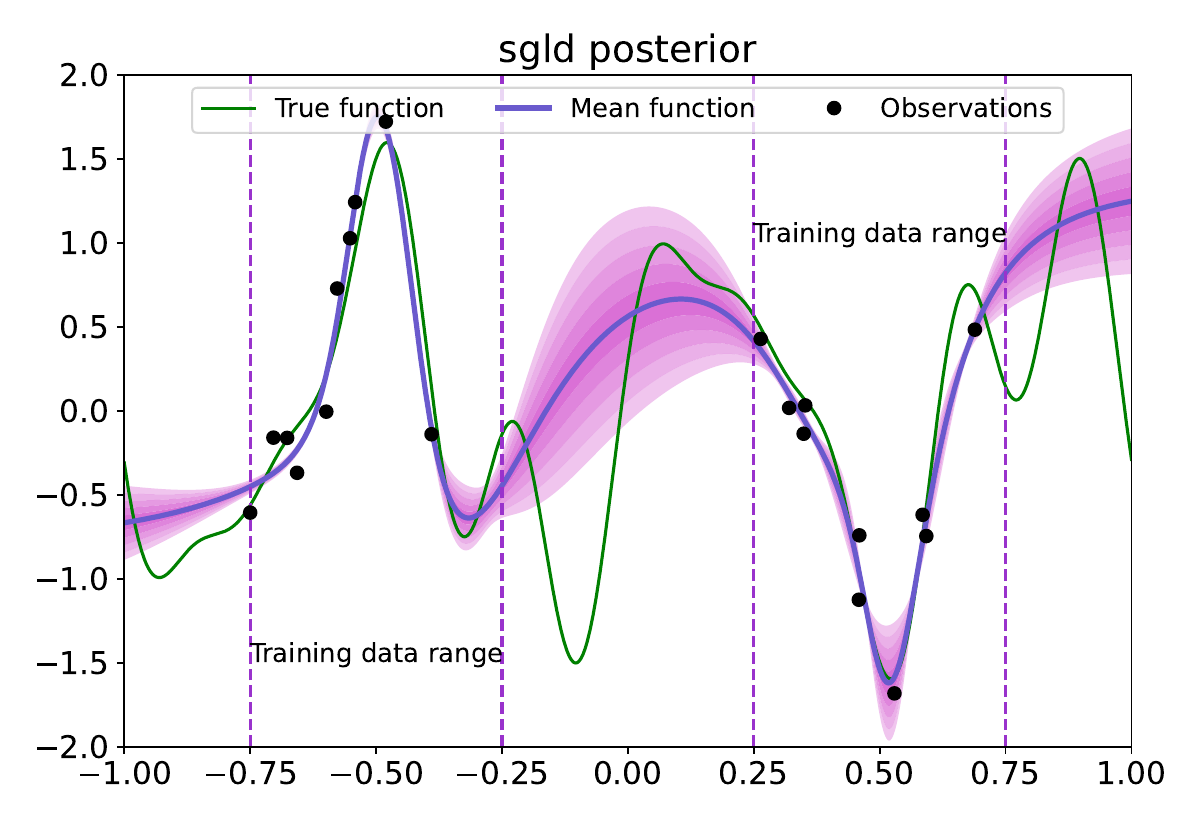}}%
    
    \subfigure[fSGHMC-10]{\label{fig:fsghmc10}%
       \includegraphics[width=0.3\linewidth]{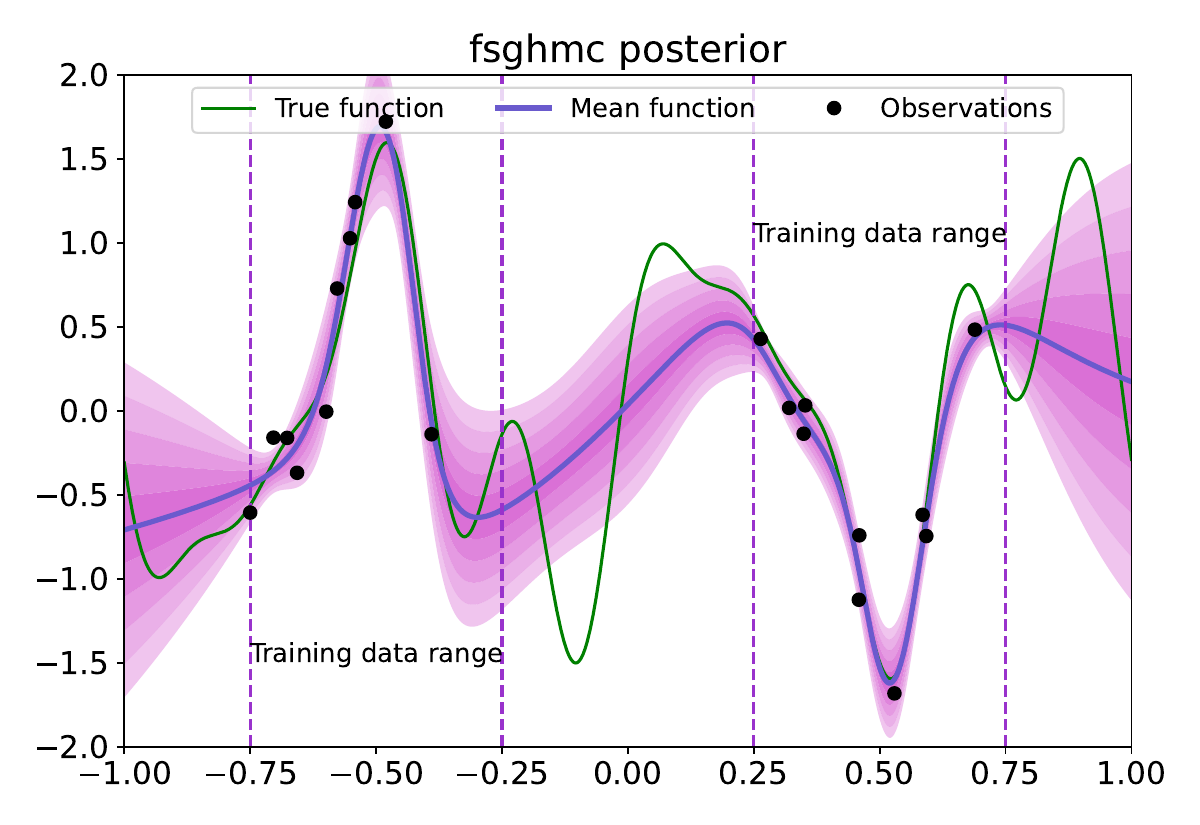}}%
    \subfigure[fSGHMC-80]{\label{fig:fsghmc80}%
       \includegraphics[width=0.3\linewidth]{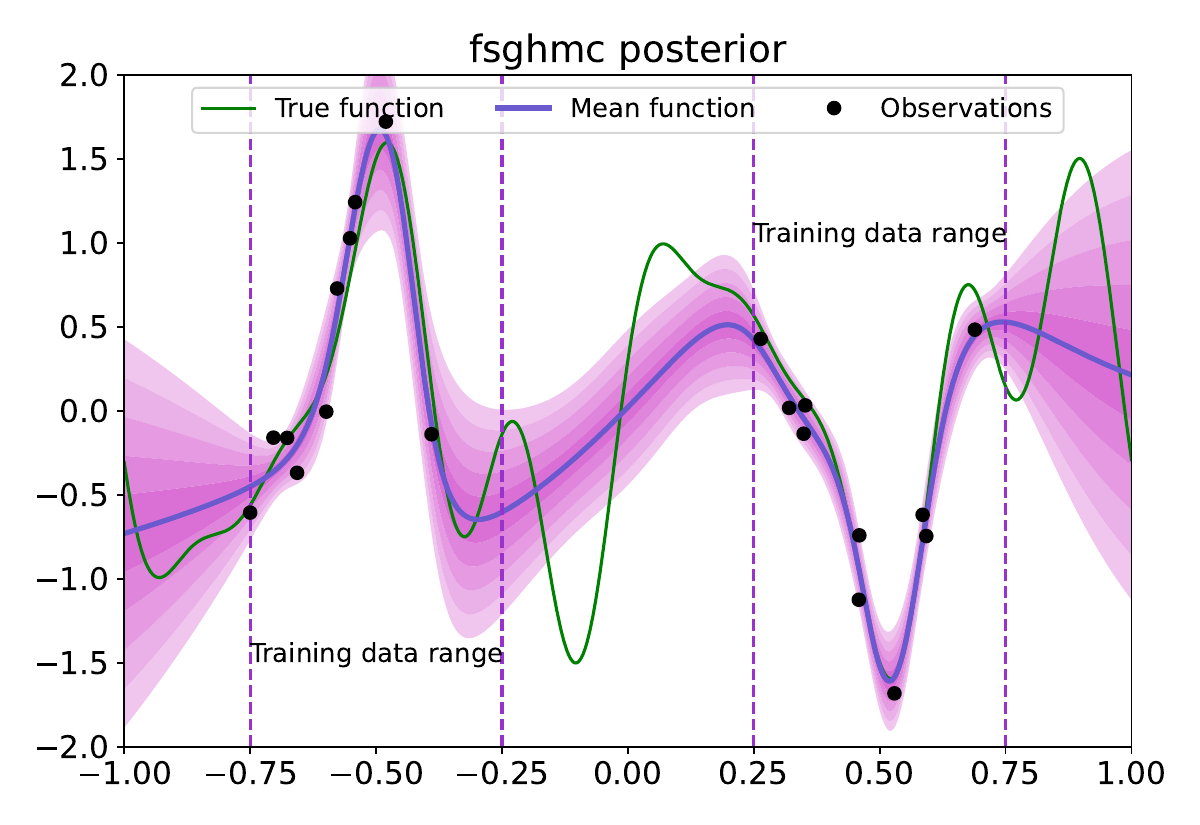}}%
   \subfigure[fSGHMC-200]{\label{fig:fsghmc200}%
       \includegraphics[width=0.3\linewidth]{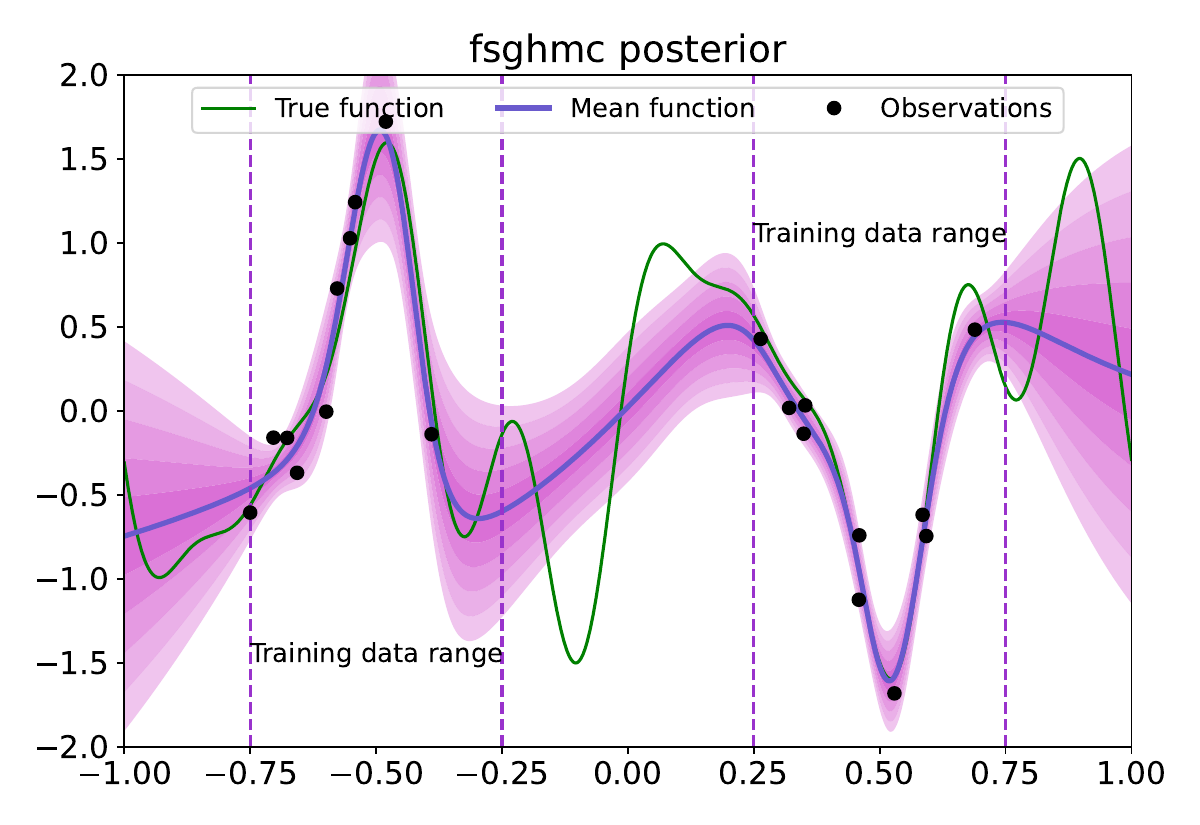}}%
    
    \subfigure[SGHMC-10]{\label{fig:sghmc10}%
       \includegraphics[width=0.3\linewidth]{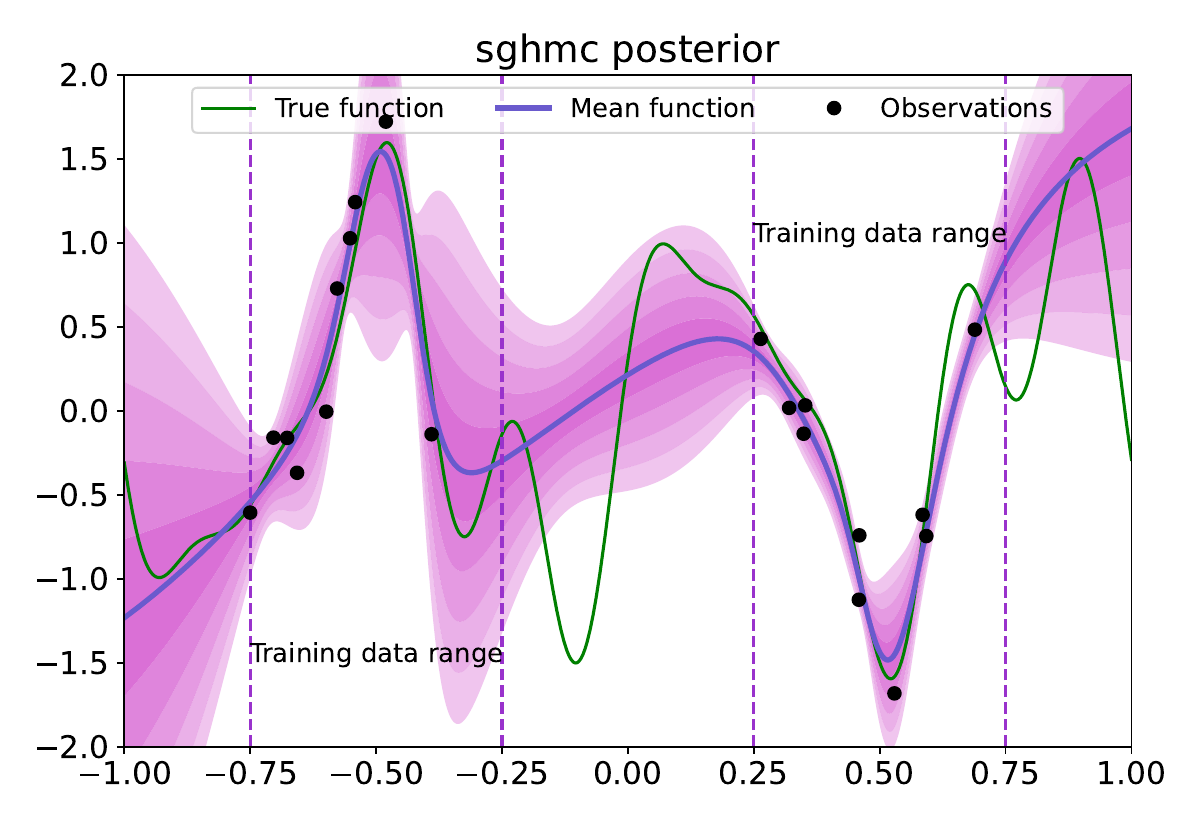}}%
    \subfigure[SGHMC-80]{\label{fig:sghmc80}%
       \includegraphics[width=0.3\linewidth]{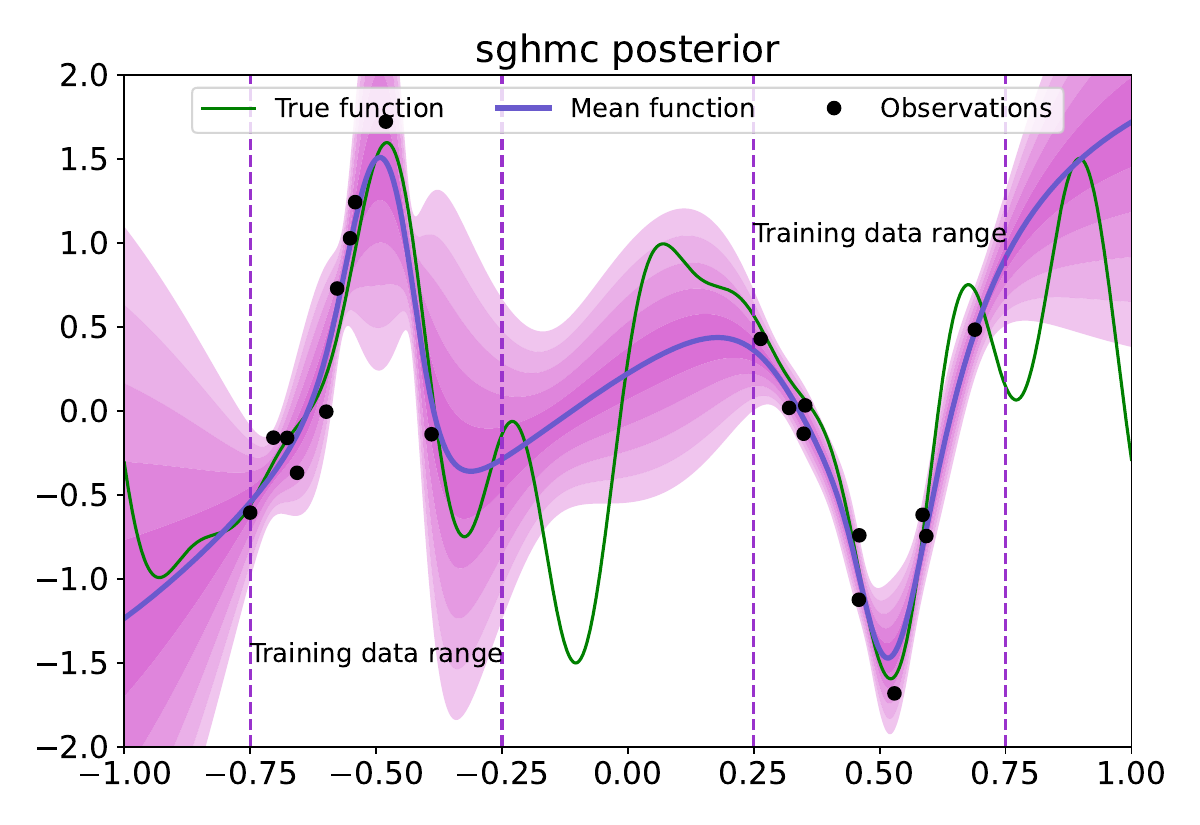}}%
   \subfigure[SGHMC-200]{\label{fig:sghmc200}%
       \includegraphics[width=0.3\linewidth]{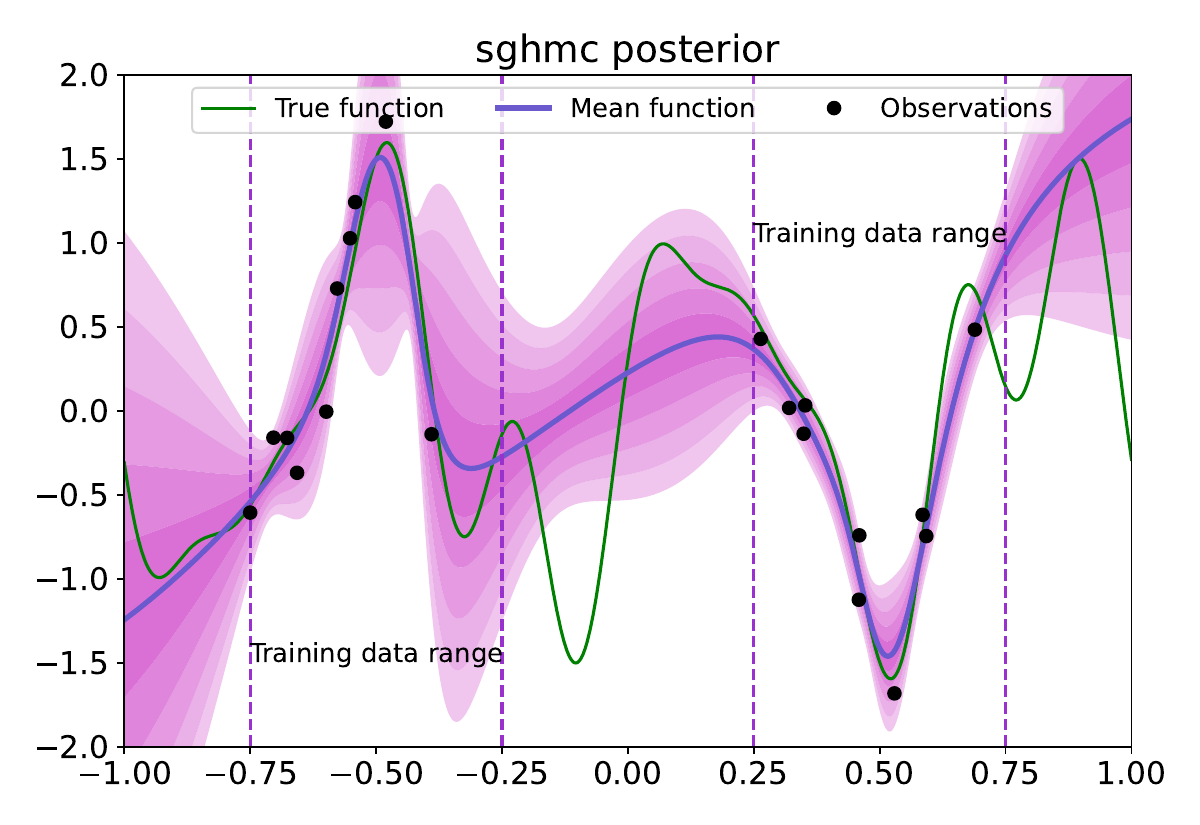}}%
    \caption{The effect of sample size on 1-D extrapolation example for fSGLD, SGLD, fSGHMC, SGHMC. The number after the short line in each subheading represents the sample size.}
    \label{fig:sample_size}}
\end{figure*}

In this section, we investigate the effect of sample size on the results of naive SGMCMC and our functional SGMCMC  by varying the number of samples. For the 1-D extrapolation example in \cref{sec:1d}. We consider three different sample sizes for SGLD, SGHMC, fSGLD, and fSGHMC: 10, 80, and 200, respectively. The extrapolation results are shown in \cref{fig:sample_size}. We can see that no matter the sample size, there is almost no difference in the fitting effect and uncertainty estimation for both naive SGMCMC and our functional SGMCMC, which indicates that the sample size has almost no effect on our experimental results.

For the UCI regressions in \cref{sec:uci}, the sample size is 15 for all naive SGMCMC and functional SGMCMC methods. We now consider a bigger sample size of 150 for all methods on the \textit{Yacht} and \textit{Boston} datasets. The RMSE and NLL results for the naive SGMCMC and functional SGMCMC are shown in \cref{tab:size_rmse} and \cref{tab:size_nll}, respectively. We randomly split each dataset into 90\% training data and 10\% test data, which is repeated 5 times to ensure validity. These two results have only very slight fluctuations compared to the results in \cref{tab:rmse} and \cref{tab:nll}, which still illustrates that our results are almost independent of the sample size. Our functional fSGLD and fSGHMC consistently demonstrate superior performance.

\begin{table*}[!t]
\small
\centering
    \caption{The table shows the average RMSE results for sample size effects (150 samples).}
    \label{tab:size_rmse}
    \begin{sc}
    \begin{tabular}{lcccccc}
     \hline & \multicolumn{4}{c}{\text { RMSE }} &  \\
    \cline { 2 - 5 } & \text { SGLD } & \text { fSGLD } & \text { SGHMC } & \text { fSGHMC } \\
    \hline 
    \text{Yacht} & 
    1.095 $\pm$ 0.128 & 0.409 $\pm$ 0.112 & 1.187 $\pm$ 0.141 & \textbf{0.246} $\pm$ \textbf{0.142} \\
    \text{Boston} & 
    1.248 $\pm$ 0.072 & 0.392 $\pm$ 0.112 & 1.332 $\pm$ 0.068 & \textbf{0.270} $\pm$ \textbf{0.135}  
    \\
    \hline 
    \end{tabular}
    \end{sc}
\end{table*}

\begin{table*}[!t]
\small
\centering
    \caption{The table shows the average NLL results for sample size effects (150 samples).}
    \label{tab:size_nll}
    \begin{sc}
    \begin{tabular}{lcccccc}
     \hline & \multicolumn{4}{c}{\text { NLL }} &  \\
    \cline { 2 - 5 } & \text { SGLD } & \text { fSGLD } & \text { SGHMC } & \text { fSGHMC } \\
    \hline 
    \text{Yacht} & 
    -0.352 $\pm$ 0.073 & -2.332 $\pm$ 0.493 & -2.300 $\pm$ 0.181 & \textbf{-3.406} $\pm$ \textbf{0.517} \\
    \text{Boston} & 
    -0.509 $\pm$ 0.129 & -2.068 $\pm$ 0.159 & -2.352 $\pm$ 0.100 & \textbf{-3.060} $\pm$ \textbf{0.284}  
    \\
    \hline 
    \end{tabular}
    \end{sc}
\end{table*}

\subsection{The effects of the number of measurement points on the gradient estimation of functional prior}
\begin{figure*}[t!]
    \centering
  {
    \subfigure[fSGLD-m10]{\label{fig:fsgld_m10}%
       \includegraphics[width=0.3\linewidth]{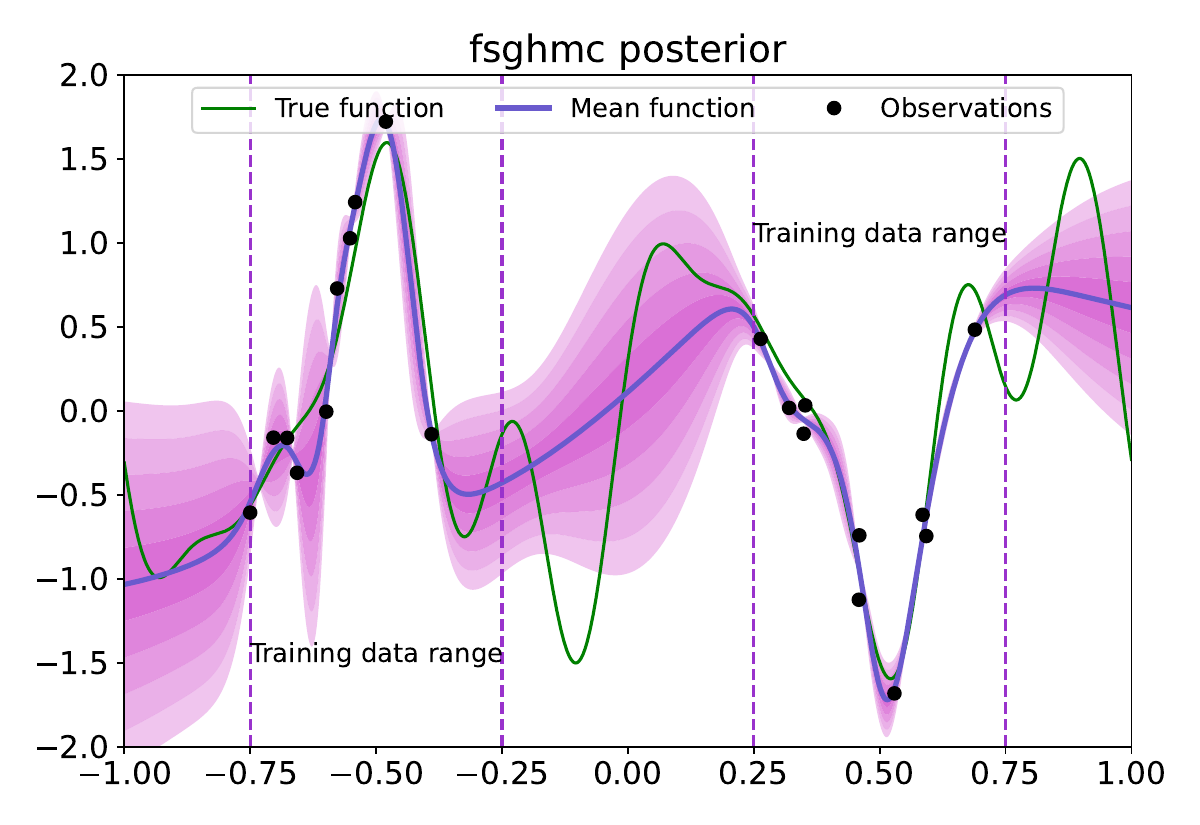}}%
    \subfigure[fSGLD-m40]{\label{fig:fsgld_m40}%
      \includegraphics[width=0.3\linewidth]{./ablation/fsgld_80}}%
    \subfigure[fSGLD-m400]{\label{fig:fsgld_m400}%
      \includegraphics[width=0.3\linewidth]{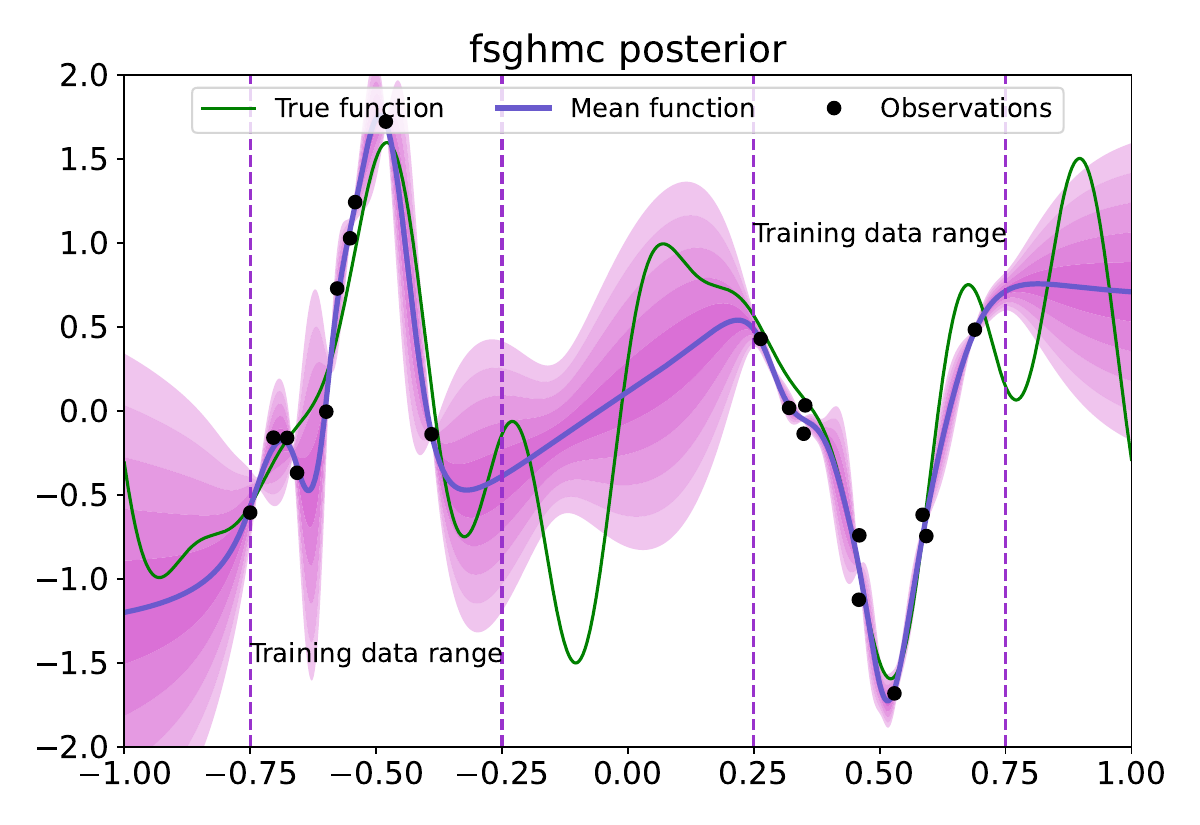}}%
    
    \subfigure[fSGHMC-m10]{\label{fig:fsghmc_m10}%
       \includegraphics[width=0.3\linewidth]{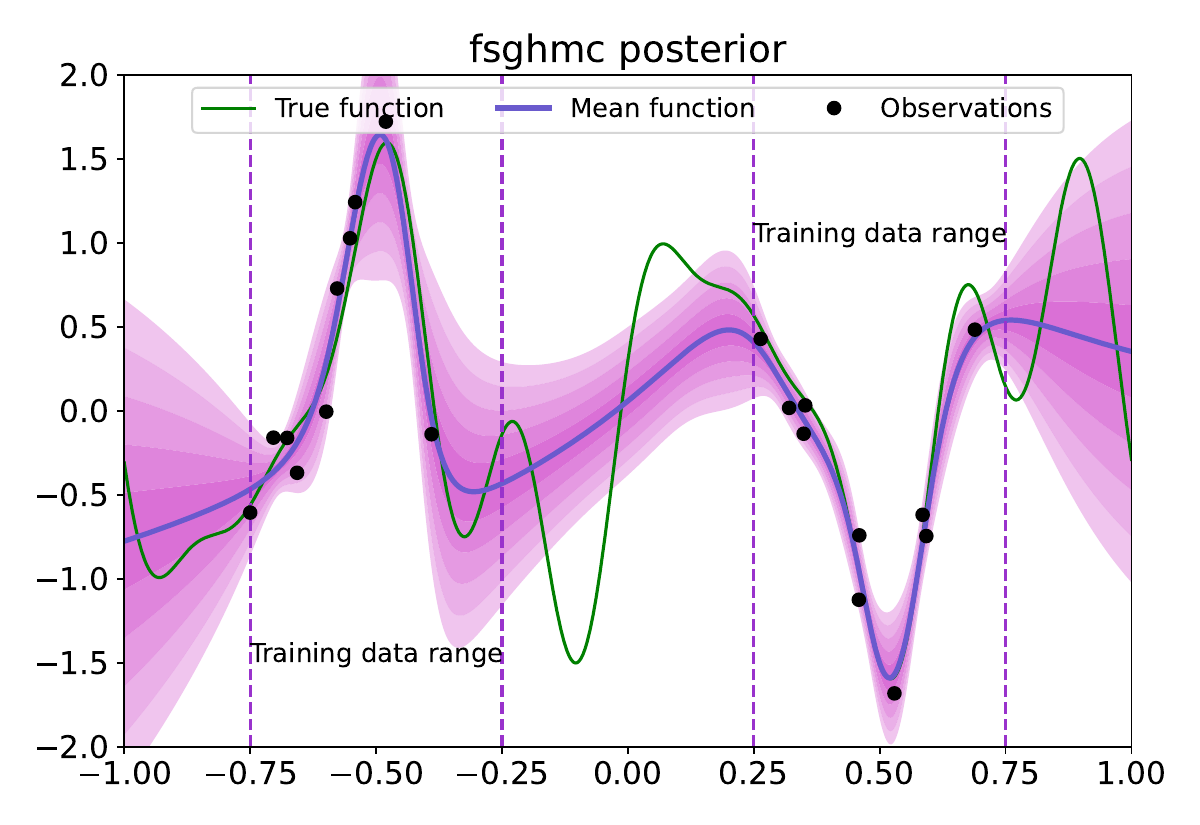}}%
    \subfigure[fSGHMC-m40]{\label{fig:fsghmc_m40}%
       \includegraphics[width=0.3\linewidth]{./ablation/fhmc_70}}%
   \subfigure[fSGHMC-m400]{\label{fig:fsghmc_m400}%
       \includegraphics[width=0.3\linewidth]{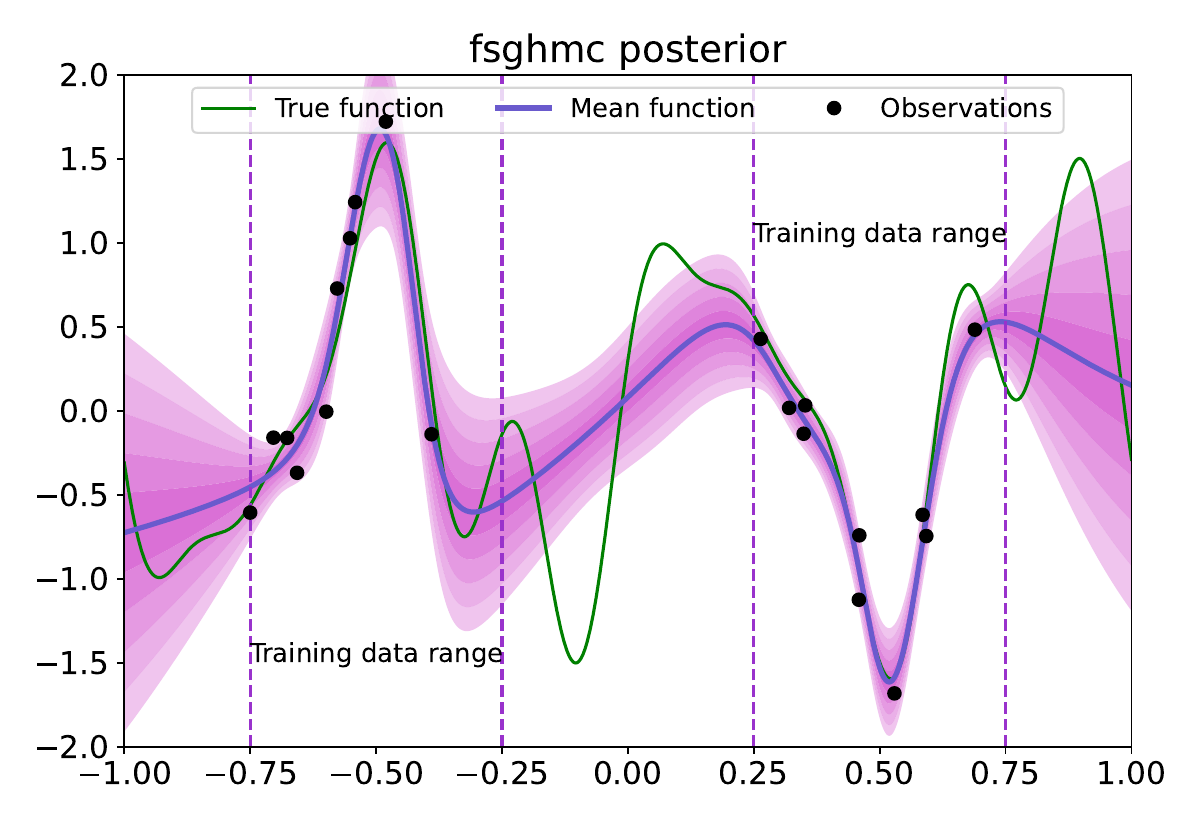}}%
    \caption{The effects of the number of measurement points on the gradient estimation of functional prior for fSGLD and fSGHMC. The number after the short line in each subheading represents the number of measurement points.}
    \label{fig:m_points}}
\end{figure*}

In the 1-D extrapolation example in \cref{sec:1d}, we sampled 40 measurement points from 20 training data and an additional 40 inducing points from Uniform $(-1, 1)$ to estimate the gradient of log functional prior for fSGLD and fSGHMC. We now investigate the effects of the number of measurement points on the approximate estimation of the gradient of the log functional prior in functional SGMCMC. We consider two other cases for the number of measurement points: 10 measurement points sampled from 20 training data and an additional 40 inducing points from Uniform $(-1, 1)$; and 400 measurement points sampled from 20 training data and an additional 1000 inducing points from Uniform $(-1, 1)$. The comparison results are shown in \cref{fig:m_points}. We can see that there is almost no difference in the prediction performance for the three different numbers of measurement points. The uncertainty estimations also fluctuate only slightly. Overall, we can conclude that the number of measurement points for gradient estimation of functional prior does not significantly impact the results of our methods on 1-D extrapolation example.

In the UCI regressions in \cref{sec:uci}, for datasets with sample sizes less than 1000, we estimate the gradient of the functional prior for fSGLD and fSGHMC using the training data as the finite measurement points. For datasets with sample sizes larger than 1000, we randomly sample 1000 data from the training data as measurement points. We now consider using only 40 measurement points on the \textit{Yacht} and \textit{Boston} datasets. The results of RMSE and NLL are shown in \cref{tab:m_rmse} and \cref{tab:m_nll}, respectively. Compared to the results in \cref{tab:rmse} and \cref{tab:nll}, the RMSE results are slightly worse for both fSGLD and fSGHMC. The NLL results for fSGLD get slightly worse on the \textit{Yacht} dataset and remain essentially unchanged on the \textit{Boston} dataset. The NLL results for fSGHMC are slightly worse on both datasets but still significantly outperforms all other methods. Overall, for the UCI regressions, the more measurement points we used, the better the performance of our methods, probably because more measurement points are more accurate for the gradient estimation of the functional prior.

\begin{table}[]
\small
\centering
    \caption{The table shows the average RMSE results for the effects of the number of measurement points.}
    \label{tab:m_rmse}
    \begin{sc}
    \begin{tabular}{lcccccc}
     \hline & \multicolumn{2}{c}{\text { RMSE }} &  \\
    \cline { 2-3 } & \text { fSGLD } & \text { fSGHMC } \\
    \hline 
    \text{Yacht} & 
    0.496 $\pm$ 0.072 & 0.297 $\pm$ 0.111  \\
    \text{Boston} & 
    0.448 $\pm$ 0.098 & 0.290 $\pm$ 0.126
    \\
    \hline 
    \end{tabular}
    \end{sc}
\end{table}

\begin{table}[!t]
\small
\centering
    \caption{The table shows the average NLL results for the effects of the number of measurement points.}
    \label{tab:m_nll}
    \begin{sc}
    \begin{tabular}{lcccccc}
     \hline & \multicolumn{2}{c}{\text { NLL }} &  \\
    \cline { 2-3 } & \text { fSGLD } & \text { fSGHMC } \\
    \hline 
    \text{Yacht} & 
    -2.327 $\pm$ 0.564 & -2.738 $\pm$ 0.428  \\
    \text{Boston} & 
    -2.277 $\pm$ 0.089 & -2.607 $\pm$ 0.329
    \\
    \hline 
    \end{tabular}
    \end{sc}
\end{table}

\subsection{The effects of the functional GP prior}

\begin{figure*}[!t]
    \centering
  {
     \subfigure[GP prior-RBF]{\label{fig:gp_rbf}%
       \includegraphics[width=0.3\linewidth]{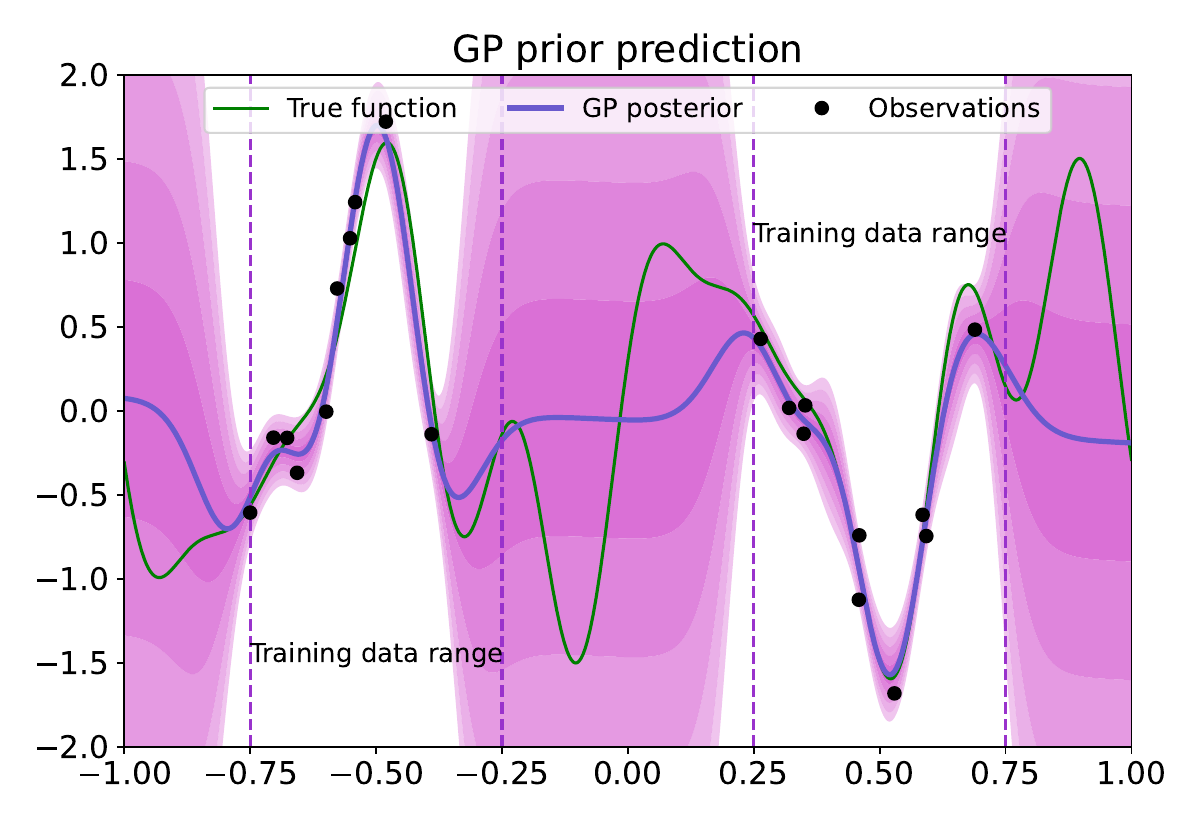}}%
    \subfigure[GP prior-Linear]{\label{fig:gp_linear}%
      \includegraphics[width=0.3\linewidth]{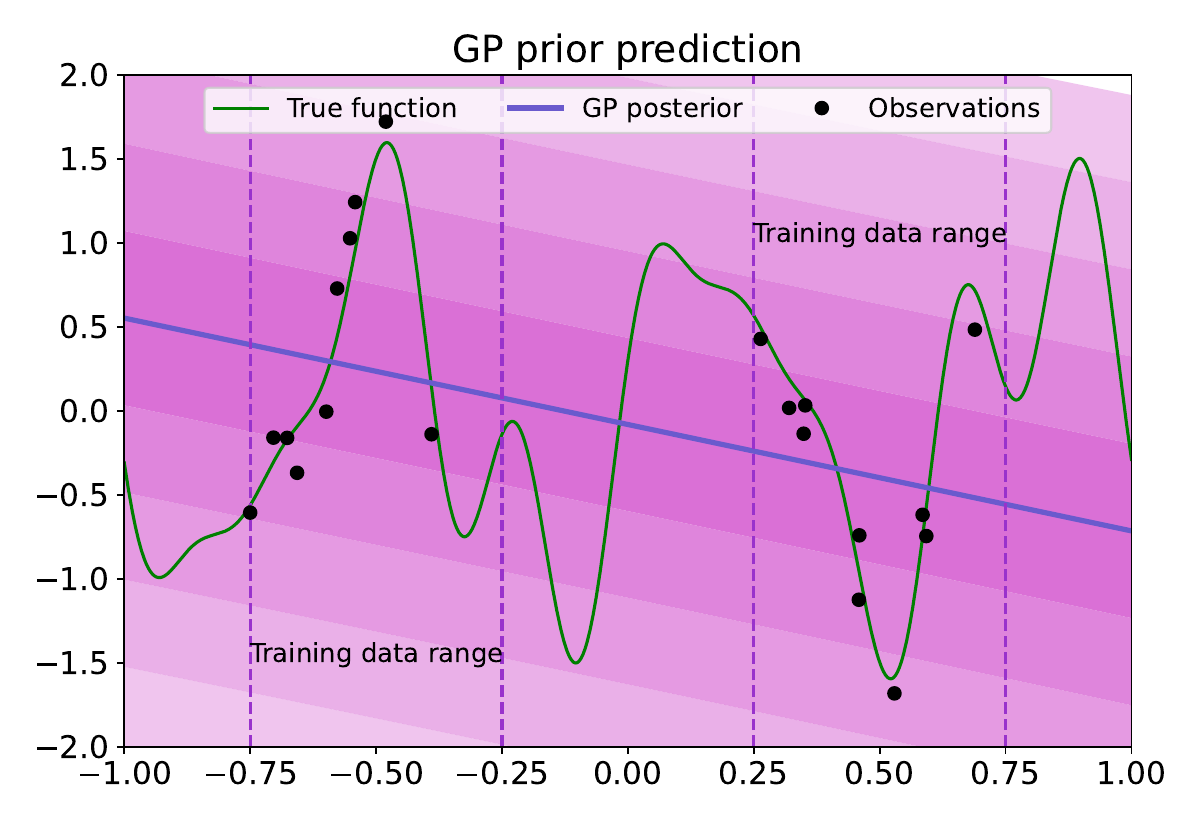}}%
    \subfigure[GP prior-Matern]{\label{fig:gp_matern}%
      \includegraphics[width=0.3\linewidth]{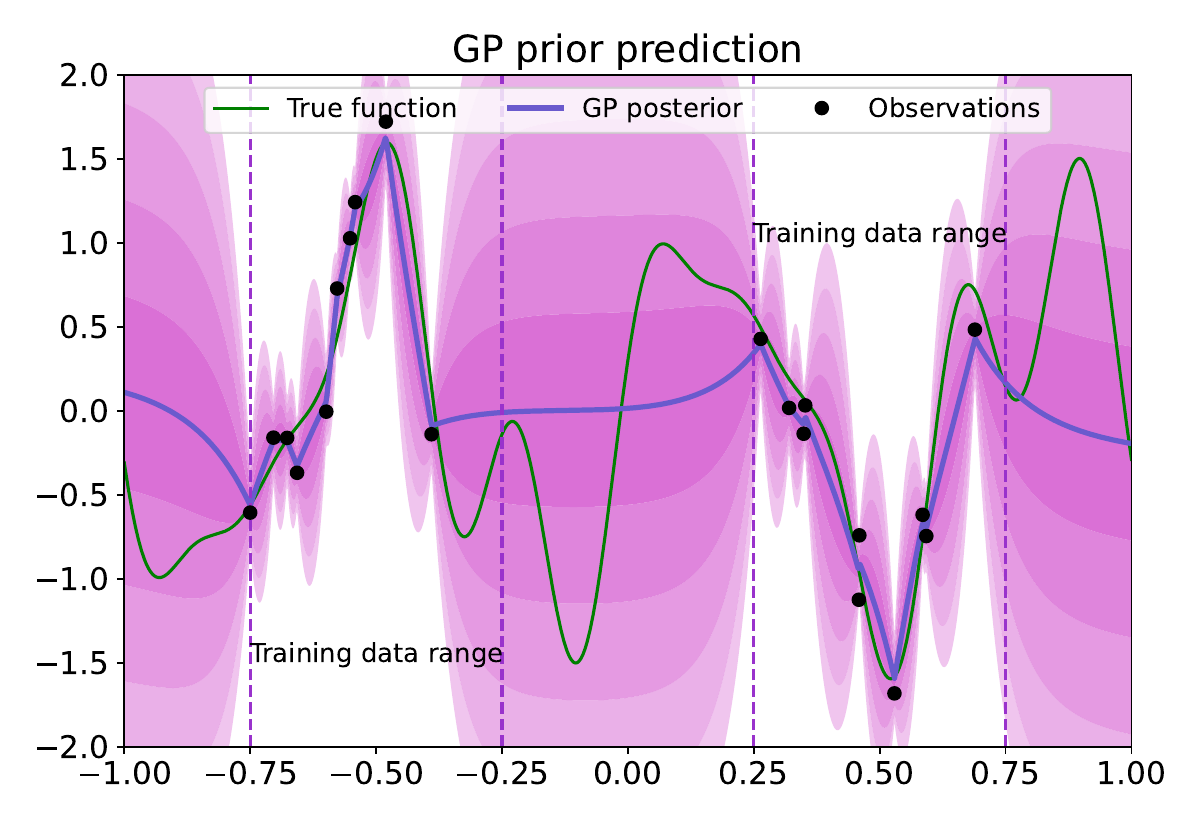}}%
      
    \subfigure[fSGLD-RBF]{\label{fig:fsgld_rbf}%
       \includegraphics[width=0.3\linewidth]{./ablation/fsgld_80}}%
    \subfigure[fSGLD-Linear]{\label{fig:fsgld_linear}%
      \includegraphics[width=0.3\linewidth]{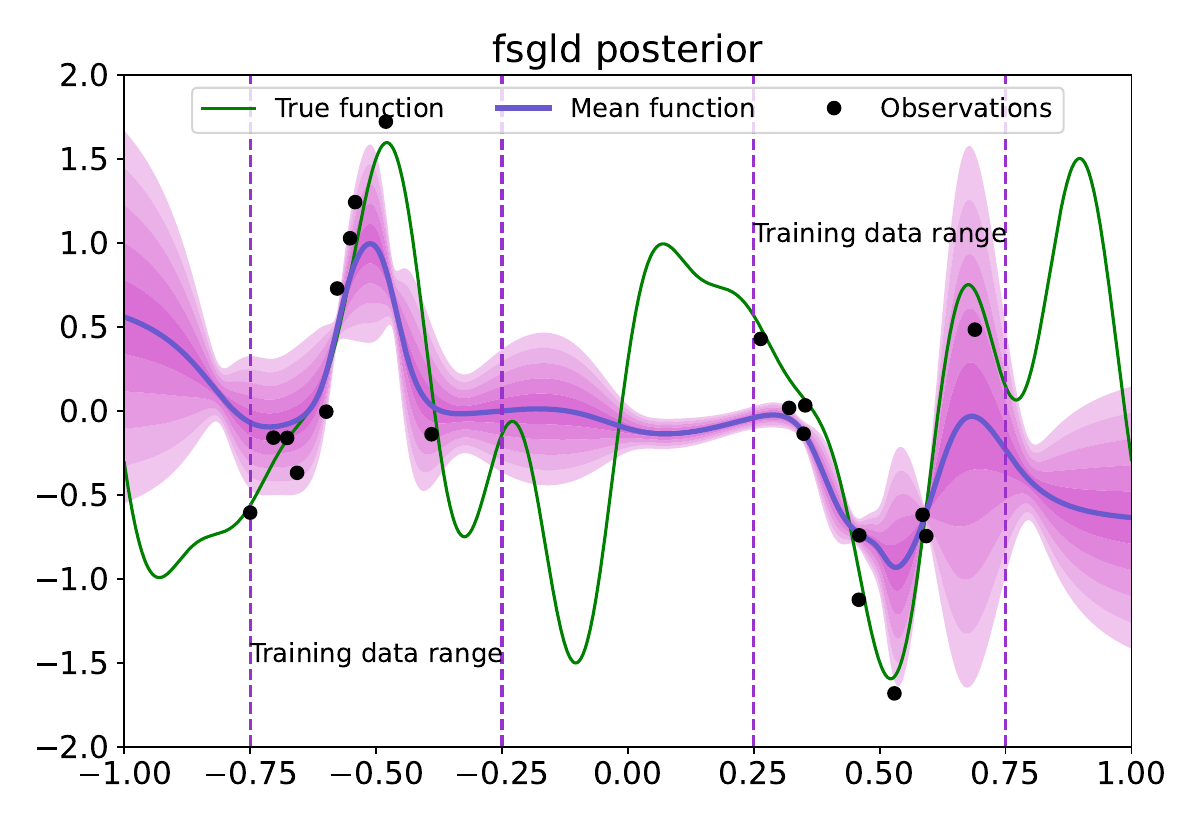}}%
    \subfigure[fSGLD-Matern]{\label{fig:fsgld_matern}%
      \includegraphics[width=0.3\linewidth]{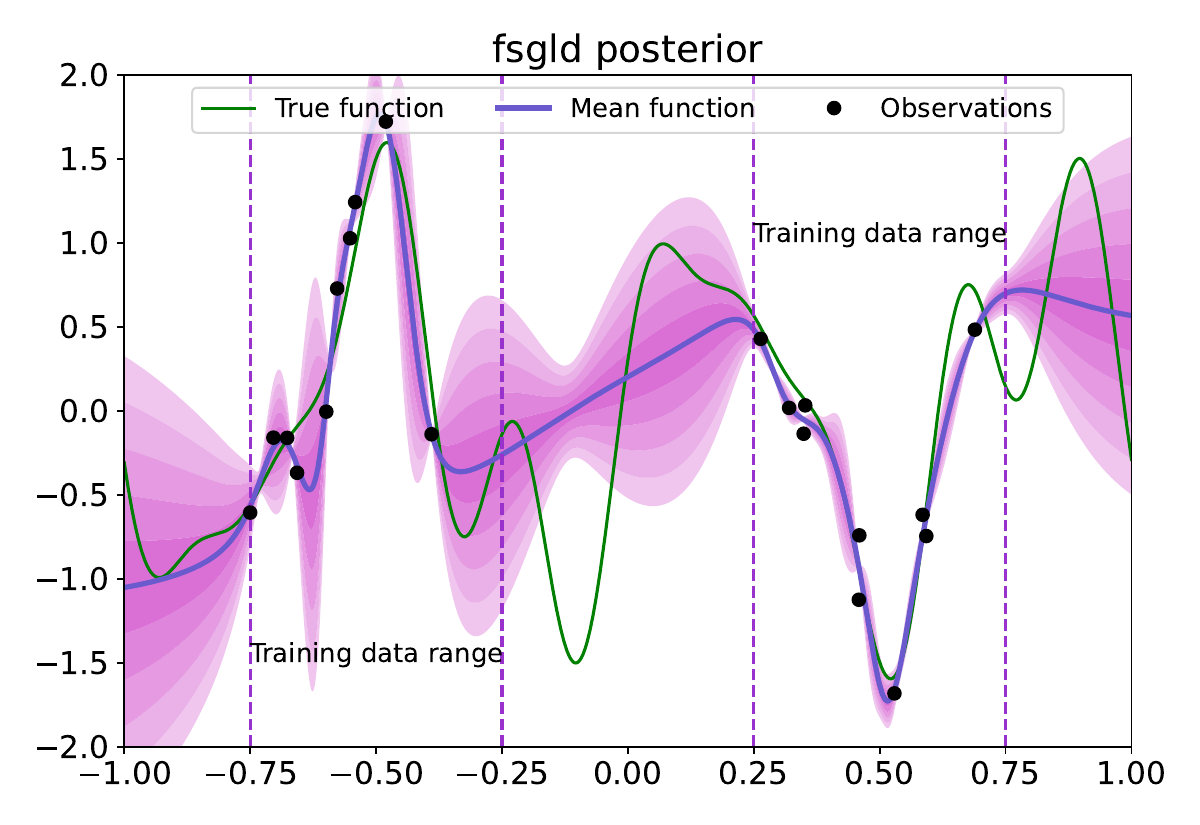}}%
    
    \subfigure[fSGHMC-RBF]{\label{fig:fsghmc_rbf}%
       \includegraphics[width=0.3\linewidth]{./ablation/fhmc_70}}%
    \subfigure[fSGHMC-Linear]{\label{fig:fsghmc_linear}%
       \includegraphics[width=0.3\linewidth]{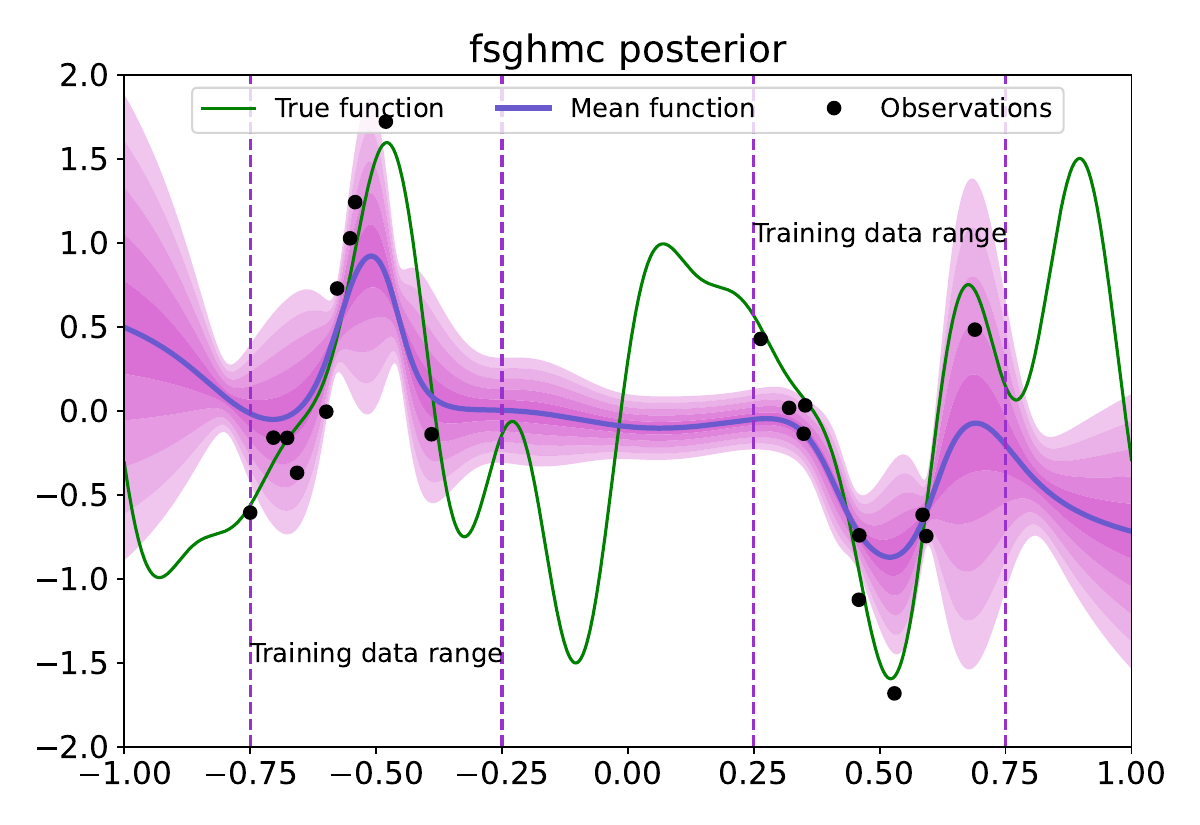}}%
   \subfigure[fSGHMC-Matern]{\label{fig:fsghmc_matern}%
       \includegraphics[width=0.3\linewidth]{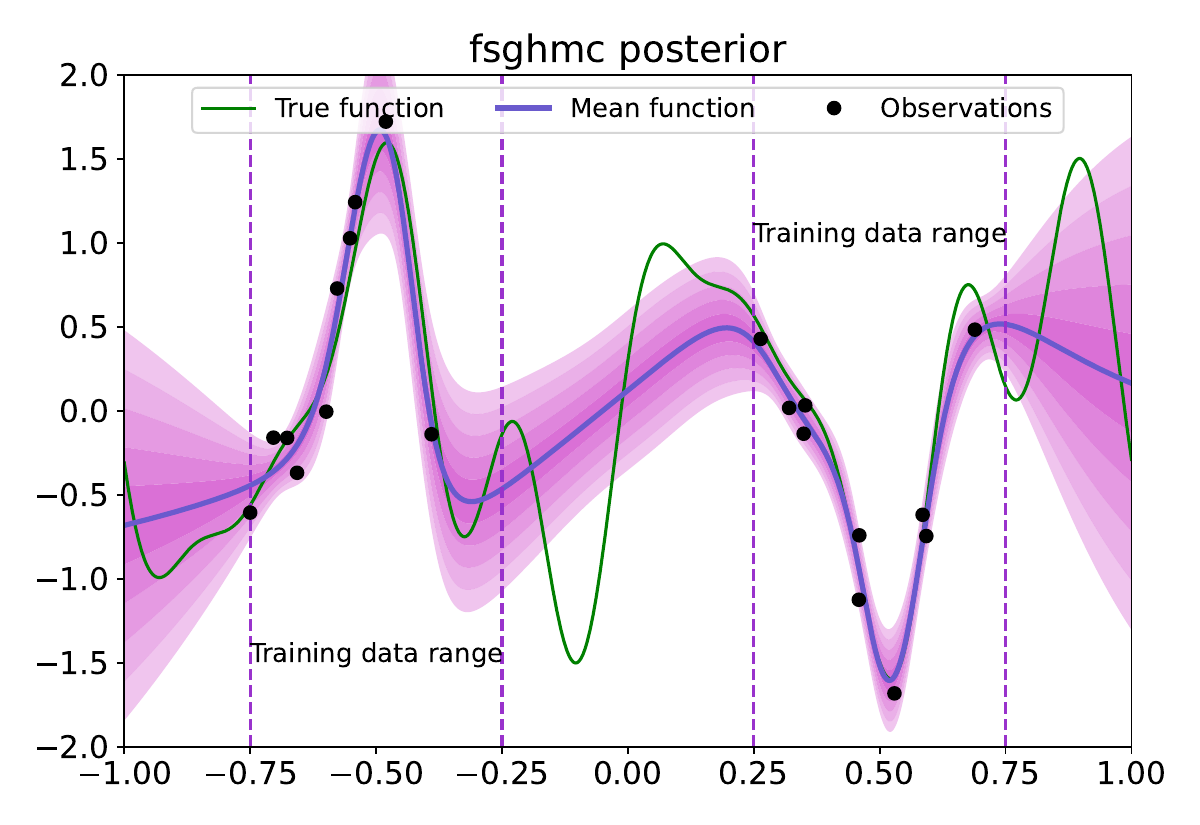}}%
    \caption{The effects of the kernel function on the functional GP prior. The text after the short line in each subheading represents the corresponding kernel functions.}
    \label{fig:kernel}}
\end{figure*}
Functional Gaussian processes (GPs) priors are able to encode prior knowledge about function properties (e.g., periodicity and smoothness) through corresponding kernel functions. We used RBF kernel in the 1-D extrapolation experiment, which is suited for modelling polynomial functions. We now consider two other kernel functions of GP priors to our functional SGMCMC: Matern kernel and Linear kernel (unsuitable for modelling polynomial oscillatory curves). The results are shown in \cref{fig:kernel}. The top row presents the GP prior predictions for RBF kernel, Linear kernel and Matern kernel, respectively. The results from the mismatched Linear kernel in \cref{fig:fsgld_linear} and \cref{fig:fsghmc_linear} for fSGLD and fSGHMC, respectively, show that the fitting performance deteriorates in the observation region and also demonstrate a strong linear trend (converging to a horizontal line) in the middle unseen region $[-0.25, 0.25]$ since the Linear GP prior prediction shows a completely linear trend and falls to fit the target function. The uncertainty intervals in the leftmost $[-1, -0.75]$ and rightmost $[0.75, 1]$ non-observation regions deviate far from the true function. On the contrary, the results of the Matern kernel are hardly different from those of the RBF kernel due to the fact that it also has a strong ability to describe polynomial functions. This indicates that our method can effectively incorporate functional prior information into the posterior inference. 

\begin{figure*}[t!]
    \centering
  {
    \subfigure[fSGLD]{\label{fig:fsgldpre}%
       \includegraphics[width=0.3\linewidth]{./toy/fsgld}}%
    \subfigure[fSGLD]{\label{fig:fsgldnopre}%
      \includegraphics[width=0.3\linewidth]{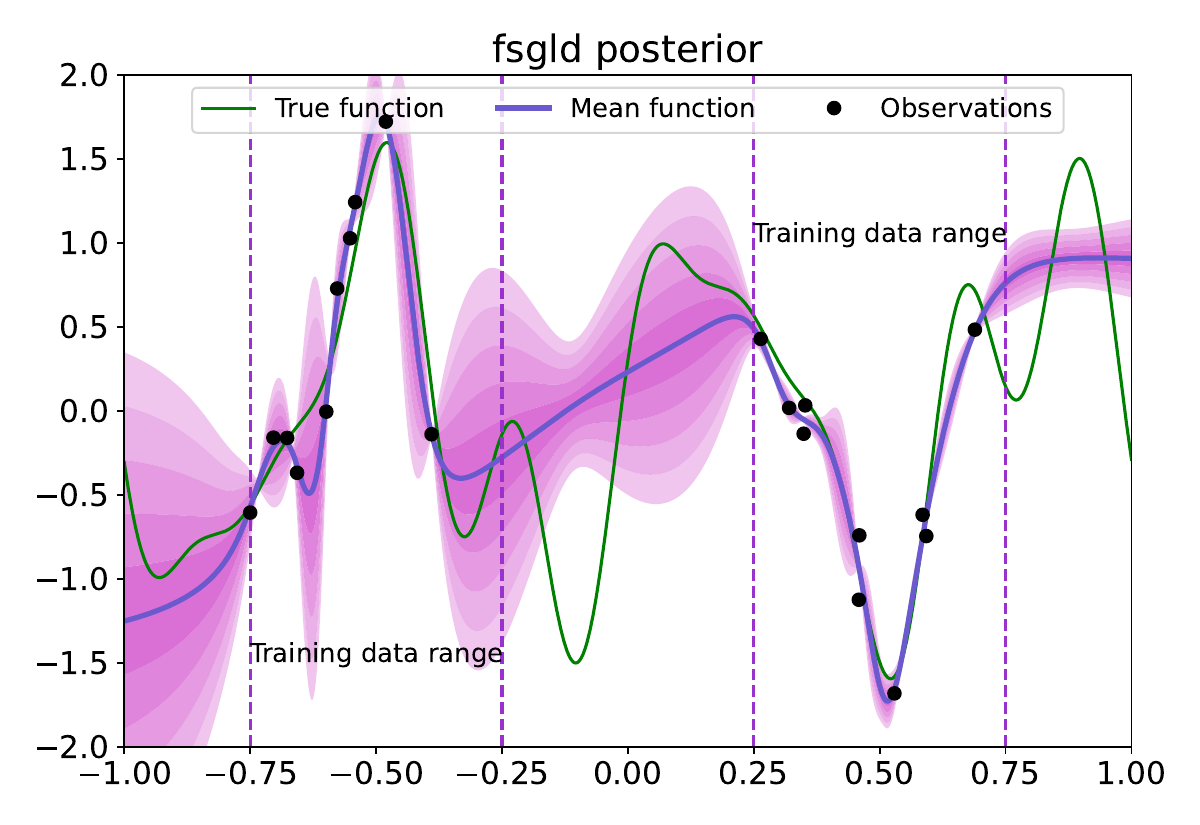}}%
   
   \subfigure[fSGHMC]{\label{fig:fhmcpre}%
      \includegraphics[width=0.3\linewidth]{./toy/fhmc2}}%
    \subfigure[fSGHMC]{\label{fig:fhmcnopre}%
      \includegraphics[width=0.3\linewidth]{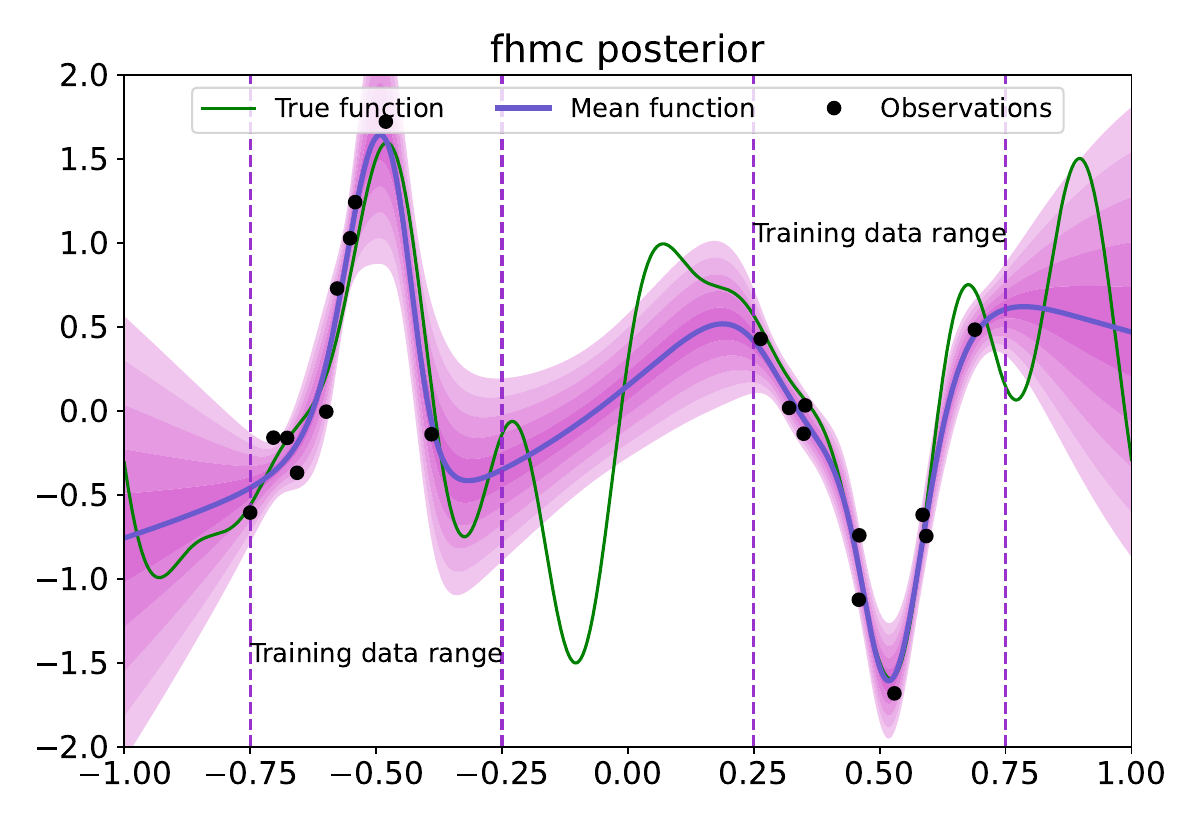}}%
    \caption{1-D extrapolation example without prior pre-training. (a) and (c) are the results with pre-trained prior; (b) and (d) are results without pre-trained prior.}
    \label{fig:nopre}}
\end{figure*}

We also plot the predictions of fSGLD and fSGHMC with no pre-trained GP priors in \cref{fig:nopre}. The leftmost columns, \cref{fig:fsgldpre} and \cref{fig:fhmcpre}, show the results for fSGLD and fSGHMC with pre-trained GP priors, as described in the main paper. The rightmost columns, \cref{fig:fsgldnopre} and \cref{fig:fhmcnopre}, display the predictions without pre-trained GP priors, using a fixed constant mean function $m(x)=0$ and a kernel function $k(x, x') = 1$. Despite not having pre-trained priors, the posterior results remain quite similar to those with pre-trained GP priors, indicating that our methods are robust to the absence of prior pre-training. In \cref{tab:ucinopre}, we report the RMSE and NLL results of fSGLD and fSGHMC on the UCI regression (Yacht) dataset without pre-trained GP priors. Even though the fixed GP priors without pre-training show slight degradation, the performance remains competitive compared to other baseline methods.

\begin{table}[t]
\caption{RMSE and NLL of fSGLD and fSGHMC on UCI regression (yacht) with no pre-trained GP priors.}
\label{tab:ucinopre}
\vskip 0.15in
\begin{center}
\begin{small}
\begin{sc}
\begin{tabular}{lcccc}
\toprule
Model & pre-trained GP & fixed GP  \\
\midrule
RMSE   \\
fSGLD    & 0.41 & 0.58 \\
fSGHMC   & 0.25 & 0.51  \\
\midrule
NLL   \\
fSGLD    & -2.46 & -1.65  \\
fSGHMC   & -3.37 & -2.52  \\
\bottomrule
\end{tabular}
\end{sc}
\end{small}
\end{center}
\vskip -0.1in
\end{table}

\begin{figure*}[t!]
    \centering
  {
    \subfigure[GP-100 epochs]{\label{fig:gp100}%
       \includegraphics[width=0.3\linewidth]{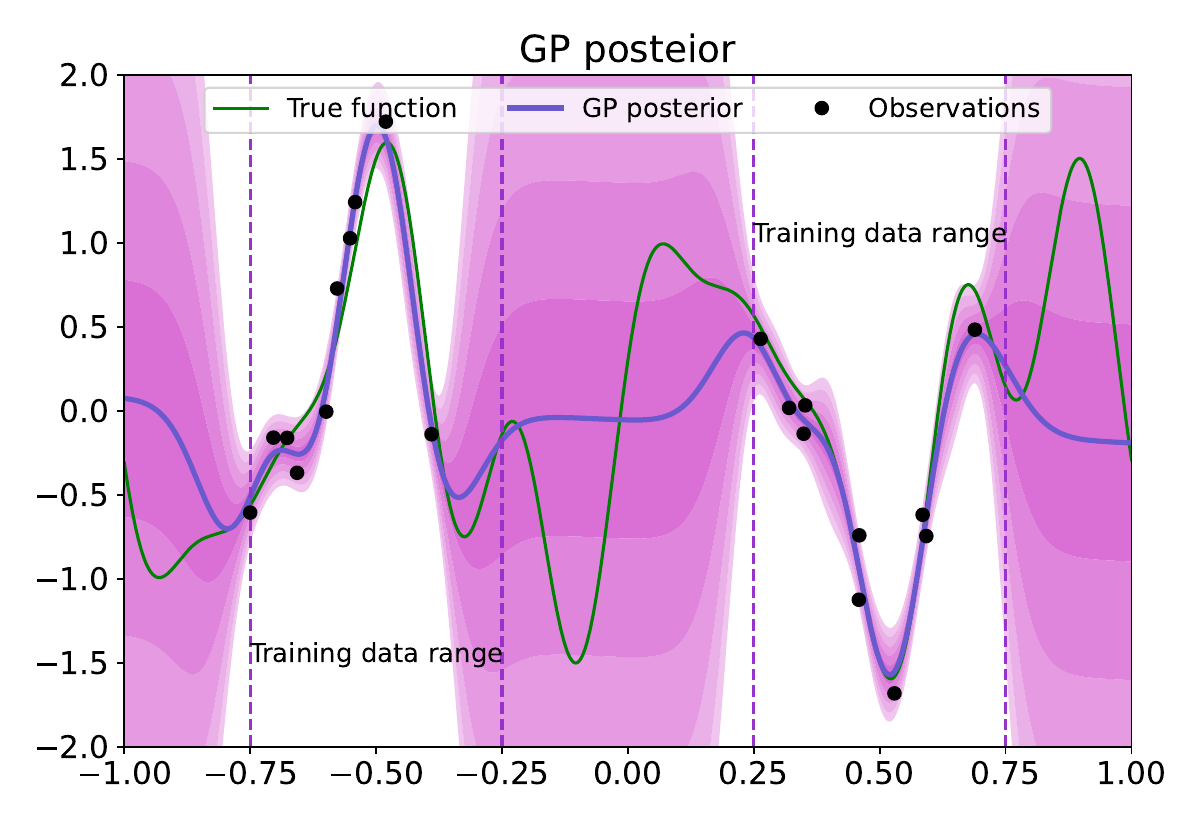}}%
    \subfigure[fSGLD]{\label{fig:fsgld100}%
      \includegraphics[width=0.3\linewidth]{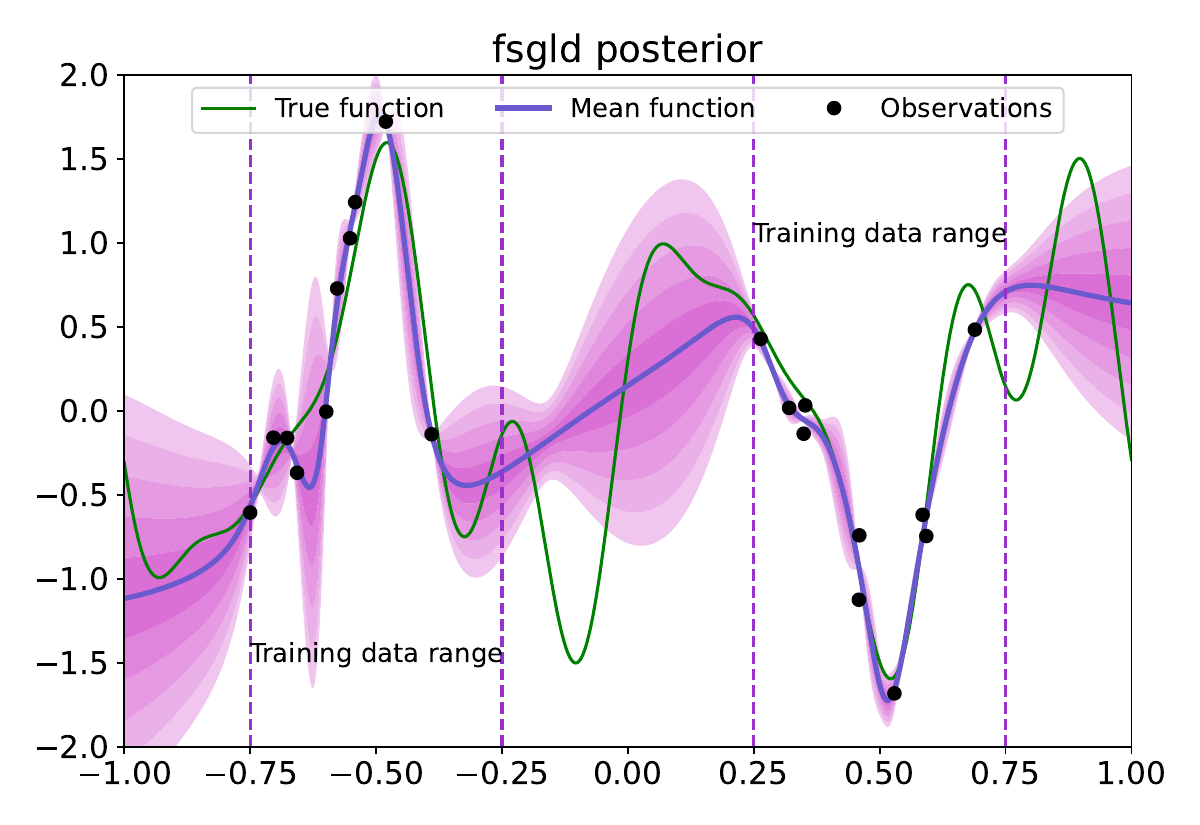}}%
    \subfigure[fSGHMC]{\label{fig:fhmc100}%
      \includegraphics[width=0.3\linewidth]{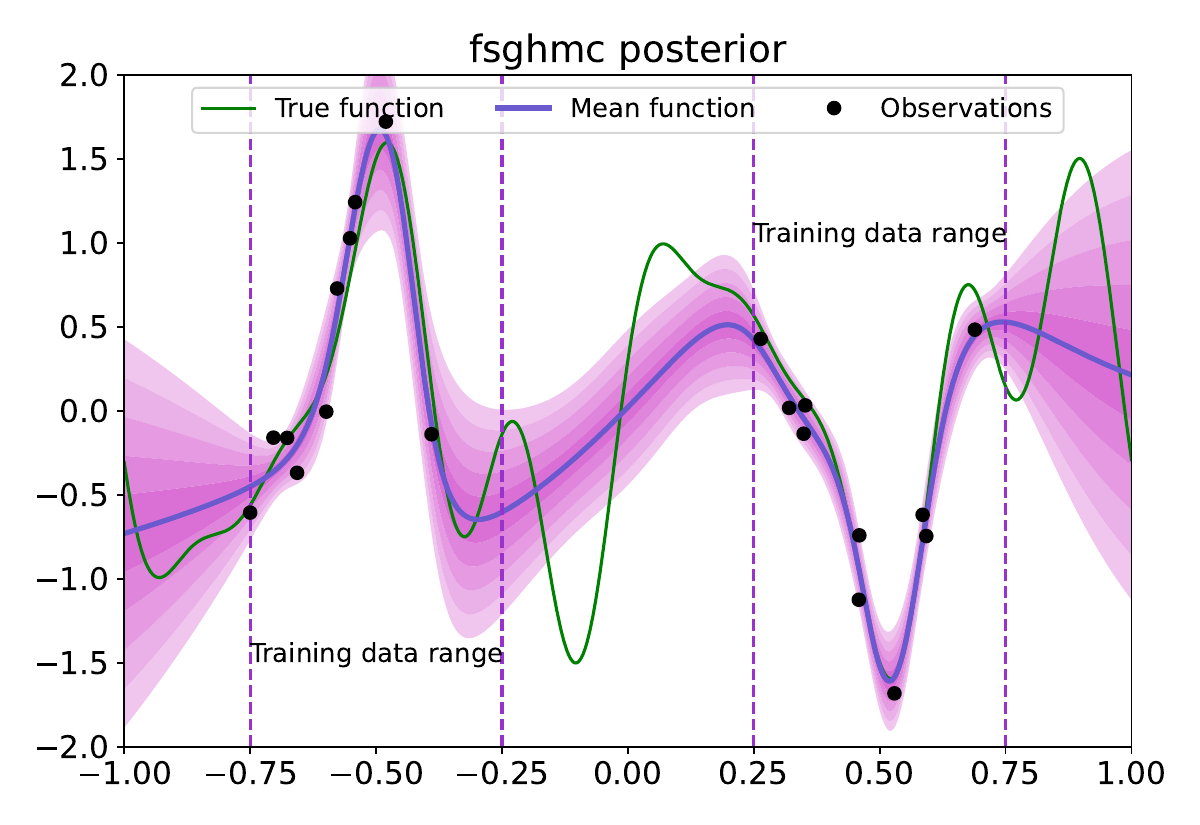}}%
    
  \subfigure[GP-20 epochs]{\label{fig:gp20}%
       \includegraphics[width=0.3\linewidth]{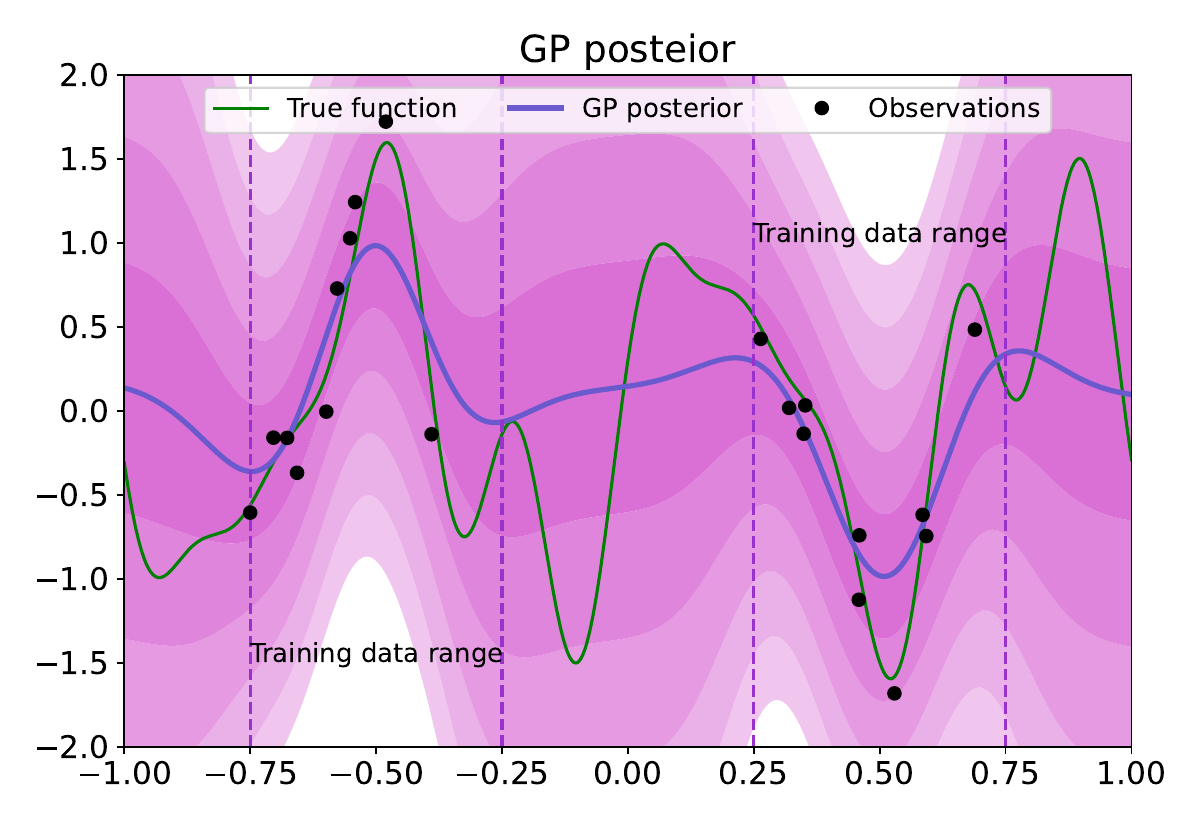}}%
    \subfigure[fSGLD]{\label{fig:fsgld20}%
      \includegraphics[width=0.3\linewidth]{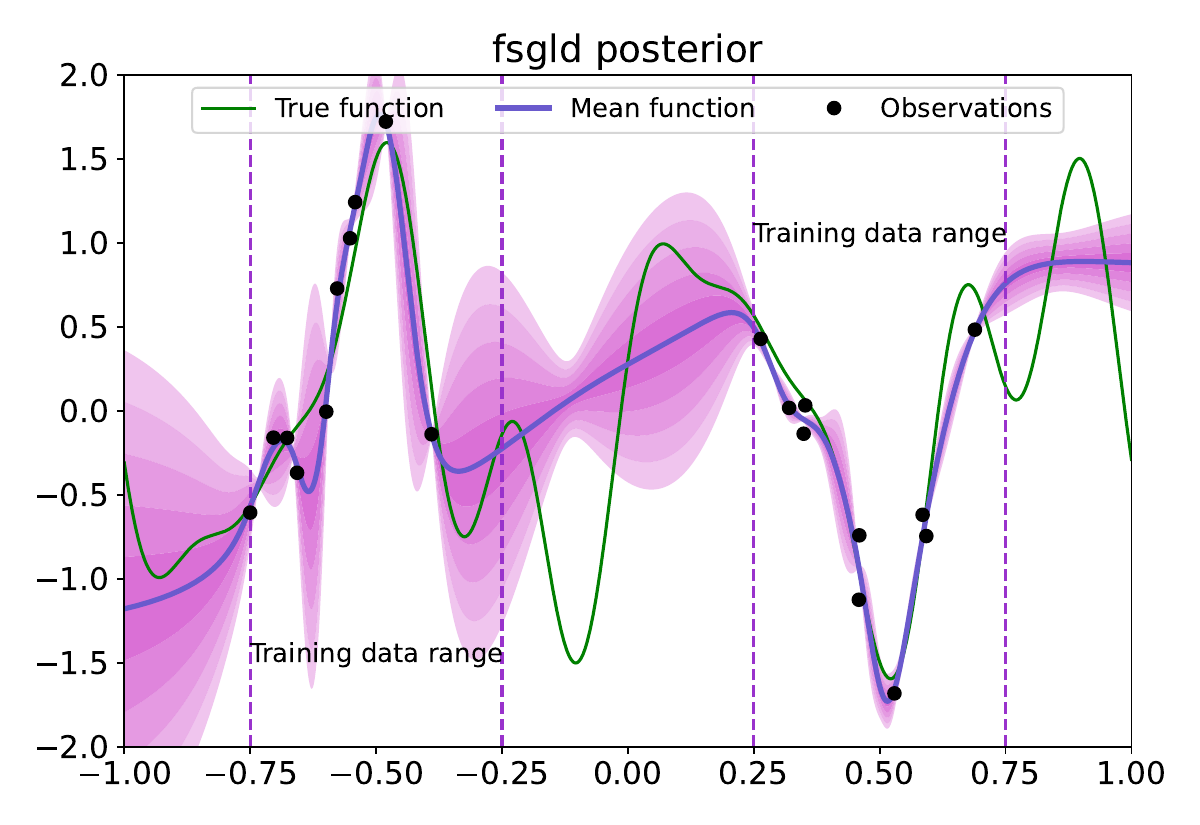}}%
    \subfigure[fSGHMC]{\label{fig:fhmc20}%
      \includegraphics[width=0.3\linewidth]{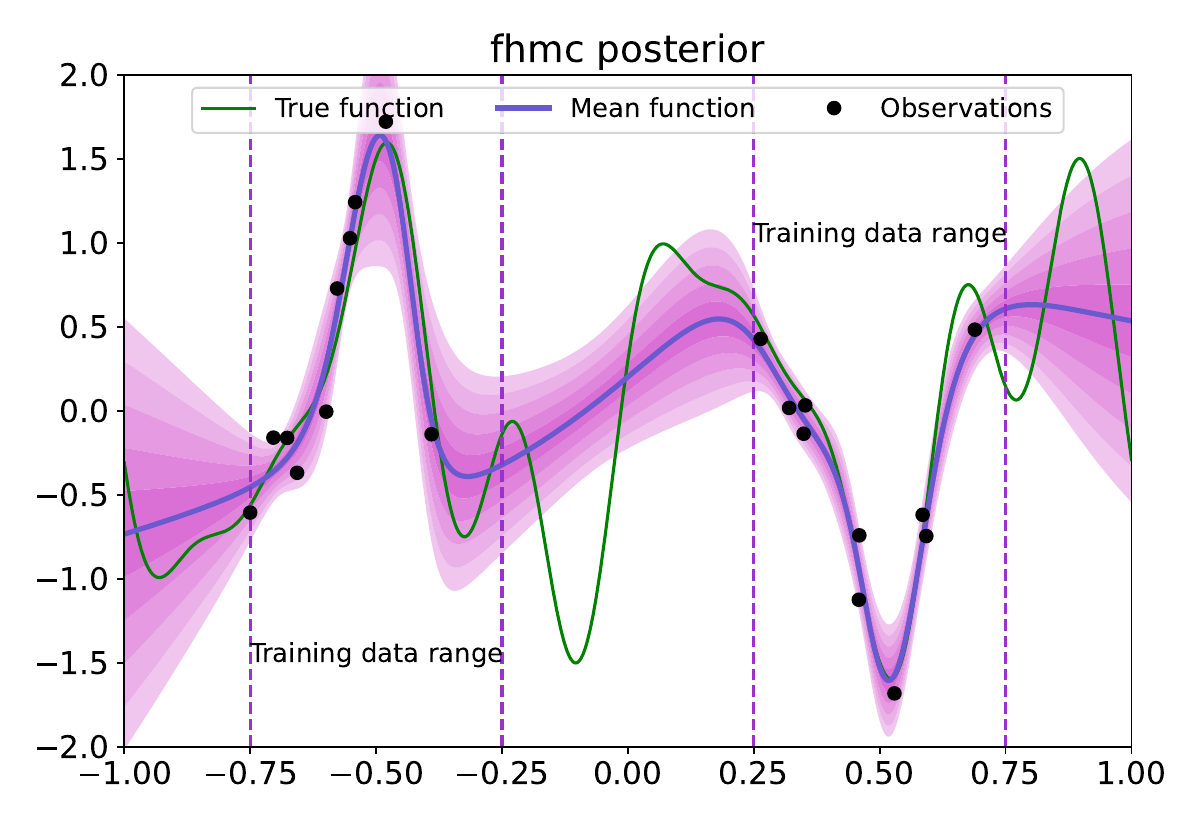}}%

     \subfigure[GP-10 epochs]{\label{fig:gp10}%
       \includegraphics[width=0.3\linewidth]{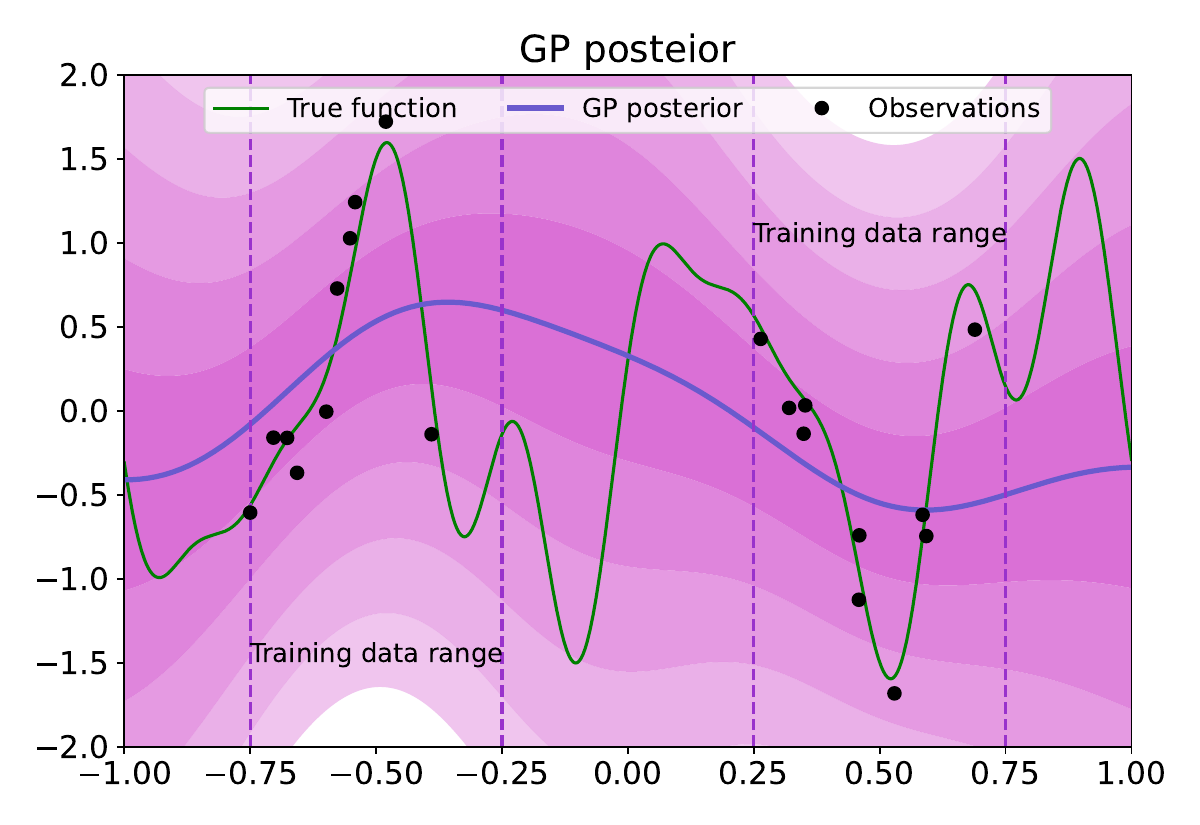}}%
    \subfigure[fSGLD]{\label{fig:fsgld10}%
      \includegraphics[width=0.3\linewidth]{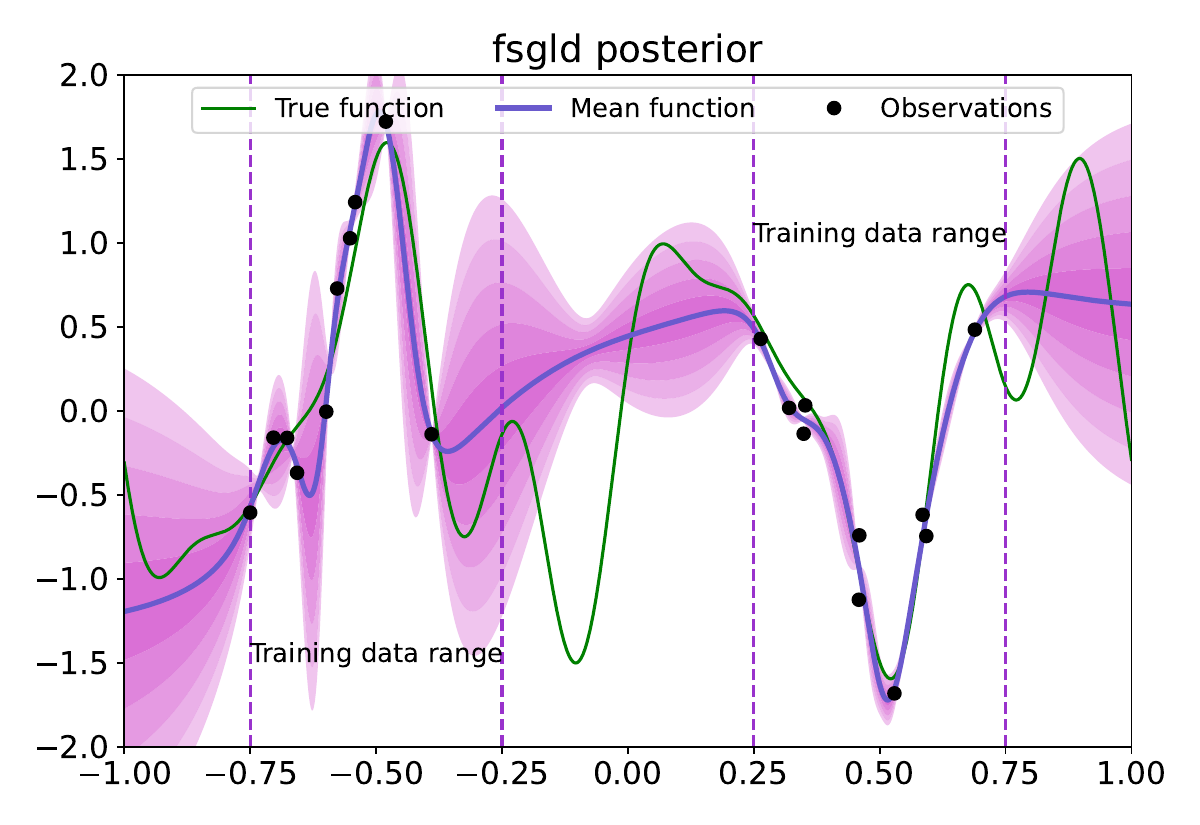}}%
    \subfigure[fSGHMC]{\label{fig:fhmc10}%
      \includegraphics[width=0.3\linewidth]{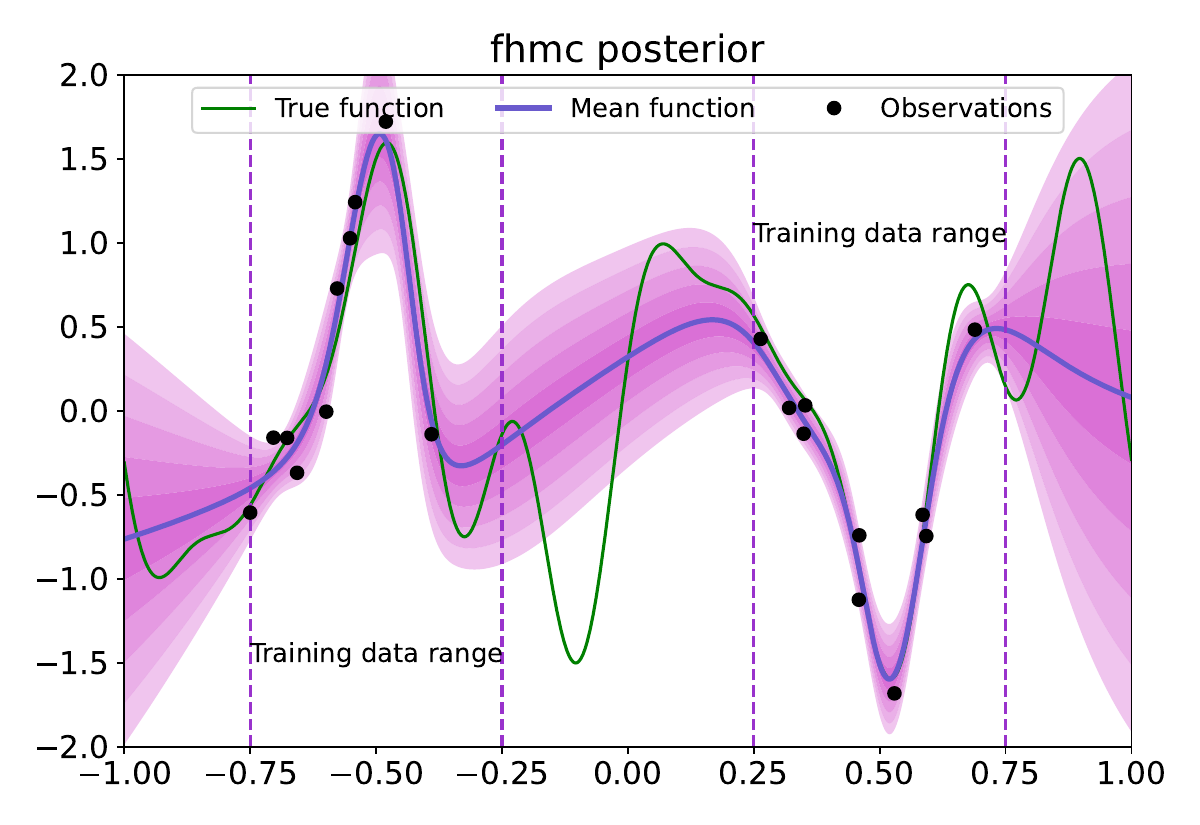}}%
    \caption{1-D extrapolation example of different pre-training epochs for GP prior. The leftmost column shows the predictions of GP prior with three different pre-training epochs: 100, 20, and 10 epochs from top to bottom. The middle and the rightmost columns are the corresponding fSGLD and fSGHMC posteriors from these three GP prior, respectively.}
    \label{fig:pre_epo}}
\end{figure*}

Furthermore, we analyze the influence of different pre-training epochs for the GP priors by comparing results with 100, 20, and 10 pre-training epochs. \cref{fig:pre_epo} provides a detailed comparison: the leftmost columns, \cref{fig:gp100}, \cref{fig:gp20} and \cref{fig:gp10} depict the predictions of the three different GP priors, while the middle and rightmost columns illustrate the fSGLD and fSGHMC posterior results from these priors, respectively. Interestingly, while the predictions from the three GP priors differ significantly, there is no noticeable difference in the corresponding fSGLD and fSGHMC posteriors. This observation highlights that the effectiveness of our methods does not strongly depend on the GP prior pre-training procedure, making them more flexible in practical applications.

\end{document}